\documentclass[11pt]{article}
\pdfoutput = 1
\usepackage{amsmath}
\usepackage{bm}
\usepackage[numbers]{natbib}
\usepackage[colorlinks,citecolor=blue,linkcolor=blue,urlcolor=blue]{hyperref}
\usepackage{latexsym, epsfig, amssymb, amsmath, amsthm, graphicx, mathrsfs, booktabs}
\usepackage{bbm}
\usepackage{rotating, tabularx, setspace,paralist}
\usepackage{color}
\usepackage{graphicx}
\usepackage[OT1]{fontenc}
\usepackage{lscape}
\usepackage{hyperref}
\usepackage{multirow,multicol}
\usepackage{longtable, caption}
\usepackage{anysize}
\renewcommand{\baselinestretch}{1.2}
\marginsize{1.1in}{1.1in}{0.55in}{0.8in} %

\makeatletter
\def\singlespace{\def\baselinestretch{1}\@normalsize}

\newtheorem{condition}{Condition}
\newtheorem{lemma}{Lemma}
\newtheorem{proposition}{Proposition}
\newtheorem{theorem}{Theorem}
\newtheorem{definition}{Definition}
\newtheorem{remark}{Remark}
\newtheorem{corollary}{Corollary}
\renewcommand{\theequation}{
\arabic{equation}%
}

\makeatother

\newcommand{\bx}{\mbox{\bf x}}

\newcommand{\Var}{\mathrm{Var}}

\newcommand{\supp}{\mathrm{supp}}

\renewcommand{\hat}{\widehat}
\def\BS{\textbf}
\usepackage{mathtools}

\usepackage{comment}
\renewcommand{\dots}{\cdots}

\def\toD{\overset{\mathscr{D}}{\longrightarrow}}

\def\independenT#1#2{\mathrel{\setbox0\hbox{$#1#2$}%
\copy0\kern-\wd0\mkern4mu\box0}}
\newcommand{\indep}{\perp \!\!\! \perp}

\begin{document}

\title{Dimension-Free Average Treatment Effect Inference with Deep Neural Networks%
\thanks{
Xinze Du is Ph.D. candidate, Department of Mathematics, University of Southern California, Los Angeles, CA 90089 (E-mail: \textit{xinzedu@usc.edu}). %
Yingying Fan is Centennial Chair in Business Administration and Professor, Data Sciences and Operations Department, Marshall School of Business, University of Southern California, Los Angeles, CA 90089 (E-mail: \textit{fanyingy@marshall.usc.edu}). %
Jinchi Lv is Kenneth King Stonier Chair in Business Administration and Professor, Data Sciences and Operations Department, Marshall School of Business, University of Southern California, Los Angeles, CA 90089 (E-mail: \textit{jinchilv@marshall.usc.edu}). %
Tianshu Sun is Robert R. Dockson Associate Professor in Business Administration, Data Sciences and Operations Department, Marshall School of Business, University of Southern California, Los Angeles, CA 90089 (E-mail: \textit{tianshus@marshall.usc.edu}). %
Patrick Vossler is Ph.D. candidate, Data Sciences and Operations Department, Marshall School of Business, University of Southern California, Los Angeles, CA 90089 (E-mail: \textit{pvossler@marshall.usc.edu}). %
This work was supported by NSF Grants DMS-1953356 and EF-2125142.
}
\date{November 30, 2021}
\author{Xinze Du, Yingying Fan, Jinchi Lv, Tianshu Sun and Patrick Vossler
\medskip\\
University of Southern California
\\
} %
}

\maketitle

\begin{abstract}
This paper investigates the estimation and inference of  the average treatment effect (ATE) using deep neural networks (DNNs) in the potential outcomes framework. Under some regularity conditions, the observed response can be formulated as the response of a mean regression problem with both the confounding variables and the treatment indicator as the independent variables. Using such formulation, we investigate two methods for ATE estimation and inference based on the estimated mean regression function via DNN regression using a specific network architecture. We show that both DNN estimates of ATE are consistent with dimension-free consistency rates under some assumptions on the underlying true mean regression model. Our model assumptions accommodate the potentially complicated dependence structure of the observed response on the covariates, including latent factors and nonlinear interactions between the treatment indicator and confounding variables.  We also establish the asymptotic normality of our estimators based on the idea of sample splitting, ensuring precise inference and uncertainty quantification. Simulation studies and real data application justify our theoretical findings and support our DNN estimation and inference methods. 
\end{abstract}

\textit{Running title}: ATED


\section{Introduction} \label{Sec1}

The estimation and inference of the average treatment effect (ATE) are foundational research topics in causal inference.  Under the potential outcomes framework, the observed outcome $Y$ of a unit corresponds to one of two potential outcomes; one value for when the unit receives treatment and the other value for when the unit does not.  The average treatment effect is defined as the population mean of the difference between these two potential outcomes.  Since only one of the two potential outcomes can be observed for each unit,  the estimation of ATE faces the common challenges in missing data problems.    There is a large literature on ATE estimation. To name a few, see, for example, \cite{athey2015machine1, athey2015machine, Demirkayaetal2021,LouizosEtAl_arxiv17,ShalitEtAl_icml17,Zhangetal2016}. See also the recent review papers \cite{Matias-survey2018, Imbens-survey2009} on the existing methods and some new developments.  

Under some regularity conditions,  the observed outcome $Y$ can be formulated as the response of a mean regression problem with covariates $(\BS X^{\top}, T)^{\top}$, where $\BS X$ is the  vector of covariates measuring the characteristics of the unit and $T$ is the treatment indicator taking values 0 and 1.  Here, $T=1$ means that the unit receives the treatment and $T=0$ otherwise.   Under such a model assumption,  the average treatment effect is the expected difference of the mean regression functions corresponding to the treated and untreated groups.  This motivates the estimation of ATE based on the estimated mean regression function, giving rise to the projection and imputation estimate \cite{Matias-survey2018}.  

In the era of big data, we have the luxury of collecting many covariates for each unit.  Since it is generally challenging to test for confounding, a conservative approach
is to include most, if not all, covariates with the aim of making the unconfoundedness assumption approximately correct.   However,  the large number of covariates,  together with the potentially complicated interactions between covariates $\BS X$ and the treatment indicator $T$, increases the challenge of ATE estimation and inference.    
On the one hand,  while parametric regression models are relatively robust to the increased dimensionality of covariates,  they impose stringent model structure assumptions which are unlikely to hold in practice, causing the issue of model misspecification. On the other hand,  nonparametric models are much more flexible with mild model structure assumptions,  but they can suffer from the curse of dimensionality, resulting in slower convergence rates. As a result, statistical inference, such as confidence interval construction, is more challenging in the nonparametric setting.

This paper explores two methods for ATE estimation and inference based on the nonparametric method of deep neural networks (DNNs) with theoretical underpinning.  In recent years,  DNNs have been popularly used to model the potentially complicated dependence structure of the response on covariates, thanks to their attractive approximation power.  We first propose directly applying DNN for estimating the underlying mean regression function and then constructing an ATE estimate based on the estimated mean regression function.  To overcome the curse of dimensionality,  we adopt the specific deep neural network structure introduced and theoretically investigated in \cite{AOS2019Bauer}.  Such a network is recursively defined using some specifically designed two-layer neural networks as building blocks. As a result, some layers of the DNN  are only sparsely connected.  The specific structure of the DNN ensures dimension-free convergence rate of the resulting nonparametric mean regression estimate, as formally revealed in \cite{AOS2019Bauer}.   

Although elegant, the results in \cite{AOS2019Bauer} are not directly applicable to  our current model setting, mainly because of the discrete treatment indicator $T$,  resulting in the nonsmoothness of  the mean regression function with respect to its covariates.  Similar to most other nonparametric regression methods,  the theoretical study of the  DNN estimate in \citep{AOS2019Bauer} requires that the mean regression function has enough smoothness with respect to all covariates.  To adapt the theory to our setting, we define a new function that linearly interpolates the values of the true mean regression function when $T=0$ and $T=1$.  This new function has enough smoothness with respect to all its covariates, and thus the theory developed in \citep{AOS2019Bauer} is applicable. We emphasize that this technical treatment is only for theoretical derivation and does not affect the practical implementation.  In fact,  the intermediate values of the newly constructed mean regression function when $T\in (0,1 )$ are not used in our applications.  

An ATE estimate based on the empirical mean over the same data for fitting the DNN can be obtained with the estimated mean regression function.  We show that such an estimate is asymptotically consistent in estimating the true ATE, and the consistency rate is dimension-free, depending only on the smoothness parameter and another parameter controlling the number of hidden neurons.  This result is consistent with that in \citep{AOS2019Bauer}.  However, despite the nice property of dimension-free consistency rate,  such ATE estimate does not enjoy the asymptotic normality because of the bias. Therefore, we exploit the idea of sample splitting,  where the ATE estimate is constructed as the empirical mean of the estimated DNN regression function evaluated on an independent inference data set. The similar sample splitting idea has been popularly used in the literature; see, for example, \cite{double-machine2018}. We show that if the sample used for DNN training is much larger than the sample used for inference,  then the resulting ATE estimate enjoys the asymptotic normality, ensuring valid statistical inference. 

We then incorporate the idea of DNN modeling into the doubly robust ATE estimation \cite{doubly-robust2011,Fan2016ImprovingCB}. We show that with the DNN estimate of the mean regression function discussed above,  only very mild conditions on the propensity score estimation are needed for the doubly robust estimator to be consistent.  For the asymptotic normality, we resort to the same sample splitting idea. We show that equally split samples, together with some additional mild assumptions on the propensity score estimation,  can be sufficient for the doubly robust estimator to obtain asymptotic normality.  In particular, we prove that the propensity score estimate based on the same DNN architecture gives us one such estimate.

The remaining of the paper is organized as follows. In Section \ref{Sec2}, we introduce our model setting and two DNN-based ATE estimation methods. In Section \ref{Sec2.2}, we study the sampling properties of these two estimators including their asymptotic normality. Sections \ref{Sec.simu} and  \ref{Sec.reda} present numerical results using simulated examples and a real data example, respectively. Section \ref{Sec.disc} contains some conclusions and directions for future study.  All technical proofs are deferred to the Appendix and the Supplementary Material.

\subsection{Notation} \label{Sec1.1}

To facilitate the technical presentation, we first introduce some necessary notation that will be used throughout the paper. We use $\|\cdot\|$ to denote the Euclidean norm of vectors. $\mathbb R$ and $\mathbb N$ stand for the collections of real numbers and positive integers, respectively, and $\mathbb N_0 = \mathbb N\cup \{0\}$. For a real-valued multivariate function $f(\BS{x}):\mathbb R^p \to \mathbb R$, denote by $\frac{\partial^kf(\BS{x})}{\partial x_1^{\alpha_1}\partial x_2^{\alpha_2}\dots\partial x_p^{\alpha_p}}$ the partial derivative of function $f$ of order $k$ for nonnegative integers $\alpha_1,\dots,\alpha_p$ such that $\sum_{i=1}^p\alpha_i = k$. 
We use $\lceil x \rceil $ and $\lfloor x \rfloor$ to represent the smallest integer greater than or equal to $x$ and the largest integer less than or equal to $x$, respectively. Denote by $\mathcal N(\epsilon,\mathcal F,\|\cdot\|)$ the covering number of some function class $\mathcal F$ with metric $\|\cdot\|$ at scale $\epsilon > 0$; see, e.g., \cite{NonpR}. That is, for a metric space $(\mathcal G, \|\cdot\|)$ with $\mathcal F \subset \mathcal G$, we define 
\begin{equation} \label{neweq005}
\mathcal N(\epsilon,\mathcal F,\|\cdot\|) = \min  \{|M|:M\subset \mathcal F \subset \bigcup_{f\in M}B(f,\epsilon)\},
\end{equation}
where $B(f,\epsilon) = \{g\in \mathcal G: \|f-g\| < \epsilon\}$ represents a ball centered at $f$ with radius $\epsilon$ in the metric space, and $|\cdot|$ denotes the cardinality of a set. 
Let $\Vert f\Vert_\infty$ be the supremum norm of $f:\mathbb R^p \to \mathbb R$, that is, $\Vert f(\BS x) \Vert_\infty = \sup_{\BS x\in \mathbb R^p}|f(\BS x) |$; for any set $A \subset\mathbb R^p$, define $\Vert f(\BS x) \Vert_{\infty,A} = \sup_{\BS x\in A}|f(\BS x) |$.
\section{ATE inference using deep neural networks}
\label{Sec2}

\subsection{Model setting} \label{Sec2.1}

Consider the  potential outcomes framework of causal inference (see, e.g., 
\cite{sekhon2008neyman}), where a set of independent and identically distributed (i.i.d.) observations $\mathcal D$ are obtained. Here, for $i=1,\cdots, n_{\mathcal D}$ with $n_{\mathcal D} := |\mathcal D|$, the $i$th observation in $\mathcal D$ is denoted as $(\BS{X}_i, T_i, Y_i)$, where $\BS{X}_i=(X_{i1}, \cdots, X_{ip})^{\top}$ represents the vector of $p$ covariates, $T_i$ is the treatment indicator ($1$ for treated and $0$ for untreated), and $Y_i \in \mathbb R$ is the  scalar response. 
The observed response takes the form $Y_i = T_iY_i(1)+(1-T_i)Y_i(0)$, with the two potential outcomes $Y_i(1)$ and $Y_i(0)$ representing the outcomes with and without treatment, respectively. A common estimate of interest is the average treatment effect (ATE) defined as
\begin{equation} \label{neweq001}
\tau = \mathbb{E}[Y_i(1)-Y_i(0)].
\end{equation}
Note that the potential outcome $Y_i(t)$, $t\in\{0,1\}$,  is latent when the individual $i$ receives the opposite treatment $T_i=1-t$, making the ATE estimation and inference challenging.

We assume the following nonparametric regression model for the observed response $Y_i$ 
\begin{equation} \label{model}
Y_i = m(\BS{X}_i,T_i) + \varepsilon_i,
\end{equation}
where $m(\BS x, t) = \mathbb{E}[Y_i | \BS{X}_i=\BS x, T_i = t]$ is the underlying regression function, and
$\varepsilon_i$ is the model error with zero mean and finite variance and is independent of both $\BS{X}_i$ and $T_i$. Throughout we make the commonly used assumptions that  1) $T_i  \indep \{Y_i(0), Y_i(1)\}  |  \textbf{X}_i$ and 2) $0 < \mathbb{P}[T_i = 1 | \textbf{X}_i] < 1$ almost surely, where the former is commonly referred to as the \textit{unconfoundedness} assumption and the latter is called  \textit{overlap} assumption. The main focus of our paper is to develop statistical inference  method for ATE using the nonparametric tool of DNN regression with theoretical underpinning. 

We start by discussing the estimation of ATE,  which is usually the first step of statistical inference.  Under the above two assumptions of unconfoundedness and overlap, the right-hand side of \eqref{neweq001} can be further written as
\begin{align} \label{new.eq.L001}
    \tau = \mathbb E[m(\BS X_i, 1) - m(\BS X_i, 0)].
\end{align}
This suggests that we estimate the ATE using the empirical counterpart 
\begin{equation}
\label{neweq002}
\widehat{\mathbb E}_{\BS X}[\widehat m(\BS{X}, 1) -\widehat m(\BS{X}, 0)], 
\end{equation}
where $\widehat m(\BS{x}, t)$ is an empirical estimate of $m(\BS x, t)$ for $t=0,1$, and $\widehat{\mathbb E}_{\BS X}$ stands for the empirical mean with respect to $\BS X$. 

For the intuitive estimate in \eqref{neweq002} to work well, we need to construct accurate estimates $\hat m(\BS x, t)$ for $t=0, 1$.  
With its appealing approximation property, DNN regression is a natural method to use for achieving this goal. For ATE estimation, the empirical mean $\widehat{\mathbb E}_{\BS X}$ in \eqref{neweq002} can be constructed using the same data as those for learning $\hat m(\BS x, t)$. However, if the goal is statistical inference, we will need to resort to data splitting and use an independent set to calculate the empirical mean to make the estimation bias under control in establishing the asymptotic normality of our estimator. A similar idea has been advocated in the literature; see, for example,  \cite{double-machine2018}. We will formalize the above statements in subsequent sections.

In what follows, we will discuss two estimators: one is constructed using the exact intuition in \eqref{neweq002}, and the other one is the doubly robust estimate that also exploits information from the propensity score. 

\subsection{ATE inference based on DNN estimate} \label{sec:dnn}

In the multivariate regression, it is well-known that classical nonparametric method can suffer from the curse of dimensionality when dimensionality $p$ is not very small.
Fortunately, under certain network architectures of the DNN, one can learn a broad class of smooth functions accurately with the aid of modern optimization circumventing the curse of dimensionality; see, e.g., the recent work in \cite{AOS2019Bauer}.  In this paper, we will consider the same DNN network architecture described by the following  functional space $\mathcal H^{(l)}_{M,p^*,p,\alpha}$ for the construction of ATE estimator.

\begin{definition} \label{Fclass}
Given positive integers $p^*, p, M, K$ and positive constant $\alpha$,  for each $l\in \mathbb N$, the function space $\mathcal H^{(l)}_{M,p^*,p,\alpha}$ 
is defined recursively as
	\begin{eqnarray*}
		\mathcal H^{(l)}_{M,p^*,p,\alpha}  = &\Big\{ h: \mathbb R^{p+1} \to \mathbb R \, \big |\, h(\BS x) = \sum_{k=1}^{K} g_k(f_{1,k}(\BS{x}),f_{2,k}(\BS{x}),\dots,f_{p^*,k}(\BS{x})) \\
		&\text{for some }g_k \in \mathcal H^{(0)}_{M,p^*,p,\alpha} \text{ and } f_{j,k} \in \mathcal H^{(l-1)}_{M,p^*,p,\alpha}\Big\},
	\end{eqnarray*}
	where 
	\begin{eqnarray*}
		\mathcal H^{(0)}_{M,p^*,p,\alpha} = &\Big\{f: \mathbb R^{p+1} \to \mathbb R \,\big |\, f(\BS x) = \sum_{i=1}^M \mu_i\cdot \sigma(\sum_{j=1}^{4p^*} \lambda_{i,j}\cdot \sigma(\sum_{v=1}^{p+1} \theta_{i,j,v}\cdot x^{(v)} +\theta_{i,j,0})+\lambda_{i,0})\\ & +\mu_0
		 \text{ with } |\mu_i| \leq \alpha, |\lambda_{i,j}|\leq \alpha, \text{ and } |\theta_{i,j,v}|\leq \alpha\Big\},
	\end{eqnarray*}
	and $\sigma(\cdot)$, specified as the sigmoid function in our theoretical study, is the activation function. Here, $\mu_i, \lambda_{i,j}, \theta_{i,j,v}\in \mathbb R$ are weight coefficients, $x^{(v)}$ denotes the $v$th component of vector $\BS x$
	, and $\cdot$ means the regular scalar multiplication which is explicitly spelled out for the presentation clarity. 
\end{definition}

The architecture of the DNN described in Definition \ref{Fclass} has been investigated in \cite{AOS2019Bauer}, with an illustrative diagram given in Figure 1 therein. As can be seen from the definition, the DNN is a feedforward network defined recursively using the two-layer network in $\mathcal H^{(0)}_{M,p^*,p,\alpha}$. As a result, many of the hidden layers are sparsely connected. The parameters $p^*$, $K$, and $M$ are all tuning parameters that need to be selected by the practitioner. 

For a function $f(\BS x)$ and some positive value $y$,  define the truncation function
 \begin{equation} \label{deftrunc}
    	\text{trunc}(f(\BS x),y) = \left\{\begin{aligned} & f(\BS x), & \text{ if }|f(\BS x)| \leq y,\\
    	&y\cdot\text{sign}(f(\BS x)), & \text{ if }|f(\BS x)| > y,	\end{aligned}\right. 
	\end{equation}
where $\text{sign}(t)$ denotes the sign function that takes value $1$ if $t > 0$, value $-1$ if $t < 0$, and value $0$ if $t = 0$. Given an i.i.d. sample $\mathcal D$, define  
	\begin{equation} \label{ERM}
	\widetilde m_{\mathcal D}(\BS x,t) = \arg \min_{h \in \mathcal H^{(l)}} \frac{1}{|\mathcal D|} \sum_{i\in \mathcal D} \big(Y_i - h(\BS{X}_i,T_i)\big)^2
	\end{equation}
	as the optimal neural network in $\mathcal H^{(l)}:=\mathcal H^{(l)}_{M,p^*,p,\alpha}$ 
	that minimizes the squared loss. Hereafter, with an abuse of notation,  we use $i\in \mathcal D$ to represent  the corresponding data $(Y_i,\BS X_i, T_i) \in \mathcal D$. To increase the robustness of the DNN estimate, we truncate $\widetilde m_{\mathcal D}(\BS x,t)$  as
	\begin{equation} \label{neweq006}
m_{\mathcal D}(\BS x,t) = \text{trunc}(\widetilde m_{\mathcal D}(\BS x, t), C\log n_{\mathcal D}),
	\end{equation}
where $C$ is some large enough universal positive constant. 

With the estimate $m_{\mathcal D}(\BS x, t)$, we are halfway done with constructing the DNN estimate of $\tau$ based on the intuition in \eqref{neweq002}. It remains to specify the empirical mean $\hat{\mathbb{E}}_{\BS X}$ in \eqref{neweq002}. A natural estimate is to average over covariates $\BS X_i$ from the same learning data $\mathcal D$, that is, 
	\begin{equation} \label{neweq007}
	\widehat \tau_{\mathcal D} = \frac{1}{|\mathcal D|}\sum_{i \in \mathcal D} [m_{\mathcal D}(\BS X_i,1) - m_{\mathcal D}(\BS X_i,0)].
	\end{equation}
We will show that such an estimate is consistent in estimating  $\tau$. However, the consistency rate is not fast enough for $\widehat \tau_{\mathcal D}$ to achieve the asymptotic normality, hindering its ability for valid statistical inference. 

To overcome such difficulty, we resort to the method of unbalanced sample splitting, which allows us to separate the randomness in the approximation step from the randomness in the inference step. Specifically, we assume that the available data set $\mathcal D$ can be randomly split into two independent data sets, the learning set $\mathcal D_1$ and the inference set $\mathcal D_2$, with  $|\mathcal D_1| = n^\gamma$ and $\gamma >1$ some constant, and  $|\mathcal D_2| =n$. Here, without loss of generality
, we assume that $n^\gamma$ is an integer. 
The learning set $\mathcal D_1$ is used to compute the estimated nonparametric mean regression function $m_{\mathcal D_1}(\BS x, t)$ as define in \eqref{neweq006}.  Then the inference set is also included to calculate the final ATE estimate, i.e.,
    \begin{equation}\label{deftauhat}
        \widehat \tau(\mathcal D_1,\mathcal D_2) = \frac{1}{|\mathcal D_2|}\sum_{i\in \mathcal D_2} \widehat \tau_i(\mathcal D_1),
    \end{equation}
    with $\widehat \tau_i(\mathcal D_1) = m_{\mathcal D_1}(\BS X_i,1) - m_{\mathcal D_1}(\BS X_i,0)$. We will show in Section \ref{Sec2.2} that the estimator defined in \eqref{deftauhat} achieves the asymptotic normality. From our technical analysis,  we will also see that the unbalanced sample splitting plays a pivotal role in establishing the asymptotic normality.


\subsection{Doubly robust estimate}\label{sec:def-double-robust}
The DNN estimate of $\tau$ discussed in the previous section does not require the estimation of the propensity score function. 
This section explores a different type of estimator, the doubly robust estimator, for its robustness to the misspecification of either the mean regression function or the propensity score function. In addition, we will make it clear that the asymptotic normality of the doubly robust estimator can be achieved with equally split samples, making its practical implementation attractive. 

We consider  the same model as in (\ref{model}). 
Given a data set $\mathcal D$ of i.i.d. observations, one can use the same DNN method as discussed in Section \ref{sec:dnn}  to estimate the regression function, yielding an estimate $ m_{\mathcal D}(\BS x, t)$. We denote by $\widehat m_{\mathcal D}(\cdot) = (\hat m_{\mathcal D,1}(\cdot), \hat m_{\mathcal D,0}(\cdot))$ with $\hat m_{\mathcal D,t}(\BS x) =  m_{\mathcal D}(\BS x, t)$ for $t=0, 1$ for notational simplicity.
We denote the propensity score estimate as $\hat e_{\mathcal D}(\BS x)$ that may be estimated by existing methods such as matching and stratification. Note that so far, we have not imposed any specific assumptions on the estimation accuracy of the propensity score function.

For a given data point $(Y_i,\BS X_i, T_i)$, let us define
\begin{equation}\label{defphi}
    \phi_i(\hat e_{\mathcal D},\hat m_{\mathcal D}) = \frac{T_i}{\hat e_{\mathcal D}(\BS X_i)}(Y_i - \hat m_{\mathcal D,1}(\BS X_i)) + \hat m_{\mathcal D,1}(\BS X_i)
\end{equation}
and
\begin{equation}\label{defpsi}
    \psi_i(\hat e_{\mathcal D},\hat m_{\mathcal D}) = \frac{1 - T_i}{1 - \hat e_{\mathcal D}(\BS X_i)}(Y_i - \hat m_{\mathcal D,0}(\BS X_i)) + \hat m_{\mathcal D,0}(\BS X_i).
\end{equation}
Then the doubly robust estimator based on data in $\mathcal D$ can be constructed as 
\begin{equation}\label{defdrestimator}
    \hat \tau_{DR, \mathcal D}(\hat e_{\mathcal D},\hat m_{\mathcal D}) = \frac{1}{|\mathcal D|}\sum_{i\in \mathcal D}\big(\phi_i(\hat e_{\mathcal D},\hat m_{\mathcal D}) - \psi_i(\hat e_{\mathcal D},\hat m_{\mathcal D})\big).
\end{equation}
We further define the population counterpart of $\hat \tau_{DR, \mathcal D}(\hat e_{\mathcal D},\hat m_{\mathcal D})$ as
\begin{equation}\label{eq:def-mean}
    \tau(\hat e_{\mathcal D},\hat m_{\mathcal D}) = \mathbb E_{(\BS X, T, Y)}\big(\phi(\hat e_{\mathcal D},\hat m_{\mathcal D}) - \psi(\hat e_{\mathcal D},\hat m_{\mathcal D})\big),
\end{equation}
where $\phi(\hat e_{\mathcal D},\hat m_{\mathcal D}) = \frac{T}{\hat e_{\mathcal D}(\BS X)}(Y - \hat m_{\mathcal D,1}(\BS X)) + \hat m_{\mathcal D,1}(\BS X)$, $(\BS X,T, Y)$ represents an independent new observation from the same distribution as $(\BS X_1, T_1, Y_1)\in \mathcal D$, $\psi(\hat e_{\mathcal D},\hat m_{\mathcal D})$ is defined analogously, and the expectation in \eqref{eq:def-mean} is taken with respect to $(\BS X,T, Y)$.

As discussed in the previous section, the above estimate \eqref{defdrestimator} is consistent in estimating the ATE under some regularity conditions. However, the estimation bias renders the asymptotic normality invalid.  Next, we discuss the doubly robust estimator based on the idea of data splitting.  Suppose we randomly split the set of available observations into two equal sized sets $\mathcal D_1$ and $\mathcal D_2$. Using data in $\mathcal D_1$, we calculate the estimates $\hat m_{\mathcal D_1, t}$, $t=0, 1$, and $\hat e_{\mathcal D_1}$  the same way as specified at the beginning of this section. Then the doubly robust estimator is constructed similar to \eqref{defdrestimator} except that \eqref{defphi} and \eqref{defpsi} are evaluated on the data in $\mathcal D_2$; that is, 
    \begin{equation}\label{eq: double-robust-split}
            \hat \tau_{DR,\mathcal D_2}(\hat e_{\mathcal D_1},\hat m_{\mathcal D_1}) = \frac{1}{|\mathcal D_2|}\sum_{i\in \mathcal D_2}\big(\phi_i(\hat e_{\mathcal D_1},\hat m_{\mathcal D_1}) - \psi_i(\hat e_{\mathcal D_1},\hat m_{\mathcal D_1})\big).
    \end{equation}

\section{Asymptotic distributions of regular and doubly robust DNN estimators for ATE} \label{Sec2.2}


Note that $m(\BS x, t)$ takes a discrete covariate $T$ as an input, which greatly increases the theoretical challenges and makes the existing tools for studying the sampling properties of DNN inapplicable. For the purpose of motivating our technical analysis, let us temporarily assume that the propensity score $e(\BS{x}) = \mathbb P(T = 1|\BS{X}=\BS{x})$ is known. Note that 
\begin{align}\label{neweq004}
& \mathbb E[Y|\BS{X}=\BS{x}, e(\BS X)=t] = \mathbb E[m(\BS{X}, T)|\BS{X}=\BS{x}, e(\BS X)=t] \nonumber \\
&=m(\BS{x}, 1)\mathbb{P}(T=1|\BS{X} = \BS{x}, e(\BS X)=t) + m(\BS{x}, 0)\mathbb{P}(T=0|\BS{X} = \BS{x}, e(\BS X)=t)\nonumber\\
& = \big[m(\BS x, 1)t + m(\BS x, 0)(1-t)\big] \mathbbm{1}\{e(\BS{x})=t\},
\end{align}
where $\mathbbm 1\{\cdot\}$ stands for the indicator function. We extend the domain of the underlying regression function $m(\BS x,t)$ to $\mathbb{R}^p \times [0,1]$  and define the intermediate values as
\begin{equation}\label{neweq004a}
m(\BS x, t)  = m(\BS x, 1)t + m(\BS x, 0) (1-t) 
\end{equation}
for $t\in (0,1)$. It is seen that the extended function is infinitely differentiable with respect to $t$ in $(0,1)$, and still satisfies our regression model assumption \eqref{model} on the boundary when $t\in\{0,1\}$.
Observe that $m(\BS x,t)$ in \eqref{neweq004a} is a function defined on $\mathbb{R}^p\times [0,1]$, and can be roughly understood as the underlying nonparametric regression function with $Y$ the response, and $(\BS X^{\top},  e(\BS X))^{\top}$ the new covariate vector\footnote{Rigorously speaking, this mean regression function is only defined on $\{(\BS x, t): e(\BS x)=t\}$. Also, the overlap assumption prevents $ e(\BS X)$ from taking values 0 and 1. We temporarily ignore these constraints for the sake of motivating our technical analysis.}. The advantage of having $m(\BS x, t)$ in \eqref{neweq004a} is that it is smooth with respect to $t$, which will greatly facilitate us in developing new machine learning theory. 

For observational studies, the propensity score function information is typically unknown. As a consequence,  $m(\BS x, t)$ in \eqref{neweq004a} is not directly estimable in the whole range of $t\in[0,1]$. Nevertheless, we still use the formulation in \eqref{neweq004a} keeping in mind that we only have observations on the boundary of the domain for $t\in [0,1]$ (i.e., the observed $T_i$'s). Since our theory does not rely on the values of $m(\BS x, t)$ when $t\in (0,1)$, such treatment should not cause any problems in our technical analyses.

To set up the technical preparation, we briefly review the major definitions and notation from \cite{AOS2019Bauer} below.

\begin{definition} \label{defsmooth}
Given $s>0$ and $C>0$,	the $(s,C)$-smooth function class for functions of $p$ real variables with $s = q + r$, $q \in \mathbb N_0$, and $0 < r \leq 1$ is defined as 
	\begin{eqnarray*}
		\mathcal S_{s,C,p} = &\Big\{ m:  \mathbb R^{p+1} \to \mathbb R \big |\, \big|\frac{\partial^{q}m(\BS{y})}{\partial x_1^{\alpha_1}\partial x_2^{\alpha_2}\dots \partial x_{p+1}^{\alpha_{p+1}}} - \frac{\partial^{q}m(\BS{z})}{\partial x_1^{\alpha_1}\partial x_2^{\alpha_2}\dots \partial x_{p+1}^{\alpha_{p+1}}}\big | \leq C\|\BS{y}-\BS{z}\|^{r}	\\
		& \text{for any } \BS{y},\BS{z} \in \mathbb R^{p+1} \text{ and } \sum_{i=1}^{p+1} \alpha_i = q \text{ with } \alpha_i\in \mathbb N_0,\, i=1,\cdots, p+1\Big\}.
	\end{eqnarray*}
\end{definition}

In what follows, for the ease of presentation, we refer to $\mathcal S_{s,C}$ as the function class that includes all the $\mathcal S_{s,C,p}$ functions for all positive integers $p$. 
The smoothness restrictions on the function class are commonly exploited for deriving nontrivial results on the rates of convergence for nonparametric estimators. In particular, the $(s,C)$-smoothness condition in Definition \ref{defsmooth}  has been used to derive  the distribution-free rates of convergence for nonparametric regression estimators; see, e.g., Section 3.2 of \cite{NonpR}. 

Now we are ready to introduce a generalized function class with some additional specific structures. These specific structures are well suited for our study and will assist us in the theoretical derivations.  
Recall that to facilitate our theory, the domain of the regression function $m(\cdot, \cdot)$ is extended to $\mathbb{R}^{p+1}$, while the values outside of the original domain do not convey any practical meaning. As will be seen in Condition \ref{condRF} below, we assume that $m(\cdot, \cdot)$ belongs to the class of $(s,C)$-smooth generalized hierarchical interactive functions, which is formally defined as follows.  

\begin{definition} \label{GHImodel}
	The $(s,C)$-smooth generalized hierarchical interactive function class of order $p^*\in \mathbb N$ and level $l \in \mathbb N$ is defined recursively as
	\begin{eqnarray*}
		\mathcal M_{p^*,l}(\mathcal S_{s,C}) = &\big\{ m:  \mathbb R^{p+1} \to \mathbb R\, \big |\, m(\BS{x}) = \sum_{k=1}^K g_k(f_{1,k}(\BS{x}),f_{2,k}(\BS{x}),\dots,f_{p^*,k}(\BS{x})) \text{ with } g_k \in \mathcal S_{s,C}, \\
		&f_{i,k} \in \mathcal M_{p^*,l-1}(\mathcal S_{s,C}) \text{ for } i = 1,2,\dots,p^* \text{ and } k = 1,2,\dots,K\big\},
	\end{eqnarray*}
	where $K$ is some positive integer and $p+1$ is the dimensionality of the augmented covariate vector. 	When $l = 0$, $\mathcal M_{p^*,0}(\mathcal S_{s,C})$ is defined as 
	$$\big\{ m:  \mathbb R^{p+1} \to \mathbb R \,\big |\, m(\BS{x}) = f(\BS{a}_1^{\top}\BS{x},\BS{a}_2^{\top}\BS{x},\dots,\BS{a}_{p^*}^{\top}\BS{x}) \text{ with } f\in \mathcal S_{s,C}, \BS{a}_i \in \mathbb R^{p+1} \text{ for } i= 1,2,\dots,p^*\}.$$
\end{definition}

The class of functions in Definition \ref{GHImodel} above is rich enough to contain numerous commonly used function classes such as the additive models, interaction models, and projection pursuit models. As a result, the assumption that the underlying mean regression function $m(\cdot, \cdot)\in \mathcal M_{p^*,l}(\mathcal S_{s,C})$ allows for rich model structures including interactions between the treatment indicator and the covariates, and also the latent factor structure in covariates. We also note that such a hierarchical structure resembles that of DNNs, which entails the approximation capabilities of DNN estimates defined in \eqref{neweq006}. See also \cite{AOS2019Bauer} for some related discussions.

We are now ready to introduce the regularity conditions that are needed to facilitate our technical analysis. 

\begin{condition}~\label{condRF}
	\begin{itemize}
		\item[(i)] The covariate vector $\BS{X}$ has bounded support and response $Y$ has subGaussian distribution with $\mathbb E\exp(cY^2) < \infty$, where $c>0$ is some constant.
		\item[(ii)] The regression function $m(\BS{x},t)\in \mathcal M_{p^*,l}(\mathcal S_{s,C})$ for some $s > 0$ and $C>0$. By the definition of $\mathcal M_{p^*,l}(\mathcal S_{s,C})$, all partial derivatives of order no larger than $q$ of  functions $g_k$ and $f_{i,k}$ are bounded by some universal positive constant in magnitude, and all the functions $g_k$ are Lipschitz continuous with Lipschitz constant $L> 0$.
		\item[(iii)] For $\mathcal H^{(l)}_{M,p^*,p,\alpha}$, 
		the parameters are taken as  $M=\left\lceil c_1n^{\frac{p^*}{2s+p^*}}\right\rceil$ and $\alpha = n^{c_2}$ for sufficiently large positive constants $c_1$ and $c_2$; the parameters $K$ and $p^*$ in defining $\mathcal H^{(l)}_{M,p^*,p,\alpha}$ are taken the same as in defining $\mathcal M_{p^*,l}(\mathcal S_{s,C})$ in part (ii) above.
		\item[(iv)] 
		There exists some constant $\delta >0$ such that $e(\BS{X})\in [\delta, 1-\delta]$ almost surely. 
	\end{itemize}
\end{condition}

The boundedness of the support of the covariate distribution is commonly assumed in nonparametric regression and helps bound the complexity of the DNN function class.
The slightly stronger assumption on overlap in Condition \ref{condRF}(iv) helps simplify the technical analysis. We also note that the parameters $K$ and $p^*$ in constructing the network $\mathcal H^{(l)}_{M,p^*,p,\alpha}$ should be correctly specified and thus equal to the ones in $\mathcal M_{p^*,l}(\mathcal S_{s,C})$, and this assumption is inherited from \cite{AOS2019Bauer}. Establishing the theory when $K$ or $p^*$ is misspecified in constructing the DNN is highly challenging and left for future investigation.  

\subsection{Asymptotic normality of the regular DNN estimator} \label{Sec2.2.1}
We start with presenting the consistency of the DNN estimator without data splitting defined in \eqref{neweq007}.   
\begin{proposition} \label{mainthm}
	Assume that Condition \ref{condRF} with the sigmoid activation function $\sigma(x) = \frac{e^x}{e^x+1}$ in $\mathcal H^{(l)}$ holds. 
	Then the estimator $\widehat \tau_{\mathcal D}$ defined in  \eqref{neweq007} satisfies that
$|\widehat\tau_{\mathcal D} -\tau| = o_P\{(\log n_{\mathcal D})^{2} n_{\mathcal D}^{-\frac{s}{2s+p^*}}\}$ as $n_{\mathcal D} = |\mathcal D| \rightarrow \infty$.
\end{proposition}

The proof of Proposition \ref{mainthm} uses some key results established in \cite{NonpR}. Thanks to the specific DNN network architecture in Definition \ref{Fclass}, the rate of convergence in Proposition \ref{mainthm} above is free of dimensionality $p$. The intuition is that the underlying regression function $m(\BS x, t)$ has the sparsity structure  specified in $\mathcal M_{p^*,l}(\mathcal S_{s,C})$, whose complexity is controlled by $p^*$.  Thus, the dimension-free convergence rate is attainable.

We now present the asymptotic normality of the data splitting estimator \eqref{deftauhat}. Recall that after splitting, we have data sets of sizes $|\mathcal D_1| = n^{\gamma}$ and $|\mathcal D_2|=n$. To gain some high-level understanding, consider the decomposition
    \begin{equation}\label{eq:asy-norm-decomp}
        \sqrt{n} (\widehat \tau(\mathcal D_1,\mathcal D_2) -\tau) = \frac{1}{\sqrt{n}}\sum_{i\in \mathcal D_2} (\tau_i - \tau) + \frac{1}{\sqrt{n}} \sum_{i\in \mathcal D_2} (\hat\tau_i(\mathcal D_1) - \tau_i),
    \end{equation}
    where $\tau_i =m(\BS X_i, 1) - m(\BS X_i, 0)$ and $\hat\tau_i(\mathcal D_1)$ is defined in Section \ref{sec:dnn}. Note that the first term on the right-hand side of (\ref{eq:asy-norm-decomp}) is the scaled summation of i.i.d. mean zero random variables and thus is asymptotically normal. For the second term on the right-hand side of (\ref{eq:asy-norm-decomp}), since the proof of Proposition \ref{mainthm} shows that $m_{\mathcal D_1}(\BS x, t)$ is consistent in estimating $m(\BS x, t)$, it follows that the second term is negligible when the sample size of $\mathcal D_1$ is much larger than that of $\mathcal D_2$. These results are formally presented in Theorem \ref{Asynormality} below.   
\begin{theorem}\label{Asynormality}
    Assume that the conditions of Proposition \ref{mainthm} hold and $\gamma >1 + \frac{p^*}{2s}$. Then we have
    \begin{equation}
        \sqrt{n}(\widehat \tau(\mathcal D_1,\mathcal D_2) - \tau) \toD N(0,\sigma^2)
    \end{equation}
     as $n\to \infty$, where $\sigma^2 = \Var(m(\BS X,1) - m(\BS X,0))$. 
 \end{theorem}

The requirement of $\gamma > 1 + \frac{p^*}{2s} $ in Theorem \ref{Asynormality} above can be relaxed if the regression function $m(\BS x, t)$ takes a more specific form,  as formally presented in the condition of the corollary below.  

\begin{condition}~\label{condpoly}
\begin{itemize}
    \item[(i)]  
    The regression function $m(\BS x,t)\in 	\mathcal M_{p^*,l}(\mathcal S_{s,C})$, where all functions $g_k$ and $f_{i,k}$ with $k= 1,\dots,K$ and $i = 1,\dots,p^*$ appearing in the definition of $\mathcal M_{p^*,l}(\mathcal S_{s,C})$ are polynomials taking the following generic form
    \begin{equation*}
    f(\BS x) = \sum_{|\bm{\alpha} | \leq q} r_{\bm{\alpha}}{\BS x}^{\bm{\alpha}}
    \end{equation*}
    with some $q \in \mathbb N_0$, $r_{\bm{\alpha}}\in \mathbb R$ the regression coefficient, $\BS x = (x_1,\cdots, x_{p+1})^{\top}$, $\bm{\alpha} = (\alpha_1,\alpha_2,\dots,\alpha_{p+1})$, and $\BS x^{\bm{\alpha}} = x_1^{\alpha_1}\dots\dots x_{p+1}^{\alpha_{p+1}}$. Here, assume that $\alpha_i \in \mathbb N_0$ and $|\bm{\alpha}| = \sum_{i=1}^{p+1} \alpha_i$.
    \item[(ii)] Denote by $q_0$ the highest order of all the polynomials in part (i). Take the parameters in $\mathcal H_{M,p^*,p,\alpha}^{(l)}$ as  $M = \left\lceil c_1n^{\frac{p^*}{2\lambda_n+p^*}} \right\rceil$ and $\alpha = n^{c_2}$ for sufficiently large positive constants $c_1$ and $c_2$, where the non-decreasing sequence $\{\lambda_n\}$ is defined as 
      $$\lambda_n = \inf\{s\in \mathbb N: n(\lambda) \geq n + 1\}$$
      with
    $$
    n(\lambda) = \inf\Bigl\{n\in \mathbb N: \frac{\log(n)}{2s+p^*} \geq \log(\frac{3}{2(2q_0+3)}) + \log(\log n)\Bigr\}.
    $$
    In addition, parameters $K$ and $p^*$ in defining $\mathcal H_{M,p^*,p,\alpha}^{(l)}$ and $\mathcal M_{p^*,l}(\mathcal S_{s,C})$ in part (i) are the same.
\end{itemize}
\end{condition}
The parameter $s$ in Condition \ref{condpoly}(i) above can take some arbitrary positive value in $\mathbb R$, in view of the specific form of functions involved in the definition. The constants $c_1$ and $c_2$ in Condition \ref{condpoly}(ii) are generally different from the corresponding constants in Condition \ref{condRF}, because the former ones depend generally on $p^*$, $p$, and $q_0$, while the latter ones can depend on parameter $p^*$, $p$, and $s$ in Condition \ref{condRF}. 

\begin{corollary} \label{cor1}
Assume that (i) and (iv) of Condition \ref{condRF} and Condition \ref{condpoly} hold with the sigmoid activation function $\sigma(x) = \frac{e^x}{e^x + 1}$ in $\mathcal H^{(l)}$.
Let $|\mathcal D_1| = n(\log n)^k$ 
with $|\mathcal D_2| = n$ for some $k > 4+ p^*$. 
Then for $\widehat \tau(\mathcal D_1,\mathcal D_2)$ defined in \eqref{deftauhat}, we have
\begin{equation}
        \sqrt{n}(\widehat \tau(\mathcal D_1,\mathcal D_2) - \tau) \toD  N(0,\sigma^2)
\end{equation}
as $n\to \infty$, where $\sigma^2$ is as defined in Theorem \ref{Asynormality}.
\end{corollary}

 Rigorously speaking, Corollary \ref{cor1} cannot be proved by directly applying Proposition \ref{mainthm} or Theorem \ref{Asynormality}. The main difficulty is that although a regression function $m(\BS x)$ satisfying Condition \ref{condpoly} belongs to $\mathcal M_{p^*,l}(\mathcal S_{s,C})$ over all $s\in \mathbb N$, the probabilistic statements  in proving Proposition \ref{mainthm} and Theorem \ref{Asynormality} do not hold uniformly for all $s\in \mathbb N$. Thus, we cannot simply set $s$ to infinity to prove Corollary \ref{cor1}. Instead, we must first establish results similar to those in \cite{AOS2019Bauer} in order to prove Corollary \ref{cor1}. Nevertheless, since the function class in Corollary \ref{cor1} is much smaller, we downgrade the importance of the result and name it a corollary.  Whether a similar result holds for a broader class of analytic functions that are  infinitely differentiable is an interesting question for future study.  
 
 Compared to Theorem \ref{Asynormality},  the weaker assumption in Corollary \ref{cor1} on $\gamma_n$ indicates that the asymptotic normality is possible with nearly balanced sample splitting.
 The fundamental reason is that, by  modifying the proof of Theorem 1 in \cite{AOS2019Bauer} to require a stronger structural assumption on the mean regression function $m(\BS x, t)$, we can show that the DNN regression function achieves a near $n^{-1/2}$ convergence rate (up to some logarithmic factor). 

For the asymptotic normality in Theorem \ref{Asynormality} and Corollary \ref{cor1} to be practically applicable, we need an accurate variance estimate. Let us consider the following natural choice
\begin{equation}\label{defsamplevar}
         \hat \sigma^2(\mathcal D_1, \mathcal D_2) = \frac{n}{n-1}\Big(\frac{1}{n}\sum_{i\in \mathcal D_2}{\widehat \tau_i^2(\mathcal D_1)} - \big(\frac{1}{n} \sum_{i\in \mathcal D_2} \widehat \tau_i(\mathcal D_1)\big)^2\Big).
\end{equation}
The independence between the data in $\mathcal D_1$ and $\mathcal D_2$ and the consistency of $m_{\mathcal D_1}(\BS x, t)$ (cf. the proof of Proposition \ref{mainthm}) ensure that $\hat \sigma^2(\mathcal D_1,\mathcal D_2)$ introduced in (\ref{defsamplevar}) is a consistent estimator of $\sigma^2$, yielding the following asymptotic normality with the estimated variance. 

\begin{theorem}\label{varestimate}
   Under the conditions of Theorem \ref{Asynormality}, we have the asymptotic normality using the variance estimator defined in \eqref{defsamplevar}
    \begin{equation}\label{asymptoticnormalequ}
        \frac{\sqrt{n}(\widehat \tau(\mathcal D_1,\mathcal D_2) - \tau)}{\hat \sigma(\mathcal D_1, \mathcal D_2)}\toD  N(0,1)
    \end{equation}
    as $n\to \infty$.
    Moreover, it holds that 
    \begin{equation}
        |\hat\sigma^2(\mathcal D_1, \mathcal D_2) - \sigma^2| = o_P((\log n)^4 n^{-1/2})
    \end{equation}
    for large enough $n$.
\end{theorem}

Theorem \ref{varestimate} above makes the practical construction of confidence intervals (CIs) possible when sample size $n$ is large. In particular, a level $100(1-\alpha)\%$ CI for $\tau$ is given by 
\begin{equation} \label{new.eq.L002}
(\widehat \tau(\mathcal D_1,\mathcal D_2)-n^{-1/2}\hat \sigma(\mathcal D_1, \mathcal D_2)z_{\alpha/2}, \widehat \tau(\mathcal D_1,\mathcal D_2)+n^{-1/2}\hat \sigma(\mathcal D_1, \mathcal D_2)z_{\alpha/2}),
\end{equation}
where $z_{\alpha/2}$ is the $100(1-\alpha/2)$th percentile of the standard normal distribution. 
Corollary \ref{cor2} below  summarizes the results that are parallel to those in Corollary \ref{cor1}.

\begin{corollary}\label{cor2}
 Under the conditions of Corollary \ref{cor1}, the asymptotic normality in \eqref{asymptoticnormalequ} holds. In addition, we have $
    |\hat\sigma^2(\mathcal D_1,\mathcal D_2) - \sigma^2| = O_P((\log(n))^4 n^{-1/2}).$
\end{corollary}

\subsection{Asymptotic normality of the doubly robust DNN estimator}
Recall that we use the balanced sample splitting in constructing the doubly robust estimator. We slightly abuse the notation and use $n$ to denote the common sample size for both $\mathcal D_1$ and $\mathcal D_2$ in this section.    
We require the following condition on the propensity score estimation for investigating the sampling properties of the doubly robust estimator.  
\begin{condition}~\label{condPS}
\begin{itemize}
    \item[(i)] There exists some constant $C_2>0$ such that for any $n$, the propensity score estimate $\hat e_{\mathcal D_1}(\BS x)$ constructed from sample $\mathcal D_1$ satisfies that 
    \begin{equation}
        \frac{1}{C_2\log n} \leq \hat e_{\mathcal D_1}(\BS X) \leq 1 - \frac{1}{C_2\log n},
    \end{equation}
    for $\BS X$ almost surely, where $\BS X$ is an independent observation from the same distribution as $\BS X_1$.
    \item[(ii)] It holds that
    \begin{equation}
        \mathbb E\left\{\frac{1}{n}\sum_{i\in \mathcal D_1}|\hat e_{\mathcal D_1}(\BS X_i) - e(\BS X_i) |^2\right\} = o(\frac{1}{\log^2 n}).
    \end{equation}
    \item[(iii)] Assume that
    \begin{equation}
        \mathbb E_{\mathcal D_1}\mathbb E_{\BS X}|\hat e_{\mathcal D_1}(\BS X) - e(\BS X)|^2  = o(n^{-1/2}),
    \end{equation}
    where $\BS X$ is an independent observation from the same distribution as $\BS X_1$. 
\end{itemize}
\end{condition}

Condition \ref{condPS}(i) can be easily satisfied if we define a truncated propensity score estimator; see \eqref{defpsestimation1} below for an example. Condition \ref{condPS}(ii) is a mild consistency assumption on $\widehat e_{\mathcal D_1}$. Condition \ref{condPS}(iii) plays a crucial role in establishing the asymptotic normality of the doubly robust estimator based on sample splitting. We will suggest a propensity score estimator that satisfies all these conditions toward the end of this section. 

\begin{proposition}\label{DRestimation}
    Assume that the conditions of Proposition \ref{mainthm} hold and the propensity score estimator $\hat e_{\mathcal D_1}$ satisfies (i) and (ii) of Condition \ref{condPS}. Then the doubly robust estimator defined in (\ref{defdrestimator}) satisfies that 
    \begin{equation}
        |\hat\tau_{DR, \mathcal D_1}(\hat e_{\mathcal D_1},\hat m_{\mathcal D_1}) -\tau |  = o_P(1).
    \end{equation}
\end{proposition}


Proposition \ref{DRestimation} does not give us an explicit convergence rate because of the very weak assumptions on the propensity score estimator $\hat e_{\mathcal D_1}$. The explicit rate can be derived at the cost of assuming the faster convergence rate for $\hat e_{\mathcal D_1}$ in Condition \ref{condPS}(iii).

\begin{theorem}\label{DRasynormal}
   Assume that the conditions of Proposition \ref{mainthm} hold with $p^* < 2s$. Then for any propensity score estimator $\hat e_{\mathcal D_1}$ satisfying (i) and (iii) of Condition \ref{condPS}, the doubly robust ATE estimator based on the sample splitting defined in \eqref{eq: double-robust-split} has the asymptotic normality
    \begin{equation}\label{DRasymptotic}
        \sqrt{n}(\hat\tau_{DR, \mathcal D_2}(\hat e_{\mathcal D_1},\hat m_{\mathcal D_1}) - \tau) \toD  N(0,\sigma_{DR}^2)
    \end{equation}
    as $n \to \infty$, where $\sigma_{DR}^2 = \Var(m_1(\BS X) - m_0(\BS X)) + \Var(\varepsilon)\mathbb E\frac{1}{e(\BS X)(1-e(\BS X))}$.
\end{theorem}

Comparing Theorem \ref{DRasynormal} with Theorem \ref{Asynormality}, the doubly robust estimator has larger asymptotic variance than the regular DNN estimator $\hat\tau(\mathcal D_1, \mathcal D_2)$. This is reflected in the results of our simulation studies. Similar to the DNN estimate presented in the previous section, the asymptotic variance $\sigma_{DR}^2$ can be estimated  using a plug-in estimator 
\begin{equation}\label{defsampleDRvariance}
    \hat\sigma_{DR,\mathcal D_2}^2(\hat e_{\mathcal D_1},\hat m_{\mathcal D_1}) = \frac{n}{n-1}\Big(\frac{1}{n}\sum_{i \in \mathcal D_2}\hat \tau_i^2(\hat e_{\mathcal D_1},\hat m_{\mathcal D_1}) - \big(\frac{1}{n}\sum_{i \in \mathcal D_2} \hat \tau_i(\hat e_{\mathcal D_1},\hat m_{\mathcal D_1})\big)^2\Big),
\end{equation}
where $\hat \tau_i(\hat e_{\mathcal D_1},\hat m_{\mathcal D_1}) = \phi_i(\hat e_{\mathcal D_1},\hat m_{\mathcal D_1}) - \psi_i(\hat e_{\mathcal D_1},\hat m_{\mathcal D_1})$ and all the notation is the same as in Section \ref{sec:def-double-robust}. 
\begin{theorem}\label{DRvarconsistency}
  Under the conditions of Theorem \ref{DRasynormal}, it holds that
    \begin{equation}
        \frac{\sqrt{n}(\hat \tau_{DR,\mathcal D_2}(\hat e_{\mathcal D_1},\hat m_{\mathcal D_1}) - \tau) }{\hat\sigma_{DR,\mathcal D_2}(\hat e_{\mathcal D_1},\hat m_{\mathcal D_1})}\toD N(0,1)
    \end{equation}
    as $n\rightarrow\infty$. In addition, we have 
    \begin{equation}
        |\hat \sigma_{DR,\mathcal D_2}^2(\hat e_{\mathcal D_1},\hat m_{\mathcal D_1}) - \sigma_{DR}^2|  = o_P((\log n)^2 n^{-1/4}).
    \end{equation}
\end{theorem}

Next we consider a specific propensity score estimator that satisfies the conditions of Theorem \ref{DRasynormal}. We start with introducing the condition below which restricts the structure of the true propensity score.   

\begin{condition}~\label{condpropensity}
    The propensity score $e(\BS x)\in \mathcal M_{p^*,l}(\mathcal S_{s_e,C_e})$ for some constants $s_e = q_e + r_e > 0$ with $q_e \in \mathbb N_0$ and $0<r_e \leq 1$, and $C_e > 0$. Moreover, all partial derivatives of order no larger than $q_e$ of  functions $g_k$ and $f_{i,k}$ involved in the definition of $\mathcal M_{p^*,l}(\mathcal S_{s_e,C_e})$ are bounded by some universal positive constant in magnitude, and all functions $g_k$ are Lipschitz continuous with Lipschitz constant $L_e > 0$.
\end{condition}

Observe that the above condition on the propensity score resembles Condition \ref{condRF}(ii) for $m(\BS x, t)$ except that the ambient dimensionality is $p$ instead here. The smoothness parameters in these two conditions can be different and one may use $\min\{s,s_e\}$ to unify them. 
Condition \ref{condpropensity} above  accommodates commonly used propensity score functions such as the logistic function of form $e(\BS x) = \frac{\exp(\BS{a}^T\BS x)}{1 + \exp(\BS{a}^T\BS x)}$ with $\BS a \in \mathbb R^{p}$ the regression coefficient vector. It is seen that $e(\BS x) = f(\BS a^T \BS x)$ for $f(x) = \frac{\exp(x)}{1+\exp(x)}$. Thus, the propensity score function belongs to the function class $\mathcal M_{1,0}(\mathcal S_{s,C_e})$ for any positive $s\geq 1$ and some $C_e$ depending on $s$\footnote{This can be verified by Fa{\`a} di Bruno's formula for high order derivatives of the composite function $f(x) = f_1\circ f_2 $, where $f_1(x) = \frac{x}{1+x}$ and $f_2(x) = e^x$.}. For example, by letting $s = 1$, we see that $f(x) \in \mathcal S_{1,\frac{1}{4},1}$ (see Definition \ref{defsmooth}) and the condition of $1 = p^*<2s = 2$ in Theorem \ref{DRasynormal} holds.

We next introduce the DNN estimate for the propensity score. Let us define 
   \begin{equation}\label{defpsestimation1}
       \hat e_{\mathcal D_1}(\BS x) = \frac{ 1}{2} + \text{trunc}\Big(\widetilde e_{\mathcal D_1}(\BS x,t) - \frac{ 1}{2},\frac{ 1}{2} - \frac{1}{C_2\log(n)}\Big), 
   \end{equation}
   where $C_2>0$ is some constant and
   \begin{equation}\label{defpsestimation2}
       \widetilde e_{\mathcal D_1}(\BS x) = \arg\min_{h \in \mathcal H_{M,p^*,p-1,\alpha}^{(l)}}\frac{1}{n}\sum_{i\in\mathcal D_1}|T_i  - h(\BS X_i)|^2.
   \end{equation}
We make the same assumption that parameters $K$ and $p^*$ in $\mathcal H_{M,p^*,p-1,\alpha}^{(l)}$ above are set at their true values in Condition \ref{condpropensity} for defining $\mathcal M_{p^*,l}(\mathcal S_{s_e,C_e})$.

\begin{corollary}\label{PSconditioncor}
Assume that the conditions of Proposition \ref{mainthm} hold with $p^* < 2s_e$ and Condition \ref{condpropensity} holds. Then the propensity score estimator $\hat e_{\mathcal D_1}(\cdot)$  defined in (\ref{defpsestimation1}) satisfies  (i) and (iii) of Condition \ref{condPS}. Consequently, the resulting doubly robust estimator enjoys the same asymptotic normality as in Theorems \ref{DRasynormal} and \ref{DRvarconsistency}.
\end{corollary}

\section{Simulation studies} \label{Sec.simu}

In this section, we consider simulation examples mimicking observational data to verify the theoretical results obtained in Section \ref{Sec2.2} for the two ATE estimates and illustrate their finite-sample performance.

\subsection{Simulation results of the mean difference DNN estimator for ATE} \label{Sec.simu1}

Consider the following main effect for the control group $T = 0$ 
\begin{equation} \label{new.eq.FL001}
m_0(\mathbf{x})=\mathbb{E} (Y_i(0) | \bx) = x_{1}^{2} + x_{2} + x_{3}^{2},  
\end{equation}
where we choose $\mathbf{x} = (X_1, \cdots, X_p)^\top \sim \text{Uniform}([0,1])^{p}$. The treatment propensity score $\mathbb{P}(T = 1 | \bx)$ is defined as 
\begin{equation} \label{new.eq.FL002}
e(\mathbf{x}) = \frac{1}{4}(1 + \beta_{2,4}(x_{3})),
\end{equation}
where $\beta_{2,4}$ denotes the beta distribution with shape parameters $2$ and $4$. Finally, the treatment effect is kept fixed at $\tau(\bx) = 1$ and we assume an additive model error of $\varepsilon \sim N(0,1)$.

A similar simulation setting was first proposed in \cite{wager2018a} with a linear main effect function. We use a slightly more complicated main effect function, but our goal is the same as in \cite{wager2018a}. Specifically, we intend to test the ability of our estimator to correct for bias due to an interaction between the propensity score and the main effect. This simulation setting mirrors the challenge in observational studies in which the treatment assignment is correlated with the potential outcomes. Thus, the statistical method must accurately adjust for the observed covariates to avoid a biased estimate.

We generate a data set of size $n_{\mathcal D}$ from the above observational data model in (\ref{new.eq.FL001})--(\ref{new.eq.FL002}) and we set $p=50$. Then we randomly split the data into two parts: a training sample $\mathcal D_1$ of size $n_1 = c n$ and an inference sample $\mathcal D_2$ of size $n$, where $n = 1000$ and we consider the choices of $c = 1, \cdots, 5$. For each generated data set, we apply a deep neural network (DNN) model with the feedforward network structure to the training sample. More specifically, we employ a DNN with three hidden layers, where the number of neurons in each hidden layer is set as $p + 1$ since we include the treatment assignment as an input into our network. Furthermore, we set the learning rate and batch size as $0.001$ and $128$, respectively, and allow the number of epochs to vary from $100$ to $800$. We optimize the network parameters using the Adam optimizer. Finally, we consider the two popular choices of the sigmoid activation and the ReLU activation for the activation function.  

We begin with the imbalanced samples version of the ATE estimate with DNN defined in Section \ref{sec:dnn}. A joint nonparametric regression function $\hat{m}_{\mathcal D_1}(\bx, t)$ can be constructed based on the training sample $\mathcal D_1$. Then we can construct the regular DNN ATE estimator using the inference sample $\mathcal D_2$. The simulation example is repeated $200$ times to generate the distribution of the resulting regular DNN ATE estimator.

\begin{figure}[ht]
    \centering
    \includegraphics[width=\linewidth]{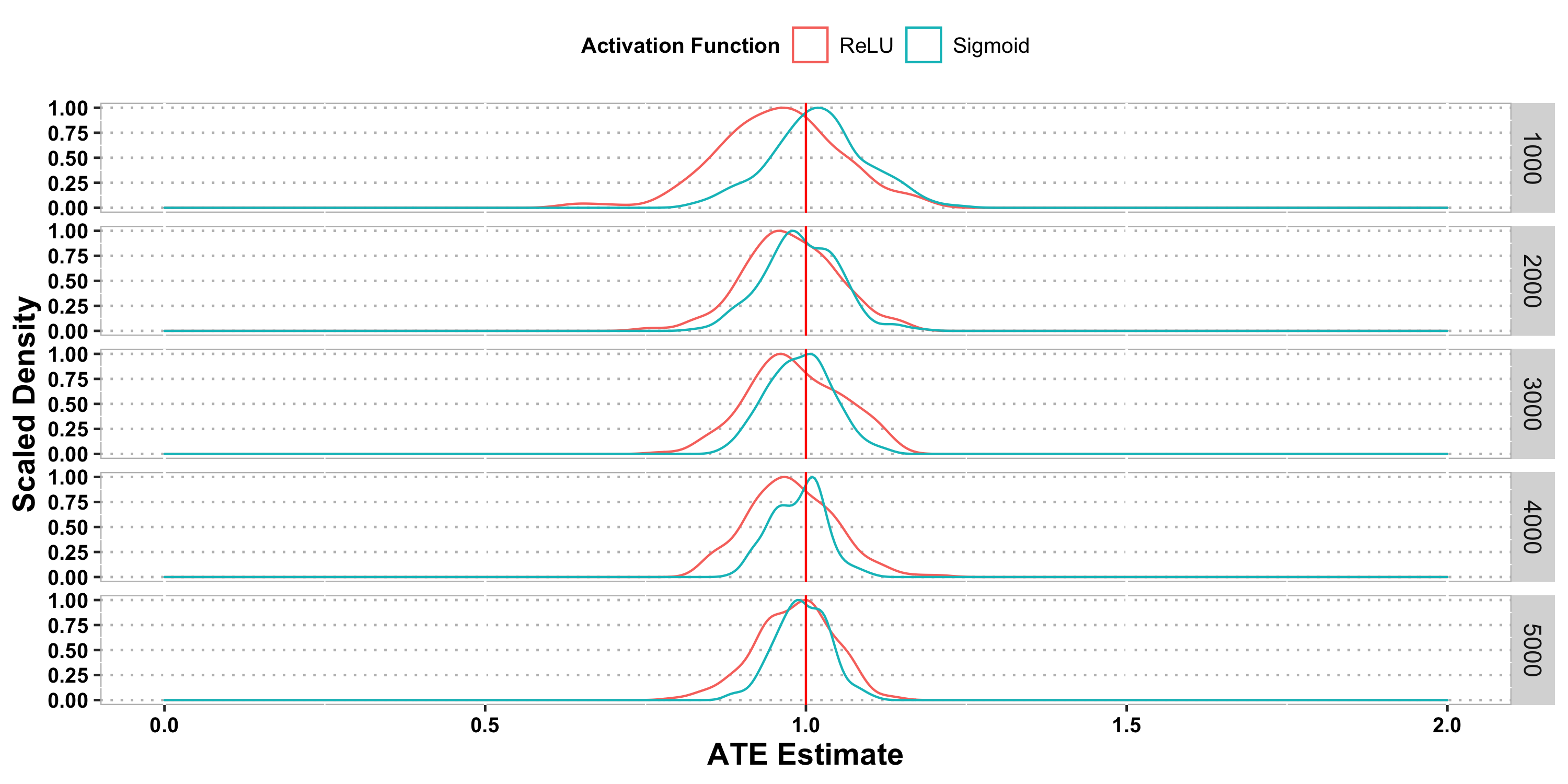}
    \caption{The scaled density of the ATE estimate over 200 replications for different training sample sizes and different activation functions. Here we use a fixed inference sample size of $n = 1000$ and train each network for 800 epochs. From top to bottom, the training sample size $n_1$ increases from 1000 to 5000. The true treatment effect of $\tau = 1$ is shown as a red vertical line. Results for different training lengths can be found in Section \ref{Sec.C} of the Supplementary Material.}
    \label{fig:sim_1_overlap_800_epochs}
\end{figure}

Figure~\ref{fig:sim_1_overlap_800_epochs} and Table~\ref{table:sim_1_800_epochs} present the results of the imbalanced samples version of the ATE estimate with DNN as a function of the choice of activation (i.e., sigmoid vs. ReLU), the training-to-inference ratio $c$, and the number of epochs varying from $100$ to $800$. 


\begin{table}[htp]
    \centering
    
\begin{tabular}[t]{cccccc}
\toprule
$n_1$ & Activation & Mean & Median & SD & MSE\\
\midrule
 & ReLU & 0.9567 & 0.9592 & 0.09532 & 0.01091\\

\multirow{-2}{*}{\centering\arraybackslash 1000} & Sigmoid & 1.0196 & 1.0175 & 0.07522 & 0.00601\\
\cmidrule{1-6}
 & ReLU & 0.9760 & 0.9703 & 0.07188 & 0.00572\\

\multirow{-2}{*}{\centering\arraybackslash 2000} & Sigmoid & 0.9927 & 0.9864 & 0.05797 & 0.00340\\
\cmidrule{1-6}
 & ReLU & 0.9837 & 0.9776 & 0.07215 & 0.00544\\

\multirow{-2}{*}{\centering\arraybackslash 3000} & Sigmoid & 0.9911 & 0.9899 & 0.04983 & 0.00255\\
\cmidrule{1-6}
 & ReLU & 0.9769 & 0.9740 & 0.06647 & 0.00493\\

\multirow{-2}{*}{\centering\arraybackslash 4000} & Sigmoid & 0.9881 & 0.9926 & 0.04098 & 0.00181\\
\cmidrule{1-6}
 & ReLU & 0.9821 & 0.9866 & 0.06029 & 0.00394\\

\multirow{-2}{*}{\centering\arraybackslash 5000} & Sigmoid & 0.9941 & 0.9931 & 0.04124 & 0.00173\\
\bottomrule
\end{tabular}

    \caption{Results of the same simulation setting as in Figure \ref{fig:sim_1_overlap_800_epochs} aggregated over 200 replications. In each replication, the networks are trained for 800 epochs. Results for different training lengths can be found in Section \ref{Sec.C} of the Supplementary Material. \vspace{0.1in}}
    \label{table:sim_1_800_epochs}
\end{table}



From Figure~\ref{fig:sim_1_overlap_800_epochs} and Table~\ref{table:sim_1_800_epochs}, we see that sigmoid activation generally outperforms ReLU activation in terms of the bias and variance. Indeed, out technical assumptions exclude ReLU because of its nonsmoothness.   
Developing theory for ReLU is an interesting research topic for future study. The empirical distribution of the ATE estimator is rather close to the normal distribution that is nearly centered around the true value of the ATE $\tau$. Furthermore, we observe that the results improve as the training-to-inference ratio $c$ increases, which is consistent with our theory. We also observe that the performance of the ATE estimator becomes better as the number of epochs grows. However, since the risk of overfitting also increases when the number of epochs is too large, we recommend to cap it to prevent overfitting of the DNN model. We present additional simulation results with different numbers of epochs in Section \ref{Sec.C} of the Supplementary Material.

\subsection{Simulation results of the doubly robust DNN estimator for ATE} \label{Sec.simu2}

We now turn to the doubly robust version of the ATE estimate with DNN as defined in Section \ref{sec:def-double-robust}. The simulation setting is the same as in Section \ref{Sec.simu1}. A key difference is that in addition to constructing an estimated regression function $\hat{m}_{\mathcal D_1}(\bx, t)$ based on the training sample $\mathcal D_1$, we will also construct the estimated propensity score $\hat{e}_{\mathcal D_1}(\BS x)$ based on the same training sample $\mathcal D_1$. Then using the inference sample $\mathcal D_2$, we can construct the doubly robust DNN ATE estimator as given in \eqref{eq: double-robust-split}. For the construction of the estimated propensity score with DNN, we can always fix a relatively small number of epochs for the training of the network (e.g., at $100$ across all the settings) and at the same time, constrain the estimated propensity score within $[0.01, 1 - 0.01]$. The main purpose of these modifications is to prevent the over- or perfect fitting of the propensity score. Moreover, we will vary the number of epochs for the construction of  $\hat{m}_{\mathcal D_1}(\bx, t)$ with DNN as in Section \ref{Sec.simu1}. 

\begin{figure}[ht]
    \centering
    \includegraphics[width=\linewidth]{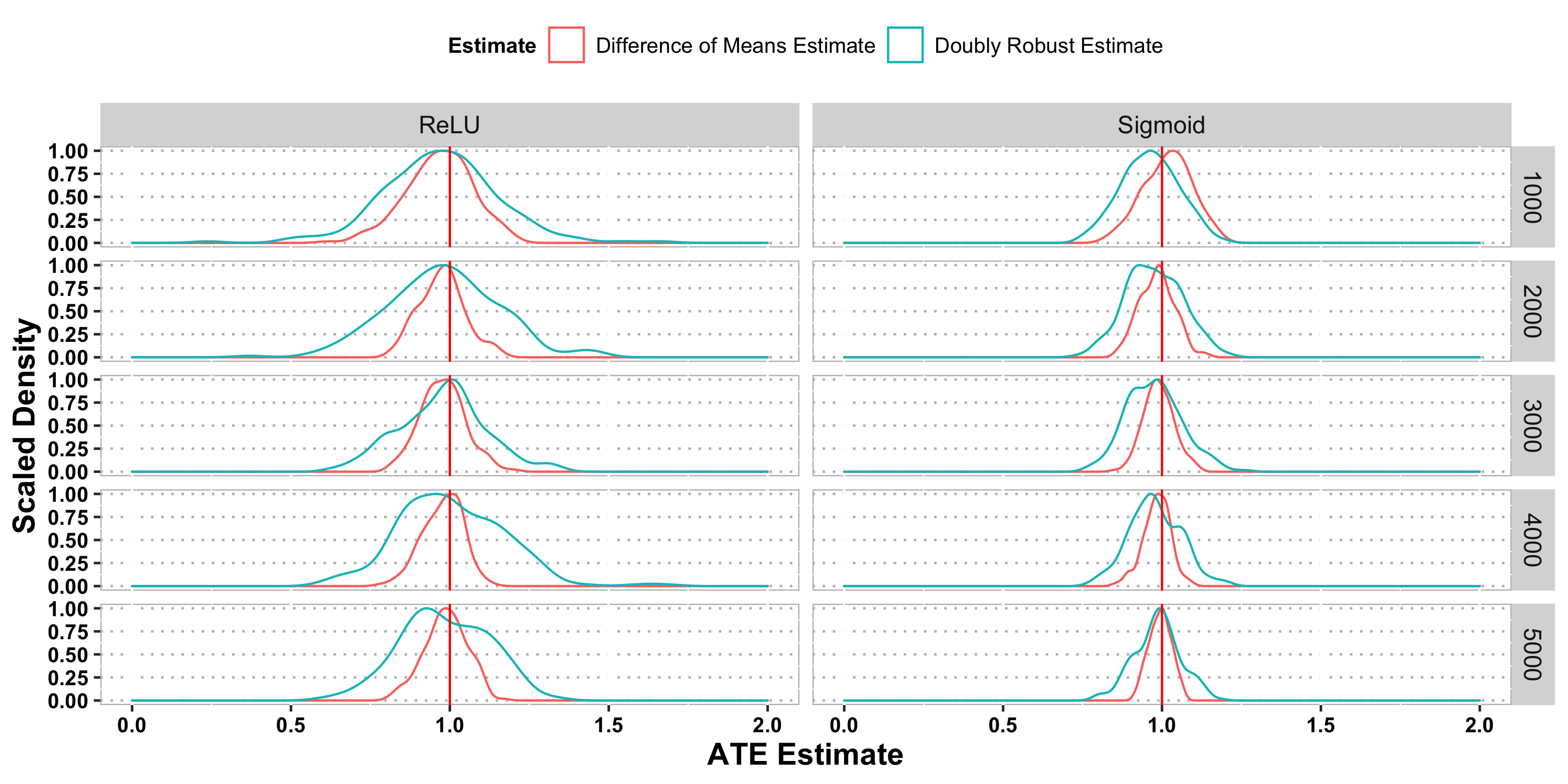}
    \caption{The scaled density of the ATE estimate over 200 replications for different training sample sizes and different activation functions. The red curves correspond to the DNN estimate defined in \eqref{deftauhat} and the blue curves correspond to the doubly robust estimate defined in \eqref{eq: double-robust-split}. The true treatment effect of $\tau = 1$ is shown as a red vertical line.  Here we use a fixed inference sample size of $n = 1000$ and train each network for 800 epochs. From top to bottom, the training sample size $n_1$ increases from 1000 to 5000. Results for different training lengths can be found in Section \ref{Sec.C} of the Supplementary Material.}
    \label{fig:sim_2_overlap_800_epochs}
\end{figure}

Figure \ref{fig:sim_2_overlap_800_epochs} and Table \ref{table:sim_2_800_epochs} present the results of the doubly robust version of the ATE estimate with DNN as a function of the choice of activation (i.e., sigmoid vs. ReLU), the training-to-inference ratio $c$, and the number of epochs varying from $100$ to $500$ (for the construction of the estimated joint regression function $\hat{m}_{\mathcal D_1}(\bx, t)$ as mentioned above). 

\begin{table}[htp]
    \centering
    
\begin{tabular}[t]{ccccccc}
\toprule
$n_1$ & Estimate Type & Activation & Mean & Median & SD & MSE\\
\midrule
 &  & ReLU & 0.9665 & 0.9751 & 0.10528 & 0.01215\\
\cmidrule{3-7}
 & \multirow{-2}{*}{\centering\arraybackslash Difference of Means Estimate} & Sigmoid & 1.0177 & 1.0193 & 0.07743 & 0.00628\\
\cmidrule{2-7}
 &  & ReLU & 0.9757 & 0.9716 & 0.19882 & 0.03992\\
\cmidrule{3-7}
\multirow{-4}{*}{\centering\arraybackslash 1000} & \multirow{-2}{*}{\centering\arraybackslash Doubly Robust Estimate} & Sigmoid & 0.9620 & 0.9631 & 0.08996 & 0.00949\\
\cmidrule{1-7}
 &  & ReLU & 0.9750 & 0.9771 & 0.07266 & 0.00588\\
\cmidrule{3-7}
 & \multirow{-2}{*}{\centering\arraybackslash Difference of Means Estimate} & Sigmoid & 0.9840 & 0.9843 & 0.05472 & 0.00324\\
\cmidrule{2-7}
 &  & ReLU & 0.9792 & 0.9743 & 0.17514 & 0.03095\\
\cmidrule{3-7}
\multirow{-4}{*}{\centering\arraybackslash 2000} & \multirow{-2}{*}{\centering\arraybackslash Doubly Robust Estimate} & Sigmoid & 0.9751 & 0.9715 & 0.08919 & 0.00854\\
\cmidrule{1-7}
 &  & ReLU & 0.9785 & 0.9808 & 0.07092 & 0.00547\\
\cmidrule{3-7}
 & \multirow{-2}{*}{\centering\arraybackslash Difference of Means Estimate} & Sigmoid & 0.9896 & 0.9882 & 0.04791 & 0.00239\\
\cmidrule{2-7}
 &  & ReLU & 0.9772 & 0.9881 & 0.13725 & 0.01926\\
\cmidrule{3-7}
\multirow{-4}{*}{\centering\arraybackslash 3000} & \multirow{-2}{*}{\centering\arraybackslash Doubly Robust Estimate} & Sigmoid & 0.9754 & 0.9749 & 0.08674 & 0.00809\\
\cmidrule{1-7}
 &  & ReLU & 0.9773 & 0.9819 & 0.06328 & 0.00450\\
\cmidrule{3-7}
 & \multirow{-2}{*}{\centering\arraybackslash Difference of Means Estimate} & Sigmoid & 0.9837 & 0.9852 & 0.04295 & 0.00210\\
\cmidrule{2-7}
 &  & ReLU & 1.0051 & 0.9869 & 0.16570 & 0.02735\\
\cmidrule{3-7}
\multirow{-4}{*}{\centering\arraybackslash 4000} & \multirow{-2}{*}{\centering\arraybackslash Doubly Robust Estimate} & Sigmoid & 0.9785 & 0.9753 & 0.08152 & 0.00707\\
\cmidrule{1-7}
 &  & ReLU & 0.9874 & 0.9886 & 0.06703 & 0.00463\\
\cmidrule{3-7}
 & \multirow{-2}{*}{\centering\arraybackslash Difference of Means Estimate} & Sigmoid & 0.9941 & 0.9953 & 0.03458 & 0.00122\\
\cmidrule{2-7}
 &  & ReLU & 0.9843 & 0.9710 & 0.13930 & 0.01955\\
\cmidrule{3-7}
\multirow{-4}{*}{\centering\arraybackslash 5000} & \multirow{-2}{*}{\centering\arraybackslash Doubly Robust Estimate} & Sigmoid & 0.9857 & 0.9902 & 0.07306 & 0.00552\\
\bottomrule
\end{tabular}
    
    \caption{The simulation results corresponding to Figure \ref{fig:sim_2_overlap_800_epochs} for 800 training epochs. Results for different training lengths can be found in Section \ref{Sec.C} of the Supplementary Material.}
    \label{table:sim_2_800_epochs}
\end{table}

From Figure \ref{fig:sim_2_overlap_800_epochs} and Table \ref{table:sim_2_800_epochs}, we see that the sigmoid doubly robust estimator has comparable performance to that of the difference of means estimate (i.e., our first method), but with slightly larger variance. This is consistent with our theoretical results in Theorems \ref{Asynormality} and \ref{DRasynormal}. It is also interesting to observe that for balanced samples (i.e., the case of $c = 1$), the performance of the sigmoid doubly robust estimator was rather close to that of the sigmoid mean difference estimator. When $c$ grows, the latter one has much improved performance while the former stays more or less the same. The fact that the training-to-inference sample ratio has more impact on difference of means estimate is also consistent with our theory. On the contrary,  the ReLU doubly robust estimator had excessively large variance. Also, the training and network tuning for the purpose of ATE inference with ReLU can be more challenging according to our empirical experience. These suggest against the use of ReLU for our application. Results corresponding to different numbers of epochs are presented in Section \ref{Sec.C} of the Supplementary Material.

\section{Real data application} \label{Sec.reda}

As a supplement to our theoretical results and our simulation studies, we demonstrate the practical usage of our proposed methods by studying the effect of 401(k) eligibility on accumulated assets as in \cite{belloni2017program, chernozhukov2004effects, abadie2003semiparametric}.   

There has been a considerable line of research focused on understanding the effect of a 401(k) plan on the accumulated assets of a household. The challenge here is that there is heterogeneity amongst savers and the decision to enroll in a 401(k) plan is non-random\footnote{This is because though a 401(k) plan is a tax-deferred retirement plan that is provided through an employer. Therefore only workers in firms that offer 401(k) plans are eligible.}. 
To address the endogeneity of 401(k) participation, \cite{Poterba1994,POTERBA19951} and \cite{BENJAMIN20031259} used data from the 1991 Survey of Income and Program Participation (SIPP) and argued that eligibility for enrolling in a 401(k) plan can be taken as exogenous after controlling for observables, particularly income. The crux of their argument is that, around the time this data was collected, 401(k) plans were still relatively new and most people based their employment decisions on income, not on whether their employer offered a 401(k) plan. Thus, eligibility for a 401(k) plan could be taken as exogenous conditional on income, and the causal effect of 401(k) eligibility could be directly estimated.


We use the same data as in \cite{belloni2017program}, which consists of 9915 observations at the household level from the 1991 SIPP. Specifically, we use net financial assets as our outcome variable and the covariates are age, income, family size, years of education, and indicators for marital status, two-earner status, defined benefit pension status, IRA participation, and home ownership. 
Since 401(k) eligibility is used as our treatment variable, it is important to note that our estimate of interest is now the average intention to treat.

We randomly sample (without replacement) with sample size varying from 20\% to 50\% of the data for the inference set and use the remaining data as our training set.
With the randomly sampled training and inference sets, we calculate our mean difference estimate and doubly robust estimate for the average intention to treatment.  
Finally, we repeat this process 100 times to generate a distribution of the estimates.

\begin{figure}[ht]
    \centering
    \includegraphics[width=\linewidth]{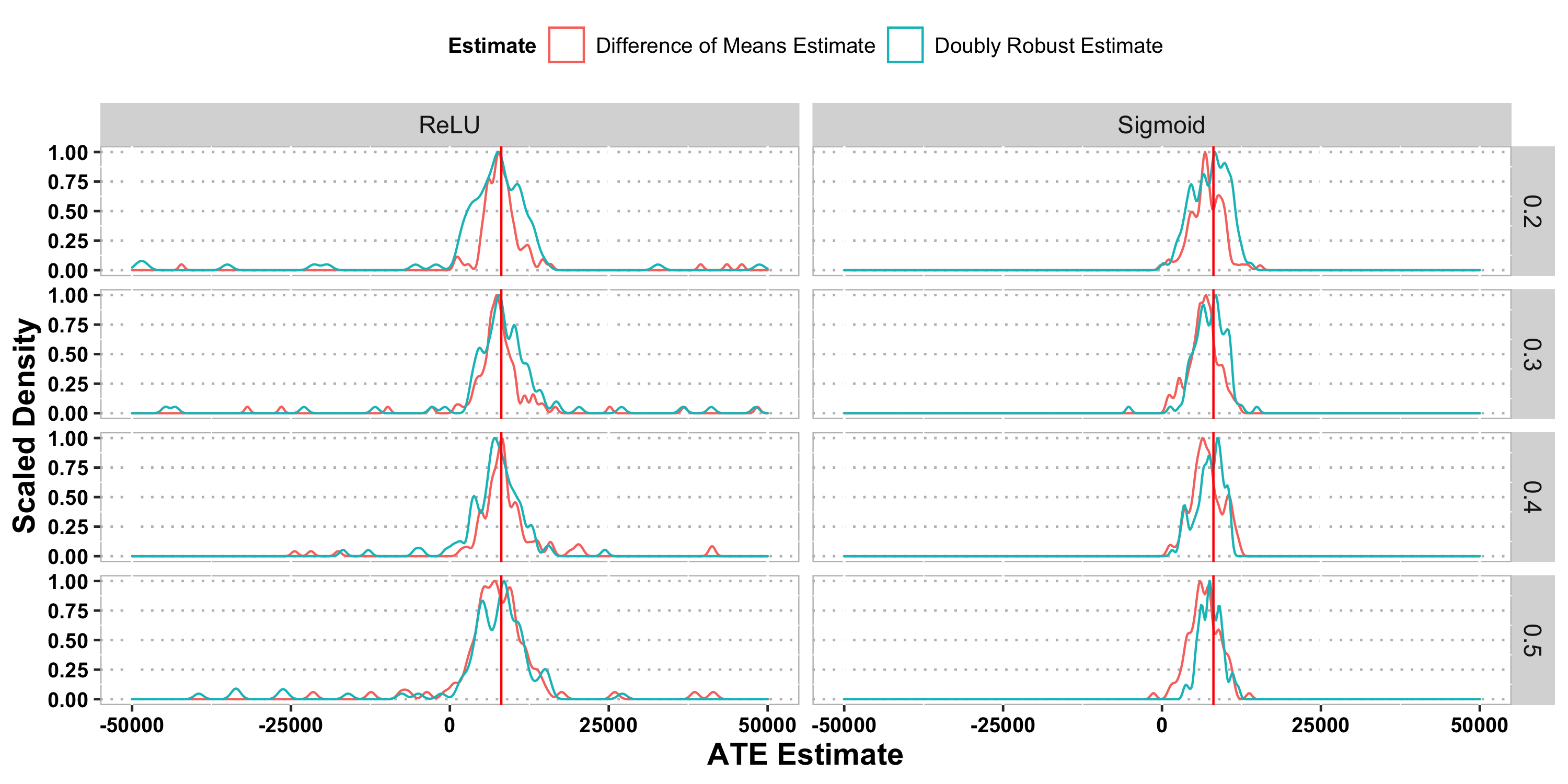}
    \caption{The scaled density of the ATE estimate over 100 replications for different training sample size proportions and different activation functions.  The red curves correspond to the DNN estimate defined in \eqref{deftauhat} and the blue curves correspond to the doubly robust estimate defined in \eqref{eq: double-robust-split}. The red vertical line is the ATE estimate reported in \cite{belloni2017program} from the quadratic spline specification without variable selection of 8093. The rows in the figure correspond to different sizes of the inference set varying from 20\% to 50\% of the data. In this figure, both estimates come from networks trained for 800 epochs. Results for different training lengths can be found in Section \ref{Sec.C} of the Supplementary Material.   
    }
    \label{fig:real_data_800_epochs}
\end{figure}

The results are summarized in Figure~\ref{fig:real_data_800_epochs} and Table~\ref{table:real_data_800_epochs}. It is seen that the distributions of both estimates are uni-modal and close to symmetric, which is similar to what we have observed in the simulation studies.  Compared to the results in \cite{belloni2017program}, both of our estimators have distributions concentrating around the ATE estimate obtained in \cite{belloni2017program} for their quadratic spline specification without variable selection of 8093. However, our estimates have larger robust standard deviations. This is expected because our methods rely on sample splitting, and as revealed in Theorems \ref{varestimate} and \ref{DRvarconsistency}, the convergence rates are determined by the inference set size,
which we vary from 20\% to 50\% of the total data in our application, whereas \cite{belloni2017program} used the entire sample and bootstrap to estimate the robust standard deviation. Comparing our mean difference estimate with our doubly robust estimate, we see that the the latter has larger standard deviations which is consistent with our theory and our simulation studies. In addition, we observe that the estimates from the ReLU network have longer-tailed distributions. Our empirical results also suggest that the intention to treat effect is indeed significantly different from zero. Results corresponding to different numbers of epochs are included in Section \ref{Sec.C} of the Supplementary Material.



\begin{table}[htp]
    \centering
    
\begin{tabular}[t]{ccccc}
\toprule
Inference Proportion & Estimate Type & Activation & Median & Robust SD\\
\midrule
 &  & ReLU & 7780 & 2362\\
\cmidrule{3-5}
 & \multirow{-2}{*}{\centering\arraybackslash Difference of Means Estimate} & Sigmoid & 6911 & 2442\\
\cmidrule{2-5}
 &  & ReLU & 7440 & 4488\\
\cmidrule{3-5}
\multirow{-4}{*}{\centering\arraybackslash 0.2} & \multirow{-2}{*}{\centering\arraybackslash Doubly Robust Estimate} & Sigmoid & 8025 & 3384\\
\cmidrule{1-5}
 &  & ReLU & 7400 & 2669\\
\cmidrule{3-5}
 & \multirow{-2}{*}{\centering\arraybackslash Difference of Means Estimate} & Sigmoid & 6659 & 2036\\
\cmidrule{2-5}
 &  & ReLU & 8127 & 3460\\
\cmidrule{3-5}
\multirow{-4}{*}{\centering\arraybackslash 0.3} & \multirow{-2}{*}{\centering\arraybackslash Doubly Robust Estimate} & Sigmoid & 7723 & 2289\\
\cmidrule{1-5}
 &  & ReLU & 8201 & 2871\\
\cmidrule{3-5}
 & \multirow{-2}{*}{\centering\arraybackslash Difference of Means Estimate} & Sigmoid & 6764 & 2429\\
\cmidrule{2-5}
 &  & ReLU & 7497 & 3310\\
\cmidrule{3-5}
\multirow{-4}{*}{\centering\arraybackslash 0.4} & \multirow{-2}{*}{\centering\arraybackslash Doubly Robust Estimate} & Sigmoid & 7614 & 2035\\
\cmidrule{1-5}
 &  & ReLU & 7473 & 3743\\
\cmidrule{3-5}
 & \multirow{-2}{*}{\centering\arraybackslash Difference of Means Estimate} & Sigmoid & 6549 & 2438\\
\cmidrule{2-5}
 &  & ReLU & 7934 & 4051\\
\cmidrule{3-5}
\multirow{-4}{*}{\centering\arraybackslash 0.5} & \multirow{-2}{*}{\centering\arraybackslash Doubly Robust Estimate} & Sigmoid & 7603 & 1872\\
\bottomrule
\end{tabular}

    \caption{The real data results corresponding to Figure \ref{fig:real_data_800_epochs} for 800 training epochs. Results for different training lengths can be found in Section \ref{Sec.C} of the Supplementary Material.}
    \label{table:real_data_800_epochs}
\end{table}

\section{Discussions} \label{Sec.disc}

In this paper, we have considered the estimation and inference of ATE using deep neural networks. Under the potential outcomes framework, the observed response follows a nonparametric mean regression model, and ATE can be written as the expected difference of the mean regression function corresponding to the treatment and control groups. We have proposed to use DNN to learn the mean regression function, and construct the ATE estimate based on the DNN estimate. We have also derived the asymptotic normality of the ATE estimate using the idea of sample splitting. These ideas and results are further extended to the doubly robust estimator based on the inverse propensity score weighting. Simulation studies and a real data application demonstrate the practical utilities of our methods.

The current theory excludes the ReLU activation because of the smoothness assumption required in establishing the main results. Developing theory for more general activation functions is an interesting topic for future study. In addition, our current consistency rates are derived for functions with finite smoothness parameter $s$. We conjecture that the rates in Propositions \ref{mainthm} and \ref{DRestimation} can be improved to nearly parametric rate when $s=\infty$. We leave such study for future investigation.


\appendix

\renewcommand{\thesubsection}{A.\arabic{subsection}}

\section{Proofs of main results} \label{SecA}

We provide the proofs of Theorems \ref{Asynormality}--\ref{DRvarconsistency}, Propositions \ref{mainthm}--\ref{DRestimation}, and Corollary \ref{PSconditioncor} in this Appendix. The remaining proofs and additional technical details are contained in the Supplementary Material. Throughout the paper, we use $C$ to denote a generic positive constant whose value may change from line to line.

\subsection{Proof of Proposition \ref{mainthm}} \label{SecA.1}

Observe that the difference between our ATE estimator $\widehat \tau_{\mathcal D}$ and the true value of the ATE $\tau$ consists of two major parts: the approximation error and the estimation error. The first part comes from the fact that $m_{\mathcal D}(\BS x,t)$ can be generally biased for nonparametric function approximation, while the second part is because we estimate the population expectation based on a given sample of size $n_{\mathcal D} = |\mathcal D|$. There is also an interplay between these two parts. The theoretical results in \cite{AOS2019Bauer} have tackled the approximation side, and our goal here is to bound the estimation error given the regression function $m_{\mathcal D}(\BS x,t)$. Since the regression function varies as $n_{\mathcal D}\to \infty$, we focus on bounding the error for all possible learned regression functions in the function class $\mathcal H^{(l)}$ to accommodate the approximation process. This is possible thanks to the relatively limited complexity of function class $\mathcal H^{(l)}$, or more precisely, the bound on the covering number of $\mathcal H^{(l)}$ according to the 
learning theory literature. Meanwhile, the correlation between the treatment group and the control group makes no difference. Due to the symmetry of the two groups in estimation, the bounds for one part can be naturally applied to the other part. Thus, for simplicity, we focus only on one part, e.g., the treatment group, in our technical analysis. Throughout the proof, we will use the notation $\mathcal D = \{(\BS X_i, T_i, Y_i)\}_{i=1}^{n_{\mathcal D}}$ to denote the available data set. We will also drop the subscript and write $n:= n_{\mathcal D}$.

Specifically, to bound $|\tau - \widehat \tau_{\mathcal D}|$, the treatment part and the control part can be separated as 
\begin{eqnarray} \label{neweq013}
|\tau - \widehat \tau_{\mathcal D}| &\leq& \big|\mathbb E_{\BS X} m(\BS{X},1) - \frac{1}{n}\sum_{i=1}^{n} m_{\mathcal D}(\BS X_i,1)\big| \nonumber\\
&& + \big|\mathbb E_{\BS X} m(\BS{X},0) - \frac{1}{n}\sum_{i=1}^{n} m_{\mathcal D}(\BS X_i,0)\big|,
\end{eqnarray}
where $\mathbb E_{\BS X}$ represents the expectation over an independent data point $\BS X$ from the same distribution as $\BS X_1$. For the treatment part of (\ref{neweq013}), we have
\begin{eqnarray*}
	\big|\mathbb E_{\BS X} m(\BS{X},1) - \frac{1}{n}\sum_{i=1}^{n} m_{\mathcal D}(\BS X_i,1)\big| &\leq& \big| \mathbb E_{\BS{X}} [m(\BS{X},1) -  m_{\mathcal D}(\BS{X},1)]\big| \\
	&& + \big|\mathbb E_{\BS{X}} m_{\mathcal D}(\BS{X},1)- \frac{1}{n}\sum_{i=1}^{n} m_{\mathcal D}(\BS X_i,1) \big|.
\end{eqnarray*}
An application of Theorem 1 in \cite{AOS2019Bauer} leads to 
\begin{equation}\label{resultinAOS}
\mathbb E_{\mathcal D}{\mathbb E_{\BS{X},T} |m(\BS{X},T) - m_{\mathcal D}(\BS{X},T)|^2}  \leq C_2 (\log n)^{3} n^{-\frac{2s}{2s+p^*}}
\end{equation}
with $C_2$ some positive constant for $n$ sufficiently large, where $\mathbb E_{\mathcal D}$ stands for the expectation over data in $\mathcal D$.  This immediately entails that  
\begin{equation}\label{resultinAOS_con}
{\mathbb E_{\BS{X},T} |m(\BS{X},T) - m_{\mathcal D}(\BS{X},T)|^2} = o_P\left((\log n)^4 n^{-\frac{2s}{2s+p^*}}\right)
\end{equation}
by Chebyshev's inequality. This together with Condition \ref{condRF}(iv) ensures that 
\begin{eqnarray} \label{keyineq}
&&\big| \mathbb E_{\BS{X}} [m(\BS{X},1) - m_{\mathcal D}(\BS{X},1)]\big| \notag\\
&\leq& \sqrt{ \mathbb E_{\BS{X}} |m(\BS{X},1) - m_{\mathcal D}(\BS{X},1)|^2} \notag\\
&\leq& \frac{1}{\sqrt{\delta}}\sqrt{ \mathbb E_{\BS{X}} \{ |m(\BS{X},1) - m_{\mathcal D}(\BS{X},1)|^2 e(\BS X) + |m(\BS{X},0) - m_{\mathcal D}(\BS{X},0)|^2(1 - e(\BS X))\}} \notag\\
&\leq& \frac{1}{\sqrt{\delta}}\sqrt{ \mathbb E_{\BS{X},T} |m(\BS{X},T) - m_{\mathcal D}(\BS{X},T)|^2}\notag\\
& = & o_P((\log n)^{2} n^{-\frac{s}{2s+p^*}}).
\end{eqnarray}

On the other hand, from Theorem 9.1 in \cite{NonpR}, one can bound the difference between the empirical average and its expectation as 
\begin{eqnarray}\label{problematic_formula}
    &&\mathbb P \left(\big|\mathbb E_{\BS{X}}m_{\mathcal D}(\BS{X},1) - \frac{1}{n}\sum_{i=1}^n m_{\mathcal D}(\BS X_i,1)\big| >  \epsilon_{n}\right) \notag\\
    &\leq&\mathbb P\left(\sup_{\hat m \in \mathcal H^{(l)}}\big|\mathbb E_{\BS{X}} \hat m(\BS{X},1)- \frac{1}{n}\sum_{i=1}^n \hat m(\BS X_i,1)\big| >  \epsilon_{n}\right) \notag\\
    &\leq& 8\mathbb E_{\mu_{n}}[\mathcal N(\epsilon_{n},\mathcal H^{(l)},L_1(\mu_{n}))]\exp\left(-\frac{n\epsilon_{n}^2}{128}\right),
\end{eqnarray}
where $\mathcal N(\epsilon_{n},\mathcal H^{(l)},L_1(\mu_{n}))$ stands for the covering number of the function class $\mathcal H^{(l)}$ with metric $\| f \|_{L_1(\mu_{n})} = \mu_n(|f|) = \frac{1}{{n}}\sum_{i=1}^n |f(\BS{X}_i)|$ at scale $\epsilon_{n} > 0$ and $\mu_n$ the empirical measure. It follows from the fundamental theory of covering numbers that 
\begin{equation}\label{ieq1}
\mathbb E_{\mu_{n}}[\mathcal N(\epsilon_{n},\mathcal H^{(l)},L_1(\mu_{n}))]\leq\mathcal N(\epsilon_{n},\mathcal H^{(l)},\|\cdot\|_\infty)
\end{equation}
and 
\begin{equation}\label{ieq2}
\mathcal N(\epsilon_{n},\mathcal H^{(l)},\|\cdot\|_\infty)\leq \exp(C_3(\log {n})M)
\end{equation}
with some positive constant $C_3$, given that $\epsilon \geq \frac{1}{n^{C_4}}$ for some positive constant $C_4$; see, e.g., Lemma 2 in \cite{AOS2019Bauer}.

With the choice of $\epsilon_{n} = \sqrt{\frac{128\log(n\cdot\mathcal N(\frac{1}{\sqrt{n}},\mathcal H^{(l)},\|\cdot\|_\infty))}{n}}$, one can deduce that 
\begin{eqnarray} \label{neweq014}
&&\mathbb P\Big(\big|\mathbb E_{\BS{X}}m_{\mathcal D}(\BS{X},1) - \frac{1}{n}\sum_{i=1}^n m_{\mathcal D}(\BS X_i,1)\big| >  C_5\sqrt{\frac{\log n+C_3(\log n)M}{n}}\Big) \nonumber\\
&\leq & \mathbb P\Big(\big|\mathbb E_{\BS{X}}m_{\mathcal D}(\BS{X},1) - \frac{1}{n}\sum_{i=1}^n m_{\mathcal D}(\BS X_i,1)\big| > \epsilon_{n}\Big) \nonumber\\
&\leq & 8\mathbb E_{\mu_{n}}[\mathcal N(\epsilon_{n},\mathcal H^{(l)},L_1(\mu_{n}))]\exp(-\frac{n\epsilon_{n}^2}{128}) \nonumber\\
&\leq & 8\frac{\mathcal N(\epsilon_{n},\mathcal H^{(l)},\|\cdot\|_\infty)}{n\cdot\mathcal N(\frac{1}{\sqrt{n}},\mathcal H^{(l)},\|\cdot\|_\infty)} \nonumber\\
&\leq & 8/n,
\end{eqnarray}
where $C_5$ is some positive constant. Here, the first inequality in (\ref{neweq014}) results from the fact that $\epsilon_{n} \leq C_5\sqrt{\frac{\log n+C_3(\log n)M}{n}}$ holds for some positive constant $C_5$. The second and third inequalities are implied by inequalities \eqref{problematic_formula}, (\ref{ieq1}), and (\ref{ieq2}). Finally, the last inequality is due to the monotone decreasing property of the covering number with respect to the scale. Hence, we can obtain by Condition \ref{condRF}(iii) that 
\begin{equation} \label{neweq015}
\big|\mathbb E_{\BS{X}}m_{\mathcal D}(\BS{X},1) - \frac{1}{n}\sum_{i=1}^n m_{\mathcal D}(\BS{X}_i,1)\big| = o_P\left(\sqrt{\log(n) n^{-\frac{2s}{2s+p^*}}}\right).
\end{equation}

Combining the above bounds in (\ref{keyineq}) and (\ref{neweq015}) yields 
\begin{eqnarray} \label{neweq016}
\big|\mathbb E m(\BS{X},1) - \frac{1}{n}\sum_{i=1}^n m_{\mathcal D}(X_i,1)\big| = o_P\left((\log n)^{2} n^{-\frac{s}{2s+p^*}}\right).
\end{eqnarray}
Similarly, it can derived for the control part that 
\begin{eqnarray} \label{neweq017}
\big|\mathbb E m(\BS{X},0) - \frac{1}{n}\sum_{i=1}^n m_{\mathcal D}(X_i,0)\big| = o_P\left((\log n)^{2} n^{-\frac{s}{2s+p^*}}\right).
\end{eqnarray}
Therefore, in view of (\ref{neweq013}), (\ref{neweq016}), and (\ref{neweq017}), we have 
\begin{equation} \label{neweq018}
|\tau -\widehat\tau_{\mathcal D} | = o_P\left((\log n)^{2} n^{-\frac{s}{2s+p^*}}\right),
\end{equation}
which completes the proof of Proposition
\ref{mainthm}.

\subsection{Proof of Theorem \ref{Asynormality}} \label{SecA.2}

The high-level idea of the proof has been summarized in the main text just before Theorem \ref{Asynormality}. For the ease of presentation, we write $\mathcal D_2 = \{(\BS X_i, T_i, Y_i)\}_{i=1}^n$. Let us consider the decomposition
\begin{eqnarray}\label{decomthm1}
   &&\sqrt{n} (\widehat \tau(\mathcal D_1,\mathcal D_2) - \tau) = \Big(\frac{1}{\sqrt{n}}\sum_{i=1}^n m(\BS X_i,1) - \mathbb E_{\BS X} m(\BS X,1)\Big)\notag\\ &&-\Big(\frac{1}{\sqrt n}\sum_{i=1}^n m(\BS X_i,0) - \mathbb E_{\BS X} m(\BS X,0)\Big)\notag\\
&&+ \frac{1}{\sqrt{n}}\sum_{i=1}^n\Big( m_{\mathcal D_1}(\BS X_i,1)-m(\BS X_i,1) \Big)- \frac{1}{\sqrt n}\sum_{i=1}^n\Big( m_{\mathcal D_1}(\BS X_i,0)- m(\BS X_i,0)\Big)\notag\\
&&:= A_1 - A_0 + B_1-B_0.
\end{eqnarray}
The first two terms together $A_1-A_0$ can be written as the sum of i.i.d random variables with bounded variance.  Thus, an application of the classical central limit theorem (CLT) leads to 
\begin{equation}\label{eq: A1A0-clt}
    A_1-A_0 \toD{ N(0, \sigma^2)}.
\end{equation}

We next prove that $B_1=o_P(1)$ and $B_0=o_P(1)$. Then these results together with \eqref{eq: A1A0-clt} can complete the proof of this theorem.  Since the proofs for terms $B_1$ and $B_0$ are almost identical, we only show the former. First, since $\mathcal D_1$ and $\mathcal D_2$ are independent, each containing i.i.d. observations, an application of Chebyshev's inequality entails that for any $x>0$, it holds that 
\begin{align*}
    &P\left(\Big|\frac{1}{n}\sum_{i=1}^n\Big( m_{\mathcal D_1}(\BS X_i,1)-m(\BS X_i,1) - \mathbb E_{\BS X}[m_{\mathcal D_1}(\BS X, 1) -m(\BS X,1)]  \Big)|>n^{-1/2}x \Big| \mathcal D_1\right)\\
    &\leq  \frac{\sum_{i=1}^n\mathbb E_{\BS X_i}|m_{\mathcal D_1}(\BS X_i,1)-m(\BS X_i,1)|^2 - n\big(\mathbb E_{\BS X}[ m_{\mathcal D_1}(\BS X, 1) - m(\BS X,1)]\big)^2 }{n x^2}\\
    & \leq \frac{\mathbb E_{\BS X}|m_{\mathcal D_1}(\BS X_i,1)-m(\BS X_i,1)|^2}{x^2}.
\end{align*}
Noting that $|\mathcal D_1| = n^{\gamma}$, by \eqref{resultinAOS} we have
\begin{equation}\label{equ0010}
    \mathbb E_{\mathcal D_1}\mathbb E_{\BS X}|m_{\mathcal D_1}(\BS X_i,1)-m(\BS X_i,1)|^2 \leq C(\gamma\log n)^3n^{-\frac{2\gamma s}{2s+p^*}}.
\end{equation}

Taking $x = (\log n)^2n^{-\frac{\gamma s}{2s+p^*}}$ and by the properties of the conditional expectation, we can deduce that 
\begin{align}\label{equ0011}
    &P(|\frac{1}{n}\sum_{i=1}^n\Big( m_{\mathcal D_1}(\BS X_i,1)-m(\BS X_i,1) - \mathbb E_{\BS X}m_{\mathcal D_1}(\BS X, 1) + \mathbb E[m(\BS X,1)]  \Big)|>n^{-1/2}x )\notag\\
    & \leq \frac{\mathbb E_{\mathcal D_1}\mathbb E_{\BS X}|m_{\mathcal D_1}(\BS X_i,1)-m(\BS X_i,1)|^2}{x^2}\rightarrow 0.
\end{align}
This result along with \eqref{keyineq} entails that 
$$
|B_1| = \Big|\frac{1}{\sqrt n}\sum_{i=1}^n\big( m_{\mathcal D_1}(\BS X_i,1)-m(\BS X_i,1)\big)\Big| = o_p\big((\log n)^2n^{\frac{1}{2}-\frac{\gamma s}{2s+p^*}}\big) = o_P(1),
$$
which concludes the proof of Theorem \ref{Asynormality}.

\subsection{Proof of Theorem \ref{varestimate}} \label{SecA.4}
Denote by $\sigma^2(\mathcal D_1)$ the population variance for $\widehat \tau(\mathcal D_1)$ conditional on $\mathcal D_1$; that is, 
\begin{equation}
    \sigma^2(\mathcal D_1) = \Var(\widehat \tau(\mathcal D_1)|\mathcal D_1),
\end{equation}
where $\widehat \tau(\mathcal D_1) = m_{\mathcal D_1}(\BS X,1) - m_{\mathcal D_1}(\BS X,0)$. Hence, we have 
\begin{equation}\label{eq:sig2-decomp}
    |\widehat \sigma^2(\mathcal D_1,\mathcal D_2) - \sigma^2| \leq |\widehat \sigma^2(\mathcal D_1,\mathcal D_2) - \sigma^2(\mathcal D_1)| + |\sigma^2(\mathcal D_1) - \sigma^2|.
\end{equation}
We can obtain that $m(\BS x, t)$ is bounded on its domain by the bounded support assumption in Condition \ref{condRF}(i) and the smoothness assumption on the mean regression function $m(\BS x, t)$. 

For the second term on the right-hand side of (\ref{eq:sig2-decomp}), it holds that 
\begin{eqnarray}
|\sigma^2(\mathcal D_1) - \sigma^2| &=& \big| \mathbb E_{\BS X}[|m_{\mathcal D_1}(\BS X,1) - \mathbb E_{\BS X}m_{\mathcal D_1}(\BS X,1) - m_{\mathcal D_1}(\BS X,0) + \mathbb E_{\BS X}m_{\mathcal D_1}(\BS X,0)|^2|\mathcal D_1]\notag\\
&&- \mathbb E_{\BS X}|m(\BS X,1) - \mathbb E_{\BS X}m(\BS X,1) - m(\BS X,0) + \mathbb E_{\BS X}m(\BS X,0)|^2\big|\notag\\
&\leq & C\log(n^\gamma)\big(\mathbb E_{\BS X} \big|[m_{\mathcal D_1}(\BS X,1) - m(\BS X,1)] -\mathbb E_{\BS X}[ m_{\mathcal D_1}(\BS X,1) - m(\BS X,1)]\big| \notag\\
&&+\mathbb E_{\BS X} \big|[m_{\mathcal D_1}(\BS X,0) - m(\BS X,0)] -\mathbb E_{\BS X}[ m_{\mathcal D_1}(\BS X,0) - m(\BS X,0)]\big|  \big)\notag\\
&\leq & 2C\log(n^\gamma)\big(\mathbb E_{\BS X} \big|m_{\mathcal D_1}(\BS X,1) - m(\BS X,1)\big| + \mathbb E_{\BS X} \big|m_{\mathcal D_1}(\BS X,0) - m(\BS X,0)\big| \big),\notag
\end{eqnarray}
where $C\log(n^\gamma)$ comes from the truncation step involved in the definition of $ m_{\mathcal D_1}$. Then an application of inequality (\ref{keyineq})  yields
\begin{equation*}
    \mathbb E_{\BS X} \big|m_{\mathcal D_1}(\BS X,1) - m(\BS X,1)\big| \leq \left\{ \mathbb E_{\BS X} \big|m_{\mathcal D_1}(\BS X,1) - m(\BS X,1)\big|^2\right\}^{1/2} = o_P((\log n^\gamma)^2n^{-\frac{s\gamma}{2s+p^*}}).
\end{equation*}
The same result can be obtained for $\mathbb E_{\BS X} \big|m_{\mathcal D_1}(\BS X,0) - m(\BS X,0)\big| \big)$ using similar arguments. Thus, we can obtain that  
\begin{equation}\label{varcontrol}
    |\sigma^2(\mathcal D_1) - \sigma^2| = o_P((\log n^\gamma)^3n^{-\frac{s\gamma}{2s+p^*}}).
\end{equation}

The first term on the right-hand side of \eqref{eq:sig2-decomp} can be tackled with an application of the weak law of large numbers for a triangular array. In particular, we set $Z_{n,i} = \hat \tau_i(\mathcal D_1)$ for $i\in \mathcal D_2$ with $\hat\tau_i(\mathcal D_1)$ defined below \eqref{deftauhat}. Observe that $|Z_{n,i}|$ is upper bounded by $C\log(n^\gamma)$ with some positive constant $C$. Then, by Chebyshev's inequality conditional on $\mathcal D_1$ for arbitrary $\epsilon > 0$, it holds that 
\begin{align*}
   & \mathbb P\Big(\Big|\frac{\sum_{i\in \mathcal D_2} (Z_{n,i} - \mathbb E [Z_{n,i}|\mathcal D_1])}{n}\Big| > \epsilon \Big) \\
   & =\mathbb E\Big[\mathbb P\Big(\Big|\frac{\sum_{i\in \mathcal D_2} (Z_{n,i} - \mathbb E [Z_{n,i}|\mathcal D_1])}{n}\Big| > \epsilon \Big\vert \mathcal D_1\Big)\Big] \\
    &\leq \mathbb E\Big[ \frac{\sum_{i=1}^n \mathbb E[Z_{n,i}^2\vert \mathcal D_1]}{n^2\epsilon^2}\Big] \leq \frac{C^2\gamma^2\log^2(n)}{\epsilon^2n}
\end{align*}
and similarly,
\begin{equation*}
    \mathbb P\Big(\Big|\frac{\sum_{i\in \mathcal D_2} (Z_{n,i}^2 - \mathbb E [Z_{n,i}^2|\mathcal D_1])}{n}\Big| > \epsilon \Big) \leq 
    \frac{C^4\gamma^4\log^4(n)}{\epsilon^2n}.
\end{equation*}

By choosing $\epsilon = \frac{\log^3(n)}{\sqrt n}$, we have with probability at least $1- \frac{C^4\gamma^4}{\log^2 n} - \frac{C^3\gamma^3}{\log^4(n)}$ that 
\begin{align*}
    \Big|\frac{\sum_{i\in \mathcal D_2} (Z_{n,i} - \mathbb E [Z_{n,i}|\mathcal D_1])}{n}\Big|\leq  \frac{\log^3(n)}{\sqrt n} \ \text{ and } \ 
    \Big|\frac{\sum_{i\in \mathcal D_2} (Z_{n,i}^2 - \mathbb E [Z_{n,i}^2|\mathcal D_1])}{n}\Big| \leq \frac{\log^3(n)}{\sqrt n}.
\end{align*}
Thus, we can deduce that 
\begin{eqnarray} \label{new.eq.L003}
&&|\hat \sigma^2(\mathcal D_1,\mathcal D_2) - \sigma^2(\mathcal D_1)| \notag\\
&= & \big|\frac{n}{n-1}\big(\frac{1}{n}\sum_{i\in \mathcal D_2} (Z_{n,i}^2 - \mathbb E[Z_{n,i}^2|\mathcal D_1])  - (\frac{1}{n}\sum_{i\in \mathcal D_2} Z_{n,i})^2 + (\mathbb E [Z_{n,i}^2|\mathcal D_1] \big) + \frac{1}{n-1}\sigma^2(\mathcal D_1)\big|\notag\\
& \leq & 2\big|\frac{1}{n}\sum_{i\in \mathcal D_2} (Z_{n,i}^2 - \mathbb E[ Z_{n,i}^2|\mathcal D_1])\big| + 4C\log(n^\gamma)\Big|\frac{1}{n}\sum_{i\in \mathcal D_2} (Z_{n,i} - \mathbb E [Z_{n,i}|\mathcal D_1])\Big| + \frac{2\sigma^2(\mathcal D_1)}{n}\notag \\
&\leq &  \frac{2\log^3(n)}{\sqrt n} + \frac{4C\gamma \log^4(n)}{\sqrt n}+ \frac{2\sigma^2}{n} + \frac{2}{n}|\sigma^2(\mathcal D_1) - \sigma^2|
\end{eqnarray}
for $n$ large enough with probability at least $1 - \frac{C^4\gamma^4}{\log^2(n)} - \frac{C^3\gamma^3}{\log^4(n)}$. 

Therefore, combining the bounds in  (\ref{varcontrol}) and (\ref{new.eq.L003}) yields that
\begin{equation}
    |\widehat \sigma^2(\mathcal D_1, \mathcal D_2) - \sigma^2| = o_P(\frac{\log^3(n)}{\sqrt n} + \frac{\log^4(n))}{\sqrt n}) + O(\frac{1}{n}) + o_P((\log n)^3n^{-\frac{s\gamma}{2s+ p^*}}).
\end{equation}
Since we assume that $\gamma > 1 + \frac{p^*}{2s}$, it follows that
\begin{equation*}
    |\widehat\sigma^2(\mathcal D_1, \mathcal D_2) - \sigma^2| = o_P(n^{-1/2}(\log n)^4).
\end{equation*}
The above consistency result together with Theorem \ref{Asynormality} and Slutsky's lemma completes the proof of Theorem \ref{varestimate}.

\subsection{Proof of Proposition \ref{DRestimation}} \label{SecA.5}
Recall that we assume that $|\mathcal D_1| = n$ and for the $i$th observation in $\mathcal D_1$, we denote it as $(\BS X_i, T_i, Y_i)$. We start with the decomposition\footnote{The subscripts $\mathcal D_1$ for $\hat e_{\mathcal D_1}$ and $\hat m_{\mathcal D_1}$ are omitted in this proof and the proofs of Theorem \ref{DRasynormal} and Theorem \ref{DRvarconsistency} for notational simplicity.}
\begin{eqnarray}\label{squaredclassdim}
    |\hat \tau_{DR,\mathcal D_1}(\hat e,\hat m) - \tau| &=& \frac{1}{n}\sum_{i\in \mathcal D_1}(\phi_i(\hat e,\hat m) - \phi_i(\hat e,m)) - \frac{1}{n}\sum_{i\in \mathcal D_1} (\psi_i(\hat e,\hat m) - \psi_i(\hat e, m))\notag\\
    &&+\frac{1}{n}\sum_{i\in \mathcal D_1}(\phi_i(\hat  e,m) - \phi_i(e, m)) - \frac{1}{n}\sum_{i\in \mathcal D_1}(\psi_i(\hat e, m) - \psi_i(e, m))\notag\\
    &&+ \frac{1}{n}\sum_{i\in \mathcal D_1}\big[(\phi_i( e, m) - \mathbb E\phi(e, m)) - (\psi_i( e,  m) - \mathbb E\psi(e, m))]\big] \notag\\
    &:=& E_1 - E_0 + F_1 - F_0 + G.
\end{eqnarray}
We will show that each term on the right-hand side of (\ref{squaredclassdim}) is an $o_P(1)$ term with some rate of convergence. Since the treatments for terms $E_1$ and $E_0$ are similar, we will only deal with the former one. The explicit form of term $E_1$ can be written as
\begin{equation}
    E_1 = \frac{1}{n}\sum_{i\in \mathcal D_1} (\frac{T_i}{\hat e(\BS X_i)} - 1) (m_1(\BS X_i) - \hat m_1(\BS X_i)).
\end{equation}

Based on (i) of Condition \ref{condPS}, we can deduce that 
\begin{eqnarray}
    |E_1| &\leq & \frac{1}{n}\sum_{i\in \mathcal D_1} 2C_2\log(n)|m_1(\BS X_i) - \hat m_1(\BS X_i)| \notag\\
    &\leq&  2C_2\log(n) \Big(\frac{1}{n}\sum_{i\in \mathcal D_1} |m_1(\BS X_i) - \hat m_1(\BS X_i)|^2\Big)^{1/2} \notag \\
    &\leq& 2C_2\log(n) \Big(\Big|\frac{1}{n}\sum_{i\in \mathcal D_1} |m_1(\BS X_i) - \hat m_1(\BS X_i)|^2 - \mathbb E_{\BS X} |m_1(\BS X) - \hat m_1(\BS X)|^2\Big| \notag\\
    &&+ \mathbb E_{\BS X} |m_1(\BS X) - \hat m_1(\BS X)|^2\Big)^{1/2}.\notag
\end{eqnarray}
 The first term inside the square root on the right-hand side above can be bounded by applying Theorem 9.1 in \cite{NonpR}, using arguments similar to those used 
 for obtaining inequality (\ref{problematic_formula}). 
 Specifically, let us define a new function class
 $$
 \widetilde{\mathcal H}^{(l)} = \{g: g(\BS x, t)=(\text{trunc}(\widetilde m(\BS x, t), C\log n)-m(\BS x, t))^2 \text{ with } \widetilde m \in\mathcal H^{(l)}\}.
 $$

Then for $n$ sufficiently large, it holds that 
 $$
\mathbb E_{\mu_{n}}[\mathcal N(\epsilon_{n},\widetilde{\mathcal H}^{(l)},L_1(\mu_{n}))]\leq\mathcal N(\frac{\epsilon_{n}}{2C(\log n)},\mathcal H^{(l)},\|\cdot\|_\infty).
$$
Thus, an application of similar arguments as in the proof of (\ref{problematic_formula}) leads to 
\begin{equation*}
    \big| \frac{1}{n}\sum_{i\in \mathcal D_1} |\hat m_1(\BS X_i) - m_1(\BS X_i)|^2 - \mathbb E_{\BS X} |\hat m_1(\BS X) - m_1(\BS X)|^2 \big|  = o_P(\sqrt{\log(n) n^{-\frac{2s}{2s+p^*}}}).
\end{equation*}
The expectation term above can be bounded similar to  (\ref{resultinAOS_con}). 
Hence, we can obtain that 
\begin{equation}
    |E_1|  = o_P(\log^3(n)n^{-\frac{s}{2s+p^*}}) + o_P(\log^{5/4}(n)n^{-\frac{s/2}{2s+ p^*}}) = o_P(1). 
\end{equation}
As for term $F_1$, we can write it as 
\begin{equation}
    F_1 = \frac{1}{n}\sum_{i\in \mathcal D_1} (\frac{1}{\widehat e(\BS X_i)} - \frac{1}{e(\BS X_i)})T_i(Y_i -  m_1(\BS X_i)) = \frac{1}{n}\sum_{i\in \mathcal D_1} (\frac{1}{\widehat e(\BS X_i)} - \frac{1}{e(\BS X_i)}) T_i\varepsilon_i,
\end{equation}
which entails that $\mathbb E F_1 = 0$. Due to (i) of Condition \ref{condPS}, it follows that 
\begin{eqnarray}
    \mathbb E[F_1^2|\BS X_1,\BS X_2,\dots,\BS X_n] &\leq& \mathbb \Var[\varepsilon]\frac{1}{n}\sum_{i\in \mathcal D_1} \big(\frac{e(\BS X_i) - \widehat e(\BS X_i)}{\widehat e(\BS X_i)e(\BS X_i)}\big)^2 \notag \\
    &\leq & \frac{\Var[\varepsilon]C_2^2 \log^2(n)}{\delta^2}\frac{1}{n}\sum_{i\in \mathcal D_1}\big(e(\BS X_i) - \widehat e(\BS X_i)\big)^2.
\end{eqnarray}
Then an application of (ii) of Condition \ref{condPS} implies that
\begin{equation}
    \mathbb E[F_1^2] = o_P(1),
\end{equation}
which shows that term $F_1$ vanishes in probability asymptotically thanks to Chebyshev's inequality. 

Applying similar arguments to terms $E_0$ and $F_0$, we can obtain that 
\begin{equation}
    |E_0| , |F_0| = o_P(1).
\end{equation}
On the other hand, due to the boundedness of both $\phi_i(e,m)$ and $\psi_i(e,m)$, an application of the law of large numbers entails that
\begin{equation}
    G = o_P(1).
\end{equation}
Therefore, it follows that the doubly robust estimator $\widehat \tau_{DR, \mathcal D_1}(\widehat e,\widehat m)$ is a consistent estimator of $\tau$, which concludes the proof of Proposition \ref{DRestimation}.

\subsection{Proof of Theorem \ref{DRasynormal}} \label{SecA.6}

The proof idea is similar to that of Theorem \ref{Asynormality}, which begins with the decomposition
\begin{eqnarray}
    \sqrt{n}|\widehat \tau_{DR,\mathcal D_2}(\widehat e,\widehat m) - \tau| &=& \frac{1}{\sqrt{n}}\sum_{i \in \mathcal D_2}\big[\phi_i(\widehat e,\widehat m) -\phi_i(\widehat e, m) \big]- \frac{1}{\sqrt{n}}\sum_{i \in \mathcal D_2}\big[\psi_i(\widehat e,\widehat m) -\psi_i(\widehat e, m) \big]\notag\\
    && +\frac{1}{\sqrt{n}}\sum_{i \in \mathcal D_2}\big[\phi_i(\widehat e, m) -\phi_i( e, m) \big]- \frac{1}{\sqrt{n}}\sum_{i \in \mathcal D_2}\big[\psi_i(\widehat e, m) -\psi_i( e, m) \big]\notag\\
    &&+ \frac{1}{\sqrt{n}}\sum_{i\in\mathcal D_2}[\phi_i(e,m) - \mathbb E_{(Y,\BS X,T)}\phi(e,m)]\notag\\
    && - \frac{1}{\sqrt{n}}\sum_{i\in\mathcal D_2}[\psi_i(e,m) - \mathbb E_{(Y,\BS X,T)}\psi(e,m)]\notag\\
    &:=& H_1 - H_0 + I_1 - I_0 + J_1 - J_0.
\end{eqnarray}

We first consider term $H_1$ above. Since $\mathcal D_1$ and $\mathcal D_2 = \{(\BS X_i, T_i, Y_i)\}_{i=1}^n$ are independent, it follows from Chebyshev's inequality that for any $x>0$,
\begin{align*}
    &\mathbb P( |H_1 - \mathbb E[H_1|\mathcal D_1]|>x |\mathcal D_1)\leq \frac{\sum_{i\in\mathcal D_2}\mathbb E[(\phi_i(\widehat e,\widehat m) -\phi_i(\widehat e, m))^2|\mathcal D_1]}{n x^2}\\
    &\leq \frac{\mathbb E[|(\frac{T_1}{\widehat e(\BS X_1)} -1 )(\widehat m_1(\BS X_1) - m_1(\BS X_1))|^2|\mathcal D_1]}{x^2}\\
    &  \leq C_2^2(\log n)^2\mathbb E[|\widehat m_1(\BS X_1) - m_1(\BS X_1)|^2|\mathcal D_1]/x^2,
\end{align*}
where in the last step we have used  (i) of Condition \ref{condPS}. 
By the definition of the conditional probability and \eqref{resultinAOS}, we can obtain that 
\begin{align*}
    \mathbb P(|H_1 - \mathbb E[H_1|\mathcal D_1]| \geq x) 
    & \leq C_2^2(\log n)^2\mathbb E_{\mathcal D_1}\mathbb E[|\widehat m_1(\BS X_1) - m_1(\BS X_1)|^2|\mathcal D_1]/x^2\\
    & \leq  C_2^2 (\log n)^2 o((\log n)^3 n^{-\frac{2s}{2s+ p^*}})/x^2.
\end{align*}
By letting $x = (\log n)^3n^{-\frac{s}{2s+p^*}}$, we have 
$$
H_1-\mathbb E[H_1|\mathcal D_1] = o_P((\log n)^3n^{-\frac{s}{2s+p^*}}).
$$

Further, we can deduce that 
\begin{align*}
    |\mathbb E[H_1|\mathcal D_1]| &= \sqrt{n}|\mathbb E_{\BS X_1,T_1}[(\frac{T_1}{\widehat e(\BS X_1)} - 1)(m_1(\BS X_1) - \widehat m_1(\BS X_1))]| \\
    & \leq C_2\sqrt{n}\log(n)\mathbb E_{\BS X_1}|(e(\BS X_1) -\widehat  e(\BS X_1))(m_1(\BS X_1) - \widehat m_1(\BS X_1))|\\
    & \leq C_2\sqrt{n}\log(n)\sqrt{\mathbb E_{\BS X_1}|e(\BS X_1) - \widehat e(\BS X_1)|^2}\sqrt{ \mathbb E_{\BS X_1}|m_1(\BS X_1) - \widehat m_1(\BS X_1)|^2}.
\end{align*}
Combining (iii) of Condition \ref{condPS}, the assumption of $p^* < 2s$, and inequality (\ref{resultinAOS}) results in 
$$
\mathbb E[H_1|\mathcal D_1] = o_P(\sqrt{n}\log(n) n^{-1/4} (\log n)^3 n^{-\frac{s}{2s+p^*}}) = o_P(1).
$$
Thus, the above results together entail that
$$
H_1 = (H_1 - \mathbb E[H_1|\mathcal D_1]) + \mathbb E[H_1|\mathcal D_1] = o_P((\log n)^3n^{-\frac{s}{2s+p^*}}) +o_P(1) = o_P(1).
$$

Similar arguments can be applied to term $I_1$. In particular, note that $\mathbb E[I_1|\mathcal D_1]=0$. Also, it holds that 
\begin{align*}
   \mathbb P\left(|I_1| \geq x\right)&= \mathbb E_{\mathcal D_1} \mathbb P\left(|I_1 - \mathbb E[I_1|\mathcal D_1]| \geq x|\mathcal D_1\right) \\
    & \leq \mathbb E_{\mathcal D_1} \frac{\mathbb E[I_1^2 |\mathcal D_1]}{x^2} \\
    & \leq \mathbb E_{\mathcal D_1} \mathbb E[|T_1(Y_1 - m_1(\BS X_1))\frac{\widehat e(\BS X_1) - e(\BS X_1)}{\widehat e(\BS X_1) e(\BS X_1)}|^2 |\mathcal D_1]/x^2 \\
    & \leq  \Var[\varepsilon]\frac{C_2^2(\log n)^2}{\delta}\mathbb E_{\mathcal D_1}\mathbb E[|\widehat e(\BS X_1) - e(\BS X_1)|^2 |\mathcal D_1]/x^2 \\
    & = o\left((\log n)^2n^{-1/2}\right)/x^2,
\end{align*}
where $(\BS X_1, T_1, Y_1)\in \mathcal D_2$ is independent of $\mathcal D_1$.  Taking $x = (\log n)^2n^{-1/4}$, we can obtain that 
$$
I_1 =  o_P\left((\log n)^2n^{-1/4}\right).
$$

Similarly, we can show that
$$
H_0 = o_P(1) \ \text{ and } \  I_0=o_P(1).
$$
Moreover, it holds that $J_1 - J_0$ converges in  distribution to $\mathcal N(0, \sigma_{DR}^2)$. Therefore, combining all these results yields that 
\begin{equation}
    \widehat \tau_{DR,\mathcal D_2}(\widehat e_{\mathcal D_1},\widehat m_{\mathcal D_1}) - \tau \toD \mathcal N(0, \sigma_{DR}^2),
\end{equation}
where $\sigma^2_{DR} = \Var_{Y_1,\BS X_1,T_1}[\phi(e,m) - \psi(e,m)] = \Var(m_1(\BS X) - m_0(\BS X)) + \Var(\varepsilon)\mathbb E\frac{1}{e(\BS X)(1-e(\BS X))}$. This completes the proof of Theorem \ref{DRasynormal}.

\subsection{Proof of Corollary \ref{PSconditioncor}}\label{SecA.7}
We only need to verify (i) and (iii) of Condition \ref{condPS}. Indeed, (i) of Condition \ref{condPS} holds due to the truncation on $\widehat e_{\mathcal D_1}(\BS x)$. 
As for (iii) of Condition \ref{condPS}, we can show by Proposition \ref{mainthm}, in which bound (\ref{resultinAOS}) can be applied to $\widehat e(\BS X)$ as well, that
\begin{equation} \label{new.eq.L004}
    \mathbb E_{\mathcal D_1}\mathbb E_{\BS X}|\widehat e_{\mathcal D_1}(\BS X) - e(\BS X)|^2 \leq  C\log^3(n)n^{-\frac{2s_e}{2s_e+p^*}}
\end{equation}
for some constant $C$ and all sufficiently large $n$. Since $p^* < 2s_e$, the right-hand side of (\ref{new.eq.L004}) is indeed an $o(n^{-1/2})$ term. Therefore, given (i) and (iii) of Condition \ref{condPS}, the desired conclusions of Corollary \ref{PSconditioncor} follow from Theorem \ref{DRasynormal}.

\subsection{Proof of Theorem \ref{DRvarconsistency}} \label{SecA.8}
Recall that $\mathcal D_2 = \{(\BS X_i, T_i, Y_i)\}_{i=1}^n$. Let us define 
$$
\sigma_{DR}^2(\widehat e_{\mathcal D_1},\widehat m_{\mathcal D_1}) = \Var[\phi_1(\widehat e_{\mathcal D_1},\widehat m_{\mathcal D_1})-\psi_1(\widehat e_{\mathcal D_1},\widehat m_{\mathcal D_1})|\mathcal D_1].
$$
Observe that
$$
\sigma^2_{DR} = \Var[\phi_1(e, m)-\psi_1( e, m)]
$$
and $\phi_1$ and $\psi_1$ are defined on the observation $(\BS X_1, T_1, Y_1)\in\mathcal D_2$. Then an application of similar arguments as in the proof of Theorem \ref{varestimate} shows that $\sigma_{DR}^2(\widehat e_{\mathcal D_1},\widehat m_{\mathcal D_1})$ is a consistent estimator of $\sigma^2_{DR}$. 
It holds that 
\begin{align*}
   &\phi_1(\widehat e_{\mathcal D_1},\widehat m_{\mathcal D_1})-\psi_1(\widehat e_{\mathcal D_1},\widehat m_{\mathcal D_1})\\
   &= \frac{T_1}{\widehat e_{\mathcal D_1}(\BS X_1)} \left( m_1(\BS X_1) - \widehat m_{\mathcal D_1,1}(\BS X_1) \right) + \widehat m_{\mathcal D_1,1}(\BS X_1) \\
   & \quad- \frac{1-T_1}{1-\widehat e_{\mathcal D_1} (\BS X_1)} \left(m_0(\BS X_1) - \widehat m_{\mathcal D_1,0}(\BS X_1) \right) - \widehat m_{\mathcal D_1,0}(\BS X_1) \\
   & \quad + \left(\frac{T_1}{\widehat e_{\mathcal D_1}(\BS X_1)} - \frac{1-T_1}{1-\widehat e_{\mathcal D_1}(\BS X_1)}\right)\varepsilon_1.
\end{align*}

Since $\varepsilon_1$ is independent of $(\BS X_1, T_1)$ and $\mathcal D_1$ and has mean zero, it follows that
\begin{align*}
  \sigma_{DR}^2(\widehat e_{\mathcal D_1},\widehat m_{\mathcal D_1}) &=  \Var\Big( \frac{T_1}{\widehat e_{\mathcal D_1}(\BS X_1)}(m_1(\BS X_1) - \widehat m_{\mathcal D_1,1}(\BS X_1))  \\
  &\quad+ \widehat m_{\mathcal D_1,1}(\BS X_1) - \frac{1-T_1}{1-\widehat e(\BS X_1)}(m_0(\BS X_1) - \widehat m_{\mathcal D_1,0}(\BS X_1)) - \widehat m_{\mathcal D_1,0}(\BS X_1) \Big|\mathcal D_1\Big)\\
  & \quad + \Var\Big(\left(\frac{T_1}{\widehat e_{\mathcal D_1}(\BS X_1)} - \frac{1-T_1}{1-\widehat e_{\mathcal D_1}(\BS X_1)}\right)\varepsilon_1\Big|\mathcal D_1\Big) \\
  & : = I_1 + I_2.
\end{align*}
Similarly, we can show that 
\begin{align*}
  \sigma_{DR}^2 & =  \Var\Big( m_{1}(\BS X_1)  - m_{0}(\BS X_1)\Big) + \Var\Big(\left(\frac{T_1}{ e(\BS X_1)} - \frac{1-T_1}{1-e(\BS X_1)}\right)\varepsilon_1\Big) \\
  & : = II_1 + II_2.
\end{align*}

First, let us consider term $I_2-II_2$. By the independence of $\varepsilon_1$ with $(\BS X_1, T_1)$ and $\mathcal D_1$, we have
\begin{align*}
    |I_2-II_2| &= \Var(\varepsilon_1)\left|\mathbb E\left[\left(\frac{T_1}{\widehat e_{\mathcal D_1}(\BS X_1)} - \frac{1-T_1}{1-\widehat e_{\mathcal D_1}(\BS X_1)}\right)^2 \Big| \mathcal D_1\right]  - \mathbb E\left[\left(\frac{T_1}{e(\BS X_1)} - \frac{1-T_1}{1- e(\BS X_1)}\right)^2\right]\right|\\
    &  = \Var(\varepsilon_1)\left|\mathbb E_{\BS X_1}\left[\frac{ e(\BS X_1)}{\widehat e_{\mathcal D_1}^2(\BS X_1)} - \frac{1-e(\BS X_1)}{(1-\widehat e_{\mathcal D_1}(\BS X_1))^2}\right]   - \mathbb E_{\BS X_1}\left[\frac{1}{e(\BS X_1)} - \frac{1}{1- e(\BS X_1)}\right]\right|\\
    &\leq \Var[\varepsilon_1]C_2^2\log^2(n)\mathbb E[|\widehat e_{\mathcal D_1}(\BS X_1)  - e(\BS X_1)||\mathcal D_1],
\end{align*}
where in the last step we have used the boundedness assumption of $\widehat e_{\mathcal D_1}$ stated in Condition \ref{condPS}(ii). Furthermore, by Condition \ref{condPS}(iii) and the fact that $Y_1$ is a sub-Gaussian random variable, it follows from Chebyshev's inequality that
\begin{align*}
    |I_2-II_2| = o_P(\log^2(n)n^{-1/4}).
\end{align*}

Next we analyze term $I_1-II_1$. Note that the random variable inside the variance in $I_1$ can be upper bounded by $C\log^2 n$ almost surely with respect to $\BS X_1$ with $C$ some generic positive constant.  By the variance representation 
\[
\Var(R_1) - \Var(R_2) = \mathbb E[(R_1+R_2)(R_1-R_2)] - (\mathbb E R_1 - \mathbb E R_2) (\mathbb E R_1 + \mathbb E R_2) 
\]
for any random variables $R_1$ and $R_2$ and some basic calculations, we can deduce that 
\begin{align*}
|I_1-II_1|& \leq C(\log n)^2\mathbb E\left[\left|\left(\frac{T_1}{\widehat e_{\mathcal D_1}(\BS X_1)}-1\right)(m_1(\BS X_1) - \widehat m_{\mathcal D_1,1}(\BS X_1)) \right.\right.\\
&\quad\left.\left.+\left(\frac{1-T_1}{1-\widehat e_{\mathcal D_1}(\BS X_1)}-1\right)(m_1(\BS X_1) - \widehat m_{\mathcal D_1,1}(\BS X_1))\right| \Big| \mathcal D_1    \right]\\
& \leq C(\log n)^3\left\{\mathbb E[|m_1(\BS X_1) - \widehat m_{\mathcal D_1,1}(\BS X_1)| |\mathcal D_1]\right. \\
&\quad\left.+ \mathbb E[|m_0(\BS X_1) - \widehat m_{\mathcal D_1,0}(\BS X_1)| |\mathcal D_1]\right\}.
\end{align*}
In view of \eqref{keyineq}, it holds that 
$$
|I_1-II_1| = o_P(\log^5(n)n^{-\frac{s}{2s+p^*}}).
$$
Thus, combining the above results leads to 
\begin{equation}\label{DRvarcontrol}
    \big|\sigma_{DR}^2(\widehat e_{\mathcal D_1},\widehat m_{\mathcal D_1}) -  \sigma_{DR}^2\big|=  o_P(\log^5(n)n^{-\frac{s}{2s+p^*}} +\log^2(n)n^{-1/4}).
\end{equation}

Denote by $Z_{n,i} = \widehat \tau_i(\widehat e_{\mathcal D_1},\widehat m_{\mathcal D_1})$ (c.f. \eqref{defsampleDRvariance}). Then similar arguments as in the proof of Theorem \ref{varestimate} can be applied with the aid of the law of large numbers. In particular, we can obtain that 
\begin{equation*}
    |\widehat \sigma_{DR,\mathcal D_2}^2(\widehat e_{\mathcal D_1},\widehat m_{\mathcal D_1}) - \sigma_{DR}^2(\widehat e_{\mathcal D_1},\widehat m_{\mathcal D_1})|  = o_P(\frac{\log^7(n)}{\sqrt{n}}) + \frac{\sigma_{DR}^2(\widehat e,\widehat m)}{n-1}.
\end{equation*}
Together with bound (\ref{DRvarcontrol}), the above inequality yields that
\begin{equation*}
    |\widehat \sigma_{DR,\mathcal D_2}^2(\widehat e_{\mathcal D_1},\widehat m_{\mathcal D_1}) - \sigma_{DR}^2|  = o_P(\frac{\log^7(n)}{\sqrt{n}}) + o_P(\log^5(n) n ^{-\frac{s}{2s+p^*}}) + o_P(\log^2(n) n^{-1/4}).
\end{equation*}
With the assumption of $p^* < 2s$, the above bound can be further simplified as 
\begin{equation}
    |\widehat \sigma_{DR,\mathcal D_2}^2(\widehat e_{\mathcal D_1},\widehat m_{\mathcal D_1}) - \sigma_{DR}^2|  = o_P(\log^2(n)n^{-1/4}).
\end{equation}
Therefore, the asymptotic normality of $\sqrt{n}\big(\widehat \tau_{DR,\mathcal D_2}(\widehat e_{\mathcal D_1},\widehat m_{\mathcal D_1}) - \tau\big)/{\widehat \sigma_{DR,\mathcal D_2}(\widehat e_{\mathcal D_1},\widehat m_{\mathcal D_1})}$ holds by Slutsky's lemma, which concludes the proof of Theorem \ref{DRvarconsistency}.

\bibliographystyle{plain}
{
\bibliography{references}}

\begin{thebibliography}{10}

\bibitem{abadie2003semiparametric}
Alberto Abadie.
\newblock Semiparametric instrumental variable estimation of treatment response
  models.
\newblock {\em Journal of Econometrics}, 113(2):231--263, 2003.

\bibitem{Matias-survey2018}
Alberto Abadie and Matias~D. Cattaneo.
\newblock Econometric methods for program evaluation.
\newblock {\em Annual Review of Economics}, 10(1):465--503, 2018.

\bibitem{athey2015machine1}
Susan Athey.
\newblock Machine learning and causal inference for policy evaluation.
\newblock In {\em Proceedings of the 21th ACM SIGKDD International Conference
  on Knowledge Discovery and Data Mining}, pages 5--6. ACM, 2015.

\bibitem{athey2015machine}
Susan Athey and Guido~W. Imbens.
\newblock Machine learning methods for estimating heterogeneous causal effects.
\newblock {\em Stat}, 1050(5), 2015.

\bibitem{AOS2019Bauer}
Benedikt Bauer and Michael Kohler.
\newblock On deep learning as a remedy for the curse of dimensionality in
  nonparametric regression.
\newblock {\em Ann. Statist.}, 47:2261--2285, 2019.

\bibitem{AOS-Bauer-SuppA}
Benedikt Bauer and Michael Kohler.
\newblock Supplement to ``on deep learning as a remedy for the curse of
  dimensionality in nonparametric regression''.
\newblock {\em Annals of Statistics}, 2019.

\bibitem{belloni2017program}
Alexandre Belloni, Victor Chernozhukov, Iv{\'a}n Fern{\'a}ndez-Val, and
  Christian Hansen.
\newblock Program evaluation and causal inference with high-dimensional data.
\newblock {\em Econometrica}, 85(1):233--298, 2017.

\bibitem{BENJAMIN20031259}
Daniel~J. Benjamin.
\newblock Does 401({K}) eligibility increase saving? evidence from propensity
  score subclassification.
\newblock {\em Journal of Public Economics}, 87(5):1259--1290, 2003.

\bibitem{double-machine2018}
Victor Chernozhukov, Denis Chetverikov, Mert Demirer, Esther Duflo, Christian
  Hansen, Whitney Newey, and James Robins.
\newblock {Double/debiased machine learning for treatment and structural
  parameters}.
\newblock {\em The Econometrics Journal}, 21(1):C1--C68, 01 2018.

\bibitem{chernozhukov2004effects}
Victor Chernozhukov and Christian Hansen.
\newblock The effects of 401({K}) participation on the wealth distribution: an
  instrumental quantile regression analysis.
\newblock {\em Review of Economics and Statistics}, 86(3):735--751, 2004.

\bibitem{Demirkayaetal2021}
Emre Demirkaya, Yingying Fan, Lan Gao, Jinchi Lv, Patrick Vossler, and Jingbo
  Wang.
\newblock Nonparametric inference of heterogeneous treatment effects with
  two-scale distributional nearest neighbors.
\newblock {\em arXiv preprint arXiv:1808.08469}, 2021.

\bibitem{Fan2016ImprovingCB}
Jianqing Fan, Kosuke Imai, Han Liu, Yang Ning, and Xiaolin Yang.
\newblock Improving covariate balancing propensity score : A doubly robust and
  efficient approach.
\newblock {\em Working paper}, 2016.

\bibitem{doubly-robust2011}
Michele~Jonsson Funk, Daniel Westreich, Chris Wiesen, Til Stürmer, M.~Alan
  Brookhart, and Marie Davidian.
\newblock {Doubly robust estimation of causal effects}.
\newblock {\em American Journal of Epidemiology}, 173(7):761--767, 03 2011.

\bibitem{NonpR}
László Gy\"{o}rfi, Michael Kohler, Adam Krzy\`zak, and Harro Walk.
\newblock {\em A Distribution-Free Theory of Nonparametric Regression}.
\newblock Springer, New York, 2002.

\bibitem{Imbens-survey2009}
Guido~W. Imbens and Jeffrey~M. Wooldridge.
\newblock Recent developments in the econometrics of program evaluation.
\newblock {\em Journal of Economic Literature}, 47(1):5--86, March 2009.

\bibitem{LouizosEtAl_arxiv17}
Christos Louizos, Uri Shalit, Joris Mooij, David Sontag, Richard~S. Zemel, and
  Max Welling.
\newblock Causal effect inference with deep latent-variable models.
\newblock In {\em Proceedings of the 31st International Conference on Neural
  Information Processing Systems}, NIPS'17, 2017.

\bibitem{Poterba1994}
James~M. Poterba and Steven~F. Venti.
\newblock {401({K}) plans and tax-deferred saving}.
\newblock In {\em {Studies in the Economics of Aging}}, NBER Chapters, pages
  105--142. National Bureau of Economic Research, Inc, June 1994.

\bibitem{POTERBA19951}
James~M. Poterba, Steven~F. Venti, and David~A. Wise.
\newblock Do 401({K}) contributions crowd out other personal saving?
\newblock {\em Journal of Public Economics}, 58(1):1--32, 1995.

\bibitem{scarselli1998universal}
Franco Scarselli and Ah~Chung Tsoi.
\newblock Universal approximation using feedforward neural networks: A survey
  of some existing methods, and some new results.
\newblock {\em Neural Networks}, 11(1):15--37, 1998.

\bibitem{sekhon2008neyman}
Jasjeet~S. Sekhon.
\newblock The neyman-rubin model of causal inference and estimation via
  matching methods.
\newblock {\em Oxford Handbook of Political Methodology}, 2008.

\bibitem{ShalitEtAl_icml17}
Uri Shalit, Fredrik~D. Johansson, and David Sontag.
\newblock Estimating individual treatment effect: generalization bounds and
  algorithms.
\newblock In {\em Proceedings of the 34th International Conference on Machine
  Learning}, pages 3076--3085, 2017.

\bibitem{wager2018a}
S.~Wager and S.~Athey.
\newblock Estimation and inference of heterogeneous treatment effects using
  random forests.
\newblock {\em Journal of the American Statistical Association},
  113:1228–1242, 2018.

\bibitem{Zhangetal2016}
Haixiang Zhang, Yinan Zheng, Zhou Zhang, Tao Gao, Brain Joyce, Grace Yoon, Wei
  Zhang, Joel Schwartz, Allan Just, Elena Colicino, Pantel Vokonas, Lihui Zhao,
  Jinchi Lv, Andrea Baccarelli, Lifang Hou, and Lei Liu.
\newblock Estimating and testing high-dimensional mediation effects in
  epigenetic studies.
\newblock {\em Bioinformatics}, 32:3150--3154, 2016.

\end{thebibliography}

	
\newpage
\appendix
\setcounter{page}{1}
\setcounter{section}{1}
\renewcommand{\theequation}{A.\arabic{equation}}
\renewcommand{\thesubsection}{A.\arabic{subsection}}
\setcounter{equation}{0}
	
\begin{center}{\bf \Large Supplementary Material to ``Dimension-Free Average Treatment Effect Inference with Deep Neural Networks''}
		
\bigskip
		
Xinze Du, Yingying Fan, Jinchi Lv, Tianshu Sun and Patrick Vossler
\end{center}
	
\noindent This Supplementary Material contains the proofs of Corollary \ref{cor1} and some technical lemmas, and additional numerical results for the simulation and real data examples in Sections \ref{Sec.simu}--\ref{Sec.reda}. All the notation is the same as defined in the main body of the paper.

\renewcommand{\thesubsection}{B.\arabic{subsection}}

\section{Additional proofs and technical details} \label{Sec.newB}

\subsection{Proof of Corollary \ref{cor1}} \label{SecA.3}
The main idea of the proof is similar to that of the proof for Theorem \ref{Asynormality}. Using the same decomposition as in (\ref{decomthm1}), it is seen that we only need to bound $B_1 = \frac{1}{\sqrt{n}}\sum_{i=1}^n \big(m_{\mathcal D_1}(\BS X_i,1) - m(\BS X_i,1)\big)$ and $B_0= \frac{1}{\sqrt{n}}\sum_{i=1}^n \big(m_{\mathcal D_1}(\BS X_i,0) - m(\BS X_i,0)\big)$ under the new conditions of Corollary \ref{cor1}. In the proof of Theorem \ref{Asynormality}, to bound terms $B_1$ and $B_0$ we have used Proposition \ref{mainthm} and the results established in \cite{AOS2019Bauer}. We will establish parallel results under the conditions of Corollary \ref{cor1} in the next subsection. Using Lemma \ref{modprop1} in Section \ref{Sec.B.3} (which contains parallel results to those in Proposition \ref{mainthm}), we can deduce that 
\begin{align*}
    |B_1| = &  \Big|\frac{1}{\sqrt{n}}\sum_{i=1}^n \big(m_{\mathcal D_1}(\BS X_i,1) - m(\BS X_i,1)\big)\Big|\\
    \leq & o_P((\log(|\mathcal D_1|))^{2 + p^*/2}|\mathcal D_1|^{-1/2}) + \sqrt{n}\big|\mathbb E_{\BS X}\big(m_{\mathcal D_1}(\BS X,1) - m(\BS X,1)\big)\big|\\
    \leq & o_P\big((\log(|\mathcal D_1|))^{2+ p^*/2}\big(\frac{|\mathcal D_2|}{|\mathcal D_1|}\big)^{1/2}\big)\\
    = & o_P\Big(\frac{(\log(n) + k\log(\log(n)))^{2+p^*/2}}{(\log(n))^{k/2}}\Big) = o_P(1).
\end{align*}
Similarly, we can also obtain that $|B_0| = o_P(1)$. Therefore, the asymptotic normality in Corollary \ref{cor1} holds.

\subsection{Some key lemmas for proving Corollary \ref{cor1}} \label{Sec.B.2}

In \cite{AOS2019Bauer}, $(s,C)$-smooth functions for fixed $s$ and $C$ are investigated and the results derived therein involve several constants that depend on the smoothness parameter $s$ implicitly. For $m(\BS x)$ satisfying  Condition \ref{condpoly}(i), it is also $(s, C)$-smooth for any $s\in \mathbb N$. Consequently, the result of Proposition \ref{mainthm} holds for each $s\in \mathbb N$. However, since the constants involved in the proof of Proposition \ref{mainthm} are not uniform over all $s\in \mathbb N$, the consistency rate therein may not hold uniformly over all $s\in \mathbb N$. Because of this, we cannot simply send $s$ to infinity to prove the results of Corollary \ref{cor1}. Instead, we adapt the proof ideas in \cite{AOS2019Bauer} to establish our desired results. 

The above arguments also help us understand the necessity of assuming the polynomial functional form in Condition \ref{condpoly}(i).  With such an assumption, the universal approximation power of two-layer deep neural networks for polynomial functions established in \cite{scarselli1998universal} can be used to show that all the polynomial functions appearing in the definition of $m(\BS x)$ are uniformly well approximated. Hence, results parallel to Proposition \ref{mainthm}, which are summarized in Lemma \ref{modprop1}, can be obtained for $m(\BS x)$ satisfying Condition \ref{condpoly}(i).  Then Corollary \ref{cor1} follows naturally.  On the contrary, without the polynomial function assumption, we would likely encounter approximation errors that are nonuniform across $s$ when using two-layer neural networks, which takes us back to the challenges discussed in the previous paragraph. This provides some justifications on Condition \ref{condpoly}(i).

\textit{Notation}. We first introduce some notation that will be used in our subsequent proofs. Let $f^{(n)}(x)$ be the $n$th order derivative of function $f:\mathbb R\to \mathbb R$ at $x$. For a polytope $K\subset \mathbb R^p$ bounded by hyperplanes $\BS u_j \cdot \BS x + w_j \leq 0$ ($j = 1,\dots, H$) with $\BS u_1,\dots, \BS u_H \in \mathbb R^p$ and $w_1,\dots, w_H\in\mathbb R$, define $K_\delta^0$ and $K_\delta^{C}$ for $\delta > 0$ as
$$
K_{\delta}^0 = \big\{ \BS x\in \mathbb R^p : \BS u_j \cdot \BS x + w_j \leq -\delta, \quad \forall j\in \{1,\dots,H\}\big\},
$$
$$
K_{\delta}^C= \big\{ \BS x\in \mathbb R^p : \BS u_j \cdot \BS x + w_j \geq \delta, \quad \text{for some } j\in \{1,\dots,H\}\big\}.
$$
Denote by $x^{(v)}$ the $v$th component of vector $\BS x\in \mathbb R^p$, and $|\BS x|_1$  the $L_1$-norm defined as $|\BS x|_1 = \sum_{v=1}^d |x^{(v)}|$.

\medskip
The following lemma is adapted from  Theorem 2 in \cite{AOS-Bauer-SuppA}.

\begin{lemma}\label{modthm2}
Let $a \geq 1$ and $\lambda  >0$ be two given constants. Assume that $m: \mathbb R^p \to \mathbb R$ is a polynomial function defined as 
\begin{equation}\label{polynomialregfunction}
    m(\BS x) = \sum_{|\bm{\alpha} | \leq q_0} r_{\bm{\alpha}}x^{\bm{\alpha}}
\end{equation}
with $\max_{|\bm{\alpha} | \leq q_0}\big| r_{\bm{\alpha}}\big| = \bar{r}_m$, and $\nu$ is an arbitrary probability measure on $\mathbb R^p$. Let $N \in \mathbb N_0$ be chosen such that $N\geq q_0$ and $\sigma :\mathbb R \to [0,1]$ the sigmoid function. Then for any $\eta \in (0,1)$ and $M\in \mathbb N$ such that $M^\lambda \geq 2(N+|t_\sigma|)(\frac{2^{N+1}}{\sigma^{(N)}(t_\sigma)} + 1)$ in which $t_{\sigma}\in (0,1)$ can be chosen such that $\sigma^{(i)}(t_{\sigma}) \neq 0$ for all $i\in \mathbb N_0$, and $M \geq a$, there exists a neural network of type
\begin{equation}
    t(\BS x) = \sum_{i=1}^{ {p+N \choose p} (N+1)(M+1)^p }\mu_i\sigma\Big(\sum_{l=1}^{4d} \lambda_{i,l} \sigma\big(\sum_{v = 1}^p \theta_{i,l,v} x^{(v)} + \theta_{i,l,0}\big) + \lambda_{i,0}\Big)
\end{equation}
such that 
\begin{equation}\label{neweq0040}
   |t(\BS x) - m(\BS x)| \leq c_{13} a^{N+q_0 +3} M^{-\lambda} 
\end{equation}
holds for all $\BS x \in [-a, a]^p$ up to a set of $\nu$-measure less than or equal to $\eta$. The coefficients of $t(\BS x)$ can be bounded by
$$
|\mu_i| \leq c_{14}a^{q_0}M^{N \lambda},
$$
$$
|\lambda_{i,l}|\leq M^{p+ \lambda (N+2)},
$$
$$
|\theta_{i,l,v} | \leq 6 \frac{p}{\eta}M^{p+ \lambda(2N+3) + 1}
$$
for all $i\in \{1,\dots,{p+N \choose p} (N+1)(M+1)^p\}$, $l \in \{0,\dots,4d\}$, and $v\in \{0,\dots,p\}$, where the positive constants $c_{13}$ and $c_{14}$ are free of $M$ and $\lambda$. 
\end{lemma}

\begin{remark}
In the proof of Theorem 2 in \cite{AOS2019Bauer}, parameter $\lambda$ that controls both the bounds for the coefficients and affects the bound for the approximation error is chosen to be slightly greater than $q_0$ (to be exact $\lambda = q_0 +r$ for some $r\in (0,1]$). If we assume Condition \ref{condRF}(iii) instead of Condition \ref{condpoly}(i) so that $m(\BS x)$ does not take the polynomial form, for the Taylor expansion $p(\BS x)$ of  $m(\BS x)$ at point $\BS x_0$ to order $q_0$, it holds that 
$$
|t(\BS x) - m(\BS x)| \leq |t(\BS x) - p(\BS x)| + |p(\BS x) - m(\BS x)|.
$$
The first term on the right-hand side above enjoys the same bound as in \eqref{neweq0040} because $p(\BS x)$ is a polynomial function, while the second term can be bounded by $c\Vert \BS x - \BS x_0\Vert^{q_0}$ for some constant $c$ that depends only on $q_0$ and $p$, according to Lemma 8 in \cite{AOS-Bauer-SuppA}. Within each cube $C_{\bm{i}}$ that will be defined in \eqref{defcubes}, we have 
\[ c\Vert \BS x - \BS x_0\Vert^{q_0} \leq c p^{(q_0+1)/2}a^{q_0 + 1}M^{-q_0-1}.
\]
Thus, setting $\lambda = q_0 + r$ makes the two bounds of roughly the same order and, meanwhile, minimizes the bounds for the coefficients, yielding the minimal complexity of the neural networks.

In contrast, by assuming Condition \ref{condpoly}(i), the second term $|p(\BS x) - m(\BS x)|$ on the right-hand side above vanishes. Thus, we no longer require that $\lambda = q_0 + r$, and instead, $\lambda$ here can be some arbitrary positive number. 
In Lemmas \ref{modthm3} and \ref{modthm1} to be presented later, we will apply the result here by setting $\lambda = \lambda_n$ as specified in Condition \ref{condpoly}(ii) to obtain the desired convergence rate.
\end{remark}

\noindent \textit{Proof}. The proof is adapted from that of Theorem 2 in \cite{AOS2019Bauer}. It is presented here for the sake of completeness. Note that the existence of $t_\sigma$ is guaranteed by the discussion of $N$-admissible in the ``Sigmoidal Squasher is $N$-admissible" section in \cite{AOS-Bauer-SuppA}.
Let $\{C_{\bm i}: \bm{i} = (i_1,i_2,\dots,i_p)\in \{1,\dots,M+1\}^p\}$ be a partition of the hypercube $C = [-a - \frac{2a}{M}, a]^p$, where $C_{\bm i}$ is the subcube defined as 
\begin{eqnarray}\label{defcubes}
   & [-a + (i_1 - 2)\frac{2a}{M}, -a + (i_1 - 1)\frac{2a}{M}] \times\dots \nonumber \\
    & \dots \times [-a + (i_p - 2)\frac{2a}{M}, -a + (i_p - 1)\frac{2a}{M}].
\end{eqnarray}
Denote by $\BS x_{\bm{i}}$ the ``bottom left" corner of cube $C_{\bm{i}}$; that is, for $\bm{i} = (i_1,\dots, i_p)$,
$$
\BS x_{\bm{i}} = \big(-a + (i_1 - 2)\frac{2a}{M},\dots,-a + (i_p - 2)\frac{2a}{M}\big).
$$
We can extend the definition of $\BS x_{\bm{i}}$ to all $\bm{i}\in\{1,2,\dots,M+2\}^p$ with $C_{\bm{i}}$ defined in (\ref{defcubes}). 

For some $\lambda >0 $, we can apply Lemma 7 in \cite{AOS2019Bauer} to function $m(\BS x)$ and let $K$ defined therein be $C_{\bm{i}}$. Then it follows that for $M$ large enough such that 
$$
(\frac{N a}{(p+1)M^\lambda} + |t_\sigma|)(2\frac{2^NM^{\lambda(N+1)}}{\sigma^{(N)}(t_\sigma)} +1 ) \leq M^{p+ \lambda(N+2)}(\frac{3}{4} - M^{-p - \lambda(2N+3)})
$$
and $M \geq a$, neural networks $t(\BS x)$ of type
\begin{equation}\label{nnform}
    t(\BS x) = \sum_{j=1}^{ {p+N \choose p} (N+1)(M+1)^p }\mu_i\sigma\Big(\sum_{l=1}^{4d} \lambda_{i,l} \sigma\big(\sum_{v = 1}^p \theta_{i,l,v} x^{(v)} + \theta_{i,l,0}\big) + \lambda_{i,0}\Big)
\end{equation}
exist with coefficients bounded as 
\begin{align*}
|\mu_i| &\leq c_{14}a^{q_0}M^{Np},\\
|\lambda_{i,l}|&\leq M^{p+ \lambda (N+2)},\\
|\theta_{i,l,v} | &\leq 6 \frac{p}{\eta}M^{p+ \lambda(2N+3) + 1}
\end{align*}
for all $i\in \{1,\dots,{p+N \choose p} (N+1)(M+1)^p\}$, $l \in \{0,\dots,4p\}$, and $v\in \{0,\dots,p\}$ such that
\begin{align*}
|t(\BS x) - m(\BS x)| &\leq c_{22} \bar{r}(m) a^{N+3} M^{-\lambda}  & \text{ for } \BS x \in (C_{\BS i})_\delta^0\cap[-a,a]^p,\\
|t(\BS x)| &\leq c_{23}\bar{r}(m) M^{-p-2 \lambda}   & \text{ for } \BS x \in (C_{\BS i})_\delta^C\cap[-a,a]^p,\\
|t(\BS x)| &\leq c_{24}\bar{r}(m) M^{N \cdot \lambda}  & \text{ for } \BS x \in \mathbb R^p.
\end{align*}
Here, $\bar{r}(m)$ is some constant depending on $q_0$, the order of $m(\BS x)$, and $(C_{\BS i})_{\delta}^0$ and $(C_{\BS i})_{\delta}^C$ are defined analogously to $K_{\delta}^0$ and $K_{\delta}^C$, respectively. The constants $c_{22}$, $c_{23}$, and $c_{24}$ depend only on $p$ and $N$. Since the polynomial functional form of $m(\BS x)$ stays the same across different cubes, the result above holds for all cubes with all the constants remaining unchanged. That is, the above results hold for $K = C_{\bm{i}}$ for any $\bm{i}\in \{1,\dots,M+1\}^p$.

 By Lemma 3 in \cite{AOS2019Bauer}, $\bar{r}(m)$ in the representation above can be upper bounded as
$$
\bar{r}(m) \leq c_{27}a^{q_0},
$$
where constant $c_{27}$ here can be chosen as $c_{27}$ in \cite{AOS2019Bauer} multiplied by $q_0!$ and it depends only on $q_0$. Recall that $(C_{\bm{i}})^0_{\delta}$ is defined similar to $K^0_{\delta}$. Then it holds that for $\BS x\in (C_{\bm{i}})^0_{\delta} \cap[-a,a]^p$, 
\begin{align}\label{nonpbound}
    |t(\BS x) - m(\BS x)| \leq c_{22}\bar{r}(m) a^{N+3}M^{-\lambda} = c_{13} a^{N+q_0 + 3} M^{-\lambda}.
\end{align}
Since $m(\BS x)$ takes the same functional form across different cubes, this bound holds for all $\bm{i} \in \{1,\dots,M+1\}^p$. That is, (\ref{nonpbound}) holds for all $\BS x$ in $[-a,a]^p$ except for set
\begin{equation}\label{nonpinvalidset}
\bigcup_{j= 1,\dots,p}\bigcup_{\bm{i} \in \{1,\dots,M+2\}^p} \big\{ x\in \mathbb R^p: |x^{(j)} - x_{\bm{i}}^{(j)} |\leq \delta\big\}
\end{equation}
because of the definition of $(C_{\bm{i}})^0_{\delta}$.


By slightly shifting the whole grid cubes along the $j$th component with the same value that is less than $\frac{2a}{M}$ for a fixed $j \in \{1,\dots,p\}$, we can construct different versions of $t(\BS x)$ that still satisfy (\ref{nonpbound}) for all $\BS x\in[-a,a]^p$ except for those $\BS x$ belonging to
\begin{equation}\label{eq:exception-set}
    \bigcup_{\bm{i} \in \{1,\dots,M+2\}^p} \big\{\BS x\in \mathbb R^p: |x^{(j)} - x_{\bm{i}}^{(j)} |\leq \delta\big\}.
\end{equation}
Here, all the components of $\BS x_{\bm{i}}$ increase by an amount less than $\frac{2a}{M}$, and we have at least $p/\eta$ choices to make the above different versions of sets in \eqref{eq:exception-set} pairwisely disjoint because
$$
\lfloor \frac{2a/M}{2\delta} \rfloor = \lfloor \frac{2a}{M}\frac{2p M}{2a\eta} \rfloor = \lfloor \frac{2p}{\eta} \rfloor \geq p/\eta.
$$
Since the sum of the $\nu$-measures of these sets is less than or equal to one, at least one of them must have measure less than or equal to $\eta/p$. Thus we can shift the $j$th component of $\BS x_{\bm{i}}$ accordingly so that the $\nu$-measure of (\ref{nonpinvalidset}) is less than $\eta$ by the union bound. This completes the proof of Lemma \ref{modthm2}.

\medskip
The following lemma is adapted from Theorem 3 in \cite{AOS2019Bauer}.

\begin{lemma}\label{modthm3}
Let $\BS X$ be a $\mathbb R^p$-valued random variable and $m:\mathbb R^p \to \mathbb R$ satisfy a generalized hierarchical interaction model of order $p^*$ and finite level $l$. For a nonnegative integer $q_0$, let $N\in \mathbb N_0$ with $N\geq q_0$. Assume that in Definition \ref{GHImodel}, all the functions $g_k$, $f_{j,k}$ are polynomial functions up to order $q_0$ and all functions $g_k$ are Lipschitz continuous with Lipschitz constant $L>0$. 
Let the activation function be chosen as the sigmoid function and $t_\sigma$ as defined in Lemma \ref{modthm2}. Let $\lambda_n\in \mathbb R_{+}$, $M_n \in \mathbb N$ be such that $M_n^{\lambda_n} \geq 2(N+|t_\sigma|)(\frac{2^{N+1}}{\sigma^{(N)}(t_\sigma)} + 1)$ for $n$ large enough, and let $a_n\in [1,M_n]$ be an increasing sequence with condition $a_n^{N+q_0 +3} \leq M_n^{\lambda_n}$ satisfied for $n$ sufficiently large. Assume that $\eta_n \in (0,1]$ and parameters in $\mathcal H^{(l)}_{M^*,p^*,p-1,\alpha}$ are defined as $M^* = {p^* + N\choose p^*}(N+1)(M_n+1)^{p^*}$ and $\alpha = \log(n)\frac{M_n^{p^* + \lambda_n(2N+3)+1}}{\eta_n}$. Then for arbitrary $c > 0$ and all $n$ greater than a certain $n_0(c)\in \mathbb N$, there exists a neural network $t\in \mathcal H^{(l)}_{M^*,p^*,p-1,\alpha}$  such that outside of a set of $\mathbb P_{\BS X}$-measure less than or equal to $c\eta_n$, we have 
$$
|t(\BS x) - m(\BS x)| \leq c_{29}a_n^{N+q_0+3}M_n^{\lambda_n}
$$
for all $x \in [-a_n,a_n]^p$. Here, constant $c_{29}$ depends on $c, p, p^*,q_0$, and $N$, but not on $n$. Moreover, $t(\BS x)$ can be chosen such that 
$$
|t(\BS x)| \leq c_{30} a_n^{q_0} M_n^{p^* + N \lambda_n}
$$
holds for all $\BS x\in \mathbb R^p$.
\end{lemma}

\noindent \textit{Proof}. The proof is a simple modification of that of Theorem 3 in \cite{AOS2019Bauer}. For completeness, we still present it here.
The main idea is proof by induction. We only consider the case when $c\eta_n < 1$ because if $c\eta_n \geq 1$, then the assertion is automatically true. 

For a function $m(\BS x) = f(\BS b_1^T\BS x, \dots, \BS b_{p^*}^T\BS x) = f(h(\BS x))$ in which $f:\mathbb R^{p^*} \to \mathbb R$ is a polynomial function up to $q_0$ order and $h: \mathbb R^{p} \to \mathbb R^{p^*}$ is the mapping $h(\BS x) = (\BS b_1^T\BS x, \dots, \BS b_{p^*}^T\BS x)^T$, one can apply Lemma \ref{modthm2} to $f(\BS y)$ to obtain a neural network approximation $\hat f(\BS y) $ for $\BS y \in [-\max_{k =1,\dots, p^*} |\BS b_k|_1 a_n, \max_{k =1,\dots, p^*} |\BS b_k|_1 a_n]^{p^*}$ except for a set $\widetilde{D}_0$ of $\mathbb P_{h(\BS X)}$-measure less than or equal to $c\eta_n$ with an error of 
$$
|\hat f(\BS y) - f(\BS y)| \leq c_{13} (\max_{k =1,\dots, p^*} |\BS b_k|_1 a_n)^{N+q_0 +3}M_n^{-\lambda_n}.
$$
The corresponding neural network approximation $t(\BS x)$ of $m(\BS x)$ can be obtained using the relationship of $t(\BS x) = \hat f(\BS y) = \hat f(h(\BS x))$ due to the fact that $\BS y= h(\BS x)$ is a linear transformation with $\max_{k =1,\dots, p^*} |\BS b_k|_1$ contributing to the bounds of parameters $\mu_i$ and $\theta_{i,l,v}$. That is, to write $t(\BS x)$ in  the form of \eqref{nnform}, we have
\begin{align*}
|\mu_i| &\leq c_{14}(\max_{k =1,\dots, p^*} |\BS b_k|_1 a_n)^{q_0}M_n^{N \lambda_n} \leq \alpha,\\
|\lambda_{i,l}|&\leq M^{p^*+ \lambda_n (N+2)} \leq \alpha,\\
|\theta_{i,l,v} |  &\leq 6\max_{k =1,\dots, p^*} |\BS b_k|_1 \frac{p^*}{\eta_n}M_n^{p^*+ \lambda_n (2N+3) + 1} \leq \alpha.
\end{align*}
Then the $\mathbb P_{\BS x}$-measure of the exception set $D_0:= \{\BS x\in \mathbb R^p| h(\BS x) \in \widetilde D_0\}$ is also bounded by $c\eta_n$. Outside of $D_0$, it holds that
$$
|t(\BS x) - m(\BS x)| \leq c_{13} (\max_{k =1,\dots, p^*} |\BS b_k|_1 a_n)^{N+q_0 +3}M_n^{-\lambda_n}.
$$
On the other hand, we can show that 
$$
|t(\BS x)| \leq M^* \max_{i= 1,\dots, M^*}|\mu_i| \leq c_{31}a_n^{q_0}M_n^{p^* + N \lambda_n}
$$
for all $\BS x\in \mathbb R^p$. Thus, the conclusion is true for the case of $l = 0$. 

When $l > 0$, let $m(\BS x) = \sum_{k=1}^K g_k(f_{1,k}(\BS x),\dots,f_{p^*,k}(\BS x)) = \sum_{k=1}^K g_k(h_k(\BS x))$ with $h_k(\BS x)$ the linear mapping defined analogously to $h(\BS x)$, and the neural network  approximation be $\hat{m}(\BS x)  = \sum_{k=1}^K \hat{g}_k(\hat{f}_{1,k}(\BS x),\dots,\hat{f}_{p^*,k}(\BS x)) = \sum_{k=1}^K \hat{g}_k(\hat{h}_k(\BS x))$, where $\hat{f}_{j,k} \in \mathcal H_{M^*,p^*,p-1,\alpha}^{(l-1)}$ can be found according to the induction hypothesis with $\eta_n$ replaced by $\frac{\eta_n}{2p^*K}$, since $f_{j,k}(\BS x)$ are assumed to be polynomials up to order $q_0$ of $\BS x$. Then each of the terms $|\hat{f}_{j,k}(\BS x) - f_{j,k}(\BS x)|$ can be bounded by $c_{32}a_n^{N+q_0+3}M_n^{-\lambda_n}$ for all $n$ sufficiently large and all $\BS x\in [-a_n,a_n]^p$ outside of a set $D_{j,k}$ of $\mathbb P_{\BS X}$-measure less than or equal to $\frac{c\eta_n}{2p^*K}$. Further, $\hat g_k$ can be chosen from Lemma \ref{modthm2} with $\eta = \frac{c\eta_n}{2K}$ such that 
$$
|\hat{g}_k(\BS y) - g_k(\BS y)| \leq c_{13}(\max_{j=1,\dots p^*} \Vert f_{j,k}\Vert_{\infty} + c_{32})^{N+q_0 + 3} M_n^{-\lambda_n} \leq c_{33} M_n^{-\lambda_n}
$$
holds for all $\BS y\in [ - \max_{j=1,\dots p^*} \Vert f_{j,k}\Vert_{\infty} - c_{32}, \max_{j=1,\dots p^*} \Vert f_{j,k}\Vert_{\infty} + c_{32}]^{p^*}$ except a set $\widetilde D_{k}$ that satisfies $\mathbb P_{h_k(\BS X)}(\widetilde D_k) \leq \frac{\eta_n}{2K}$ ($c_{32}$ can be modified so that $\max_{j=1,\dots p^*} \Vert f_{j,k}\Vert_{\infty} + c_{32} \geq 1 $ is satisfied). Indeed,  $\hat g_k$ can be represented in the form of (\ref{nnform}) with parameters satisfying
\begin{align*}
|\mu_i| &\leq c_{14}(\max_{j=1,\dots p^*} \Vert f_{j,k}\Vert_{\infty} + c_{32})^{q_0}M_n^{N \lambda_n} \leq \alpha,\\
|\lambda_{i,l}|&\leq M^{p^*+ \lambda_n (N+2)} \leq \alpha,\\
|\theta_{i,l,v} | &\leq 6 \frac{p}{\eta_n}M_n^{p^*+ \lambda_n (2N+3) + 1} \leq \alpha,
\end{align*}
which implies that $\hat g_k \in \mathcal H_{M^*,p^*,p^*-1,\alpha}^{(0)}$. 

Let us define $\hat{h}_k^{-1}(\widetilde D_k) := \{\BS x\in \mathbb R^{p^*}| \hat{h}_k(\BS x) \in \widetilde D_k\}$.  Since $\mathbb P_{\hat{h}_k(\BS X)}(\widetilde D_{k}) = \mathbb P_{\BS X}(\hat h_k^{-1}(\widetilde D_k))$, $\hat{g}_k(\hat{h}_k(\BS x))$ approximates $g_k(\hat{h}_k(\BS x))$ with the maximum approximation error given above for all
$$
\BS x\in [-a_n,a_n]^p\setminus \bigcup_{j=1,\dots,p^*}D_{j,k}
$$
outside of the set $D_k :=\hat{h}_k^{-1}(\widetilde D_k) $ of $\mathbb P_{\BS X}$-measure less than or equal to $\frac{c\eta_n}{2K}$. Denote by $t(\BS x) = \hat m(\BS x)$. Then from the derivations above, we have that $t(\BS x) \in \mathcal H_{M^*,p^*,p-1,\alpha}^{(l)}$ and 
\begin{align*}
    |t(\BS x) - m(\BS x)| \leq  & \Big| \sum_{k=1}^K g_k(h_k(\BS x)) - \sum_{k=1}^K g_k(\hat{h}_k(\BS x))\Big| + \Big| \sum_{k=1}^K g_k(\hat{h}_k(\BS x)) - \sum_{k=1}^K \hat{g}_k(\hat{h}_k(\BS x))\Big|\\
    \leq & \sum_{k=1}^K L\sum_{j=1}^{p^*}|f_{j,k}(\BS x) - \hat{f}_{j,k}(\BS x)| + \big| \sum_{k=1}^K g_k(\hat{h}_k(\BS x)) - \sum_{k=1}^K \hat{g}_k(\hat{h}_k(\BS x))\big|\\
    \leq & K L p^*c_{32} a_n^{N+q_0+3} M_n^{-\lambda_n} + K c_{33}M_n^{-\lambda_n} \leq c_{29} a_n^{N+q_0+3} M_n^{-\lambda_n}
\end{align*}
holds for all $\BS x\in [-a_n,a_n]^p$ outside of the set
$$
\bigcup_{\substack{j=1,\dots, p^*\\k=1,\dots, K}} D_{j,k} \cup \bigcup_{k=1,\dots, K}D_k.
$$
Meanwhile, the $\mathbb P_{\BS X}$-measure of the set is bounded by $p^*K\frac{c\eta_n}{2p^* K} + K\frac{c\eta_n}{2K} = c\eta_n$ as desired.

On the other hand, for all $\BS x\in \mathbb R^p$, we can deduce that 
\begin{align*}
    |t(\BS x)| \leq & K {p^* + N \choose p^*} (N+1)(M_n+1)^{p^*}\max_{k=1,\dots,K} c_{14} (\max_{j=1,\dots p^*} \Vert f_{j,k}\Vert_{\infty} + c_{32})^{q_0} M_n^{N \lambda_n}\\
    \leq & c_{34}M_n^{p^* + N \lambda_n},
\end{align*}
which concludes the proof of Lemma \ref{modthm3}. 

\medskip
The following lemma is adapted from Theorem 1 in \cite{AOS2019Bauer}.

\begin{lemma}\label{modthm1}
Let $\{(\BS X_i,Y_i)\}_{i=1}^n$ be an i.i.d. sample collected from an underlying distribution such that $\supp(\BS X)$ is bounded and $\mathbb E\exp(c_1 Y^2)\leq \infty$ for some constant $c_1 > 0$. Assume that Condition \ref{condpoly} with the sigmoid function $\sigma:\mathbb R \to (0,1)$ is satisfied. Let $t_\sigma$ be defined as in Lemma \ref{modthm2} and $q_0$ the highest order of all the polynomials appearing in Condition \ref{condpoly}(i) with arbitrary constant $N\in \mathbb N_0$ such that $N \geq q_0$. Denote by $m_n$ the least-squares estimate defined in  (\ref{neweq006}). Then it holds that 
$$
\mathbb E\int|m_n(\BS x) - m(\BS x)|^2\mathbb P_{\BS X}(d\BS x) \leq c_{60} \log^{p^* + 3}(n) n^{-1}
$$
for all sufficiently large $n$, where constant $c_{60}$ depends on $N$, $q_0$, $t_\sigma$, $p$, and $p^*$, but not on $n$.
\end{lemma}

\noindent \textit{Proof}. Let $a_n = \log^{\frac{3}{2(N+q_0+3)}}(n)$. For a sufficiently large $n$, it holds that  $\supp(\BS X) \in [-a_n,a_n]^p$, which entails that $\mathcal N(\delta,\mathcal G,\Vert \cdot\Vert_{\infty,\supp(\BS X)}) \leq \mathcal N(\delta,\mathcal G,\Vert \cdot\Vert_{\infty,[-a_n,a_n]^p})$ for an arbitrary function space $\mathcal G$ and $\delta > 0$. Then an application of Lemmas 1 and 2 in \cite{AOS2019Bauer} gives
\begin{align}\label{eq: overall-err}
    & \mathbb E\int|m_n(\BS x) - m(\BS x)|^2\mathbb P_{\BS X}(d\BS x)
    \leq c_{7}\log^2(n)\frac{c_{10}\log(n) M^*}{n} \notag\\
    & \quad + 2 \inf_{h\in \mathcal H^{(l)}_{M^*,p^*,p-1,\alpha}}\int |h(\BS x) - m(\BS x)|^2 \mathbb P_{\BS X}(d\BS x).
\end{align}
We next bound the second term on the right-hand side above by using Lemma \ref{modthm3}. Note that the condition that all the $g_k$'s are Lipschitz continuous with some Lipschitz constant $L> 0 $ can be guaranteed by the fact that they are polynomials on a bounded support.

We first define an integer-valued function $n(\lambda)$
$$
n(\lambda) = \inf\{n\in \mathbb N : M_n = \left\lceil n^{\frac{1}{2 \lambda+p^*}}\right\rceil, M_n^{\lambda} \geq 2(N+|t_\sigma|)(\frac{2^{N+1}}{\sigma^{(N)}(t_\sigma)} + 1), n^{\frac{1}{2 \lambda+p^*}}  \geq a_n\}
$$
for all $\lambda\geq q_0 + 1$. Clearly, $n(\lambda)$ is finite and increasing with $\lambda$. Indeed, for $\lambda$ sufficiently large, it follows that 
$$
n(\lambda) = \inf\{n: \frac{\log(n)}{2 \lambda+p^*} \geq \log(\frac{3}{2(N+q_0+3)}) + \log(\log(n))\}.
$$
Starting from $n = n(q_0 + 1)$, let us define 
$\lambda_n = \inf\{\lambda\in \mathbb N: n(\lambda) \geq n + 1\}$. Then we have $n(\lambda_n) \geq n + 1$. Since $n(\lambda) - 1 $ does not satisfy $\frac{\log(n)}{2 \lambda+p^*} \geq \log(\frac{3}{2(N+q_0+3)}) + \log(\log(n))$, it holds that
\begin{align*}
\frac{1}{2 \lambda_n+ p^*} & \leq \frac{\log(\frac{3}{2(N+q_0+3)}) + \log(\log(n(\lambda_n) - 1))}{\log(n(\lambda_n) -1)} \\
&\leq \frac{\log(\frac{3}{2(N+q_0+3)}) + \log(\log(n))}{\log(n)},
\end{align*}
where the second inequality follows from the monotonicity of function $\frac{\log(\frac{3}{2(N+q_0+3)}) + \log(\log(n))}{\log(n)}$ with respect to $n$ when $n$ is large enough.

We set $M_n = \lceil n^{\frac{1}{2 \lambda_n + p^*}} \rceil$ and $\eta_n  = \log^{\frac{3(N+3)}{N+q_0+3}}(n)n^{-\frac{2 \lambda_n(N+1) + 2p^*}{2 \lambda_n + p^*}}$. Denote by 
\[ \alpha_0 = \log(n) \frac{M_n^{p^* + \lambda_n(2N+3) +1}}{\eta_n}. \] 
Then it is seen that 
\[
\alpha_0 = \log^{\frac{-2N+q_0-6}{N+q_0+3}}(n)n^{2\frac{\lambda_n(4N+5)+3p^*}{2\lambda_n + p^*}}\leq n^{4N+6}. 
\]
Choosing constant $c_2$ in Condition \ref{condpoly}(ii) to be larger than $4N + 6$, we can obtain that $\mathcal H^{(l)}_{M^*,p^*,p-1,\alpha_0} \subset \mathcal H^{(l)}_{M^*,p^*,p-1,\alpha}$ with $\alpha = n^{c_2}$ since $\alpha \geq \alpha_0$. Consequently, it follows that
\begin{equation}\label{neweq0030}
   \inf_{h\in \mathcal H^{(l)}_{M^*,p^*,p-1,\alpha}}\int |h(\BS x) - m(\BS x)|^2 \mathbb P_{\BS X}(d\BS x) \leq \inf_{h\in \mathcal H^{(l)}_{M^*,p^*,p-1,\alpha_0}}\int |h(\BS x) - m(\BS x)|^2 \mathbb P_{\BS X}(d\BS x). 
\end{equation}
Denote by $t(\BS x)\in \mathcal H^{(l)}_{M^*,p^*,p-1,\alpha_0}$ the neural network characterized in Lemma \ref{modthm3} with $\alpha$ therein set to be $\alpha_0$ defined above, and let $D_n$ be the exception set in Lemma \ref{modthm3}, outside of which $|t(\BS x) - m(\BS x)| \leq c_{29} a_n^{N+q_0 + 3}M_n^{-\lambda_n}$ holds with $\mathbb P_{\BS X}(D_n) \leq c\eta_n$ for $c = 1$. Then  we can deduce that 
\begin{align}\label{neweq0031}
    &\inf_{h\in \mathcal H^{(l)}_{M^*,p^*,p-1,\alpha_0}}\int |h(\BS x) - m(\BS x)|^2 \mathbb P_{\BS X}(d\BS x)\nonumber\\ 
    \leq &\int |t(\BS x) - m(\BS x)|^2 \mathbb P_{\BS X}(d\BS x)\nonumber\\
    = &\int |t(\BS x) - m(\BS x)|^2 \bm{1}_{D_n^C}\mathbb P_{\BS X}(d\BS x) + \int |t(\BS x) - m(\BS x)|^2 \bm{1}_{D_n}\mathbb P_{\BS X}(d\BS x)\nonumber\\ 
        \leq & (c_{29}a_n^{N+q_0 +3} M_n^{-\lambda_n})^2 + (2c_{30}a_n^{q_0} M_n^{p^*+ N \lambda_n})^2 \eta_n \nonumber\\
    \leq & c_{11} \log^3(n) n^{-\frac{2 \lambda_n}{2 \lambda_n + p^*}}.
\end{align}

Therefore, in view of \eqref{eq: overall-err}, \eqref{neweq0030}, and \eqref{neweq0031}, it holds for $n$ sufficiently large that
\begin{align*}
    &\mathbb E\int|m_n(\BS x) - m(\BS x)|^2\mathbb P_{\BS X}(d\BS x) \\
    \leq &c_4 \log^3(n)n^{\frac{p^*}{2 \lambda_n + p^*}}n^{-1} \\
    \leq &c_4 \log^3(n) n^{p^*\frac{\log(\frac{3}{2(N+q_0+3)}) + \log(\log(n))}{\log(n)}}n^{-1}\\
    \leq &c_{60} \log^{3 + p^*}(n)n^{-1},
\end{align*}
which completes the proof of Lemma \ref{modthm1}.

\subsection{Lemma \ref{modprop1} and its proof} \label{Sec.B.3}

The following lemma gives parallel results to Proposition \ref{mainthm}.

\begin{lemma}\label{modprop1}
Assume that (i) and (iv) of Condition \ref{condRF} and Condition \ref{condpoly} hold with the sigmoid activation function $\sigma(x) = \frac{e^x}{e^x+1}$ in $\mathcal H^{(l)}$. Then the estimator $\widehat{\tau}_{\mathcal D}$ defined in (\ref{neweq007}) satisfies that $|\widehat{\tau}_{\mathcal D} - \tau| = o_P(\log^{\frac{1+p^*}{2}}(n) n^{-1/2})$ as $n_{\mathcal D}:= n \to \infty$.
\end{lemma}

\noindent \textit{Proof}. The proof follows from similar arguments as in the proof of Proposition \ref{mainthm} using the newly established Lemmas \ref{modthm2}--\ref{modthm1} in Section \ref{Sec.B.2} with the caution that dimensionalilty $p$ in Lemmas \ref{modthm2}--\ref{modthm1} needs to be updated to $p+1$ for proving Lemma \ref{modprop1} here. The details are omitted for simplicity.

\section{Additional numerical results} \label{Sec.C}
\begin{figure}[ht]
    \centering
    \includegraphics[width=\linewidth]{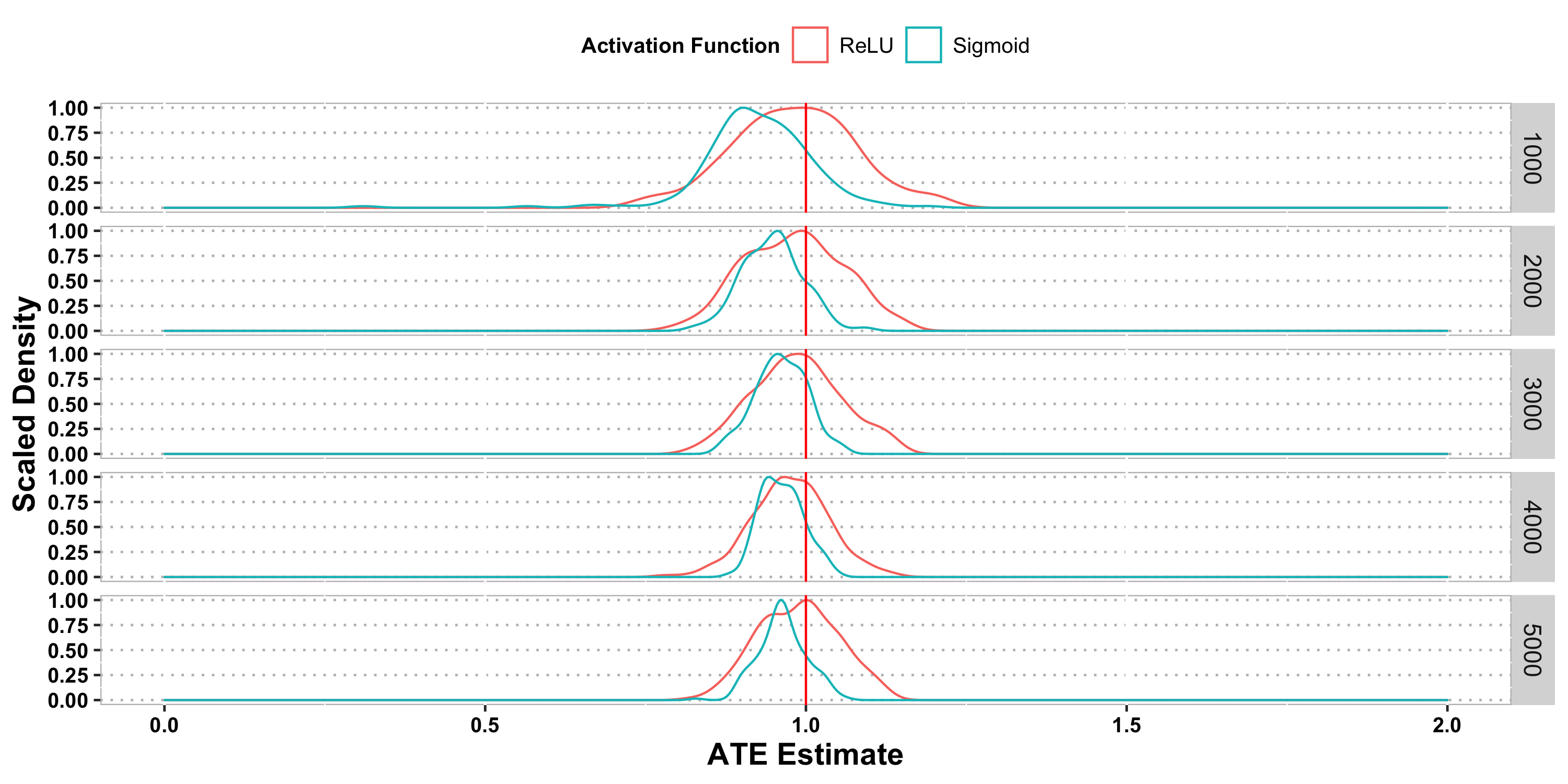}
    \caption{The scaled density of the ATE estimate over 200 replications for different training sample sizes and different activation functions. Here we use a fixed inference sample size of $n = 1000$ and train each network for 100 epochs. The true treatment effect of $\tau = 1$ is shown as a red vertical line.}
    \label{fig:sim_1_overlap_100_epochs}
\end{figure}

\begin{figure}[htp]
    \centering
    \includegraphics[width=\linewidth]{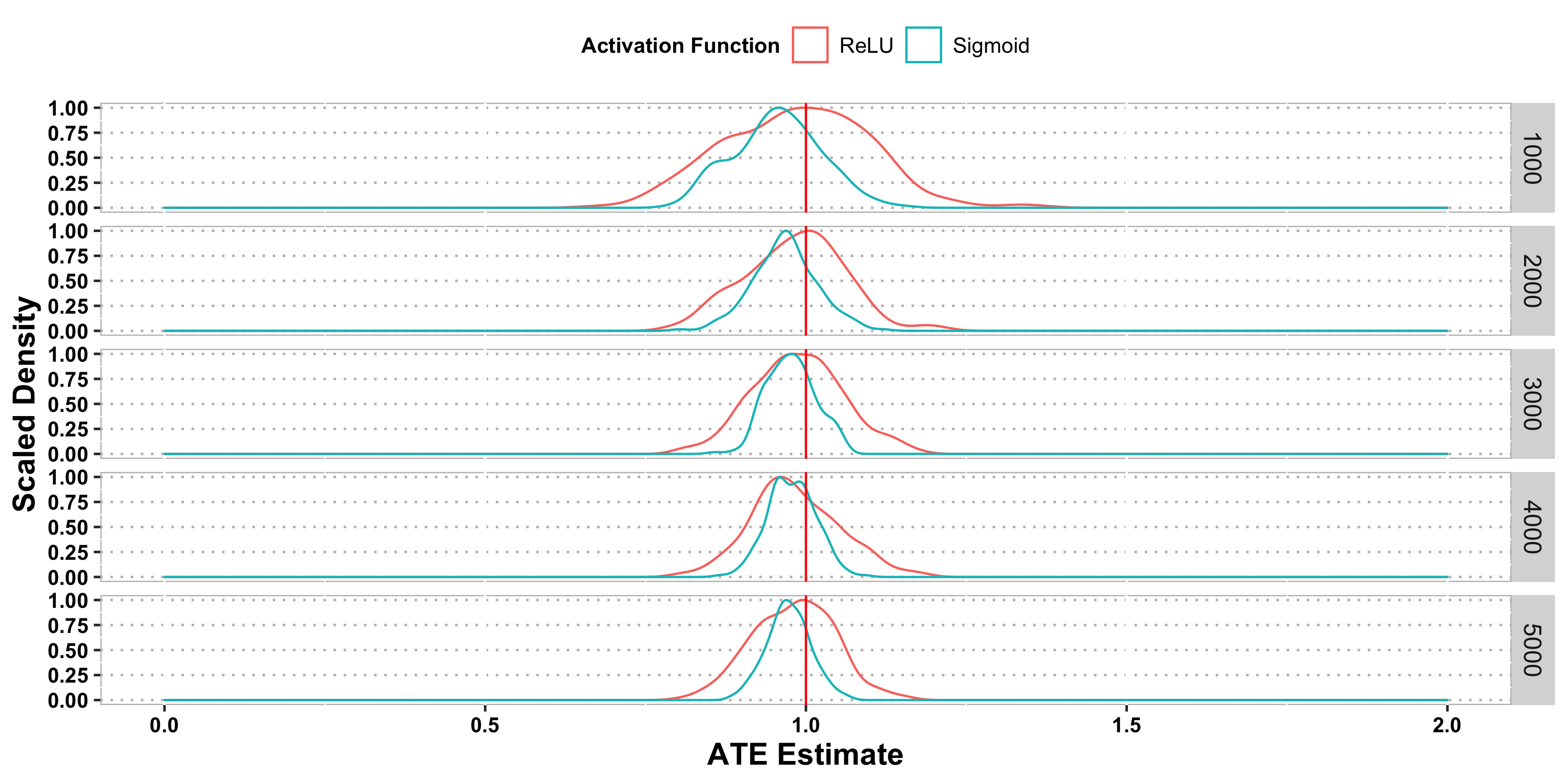}
    \caption{The scaled density of the ATE estimate over 200 replications for different training sample sizes and different activation functions. Here we use a fixed inference sample size of $n = 1000$ and train each network for 200 epochs. The true treatment effect of $\tau = 1$ is shown as a red vertical line.}
    \label{fig:sim_1_overlap_200_epochs}
\end{figure}

\begin{figure}[ht]
    \centering
    \includegraphics[width=\linewidth]{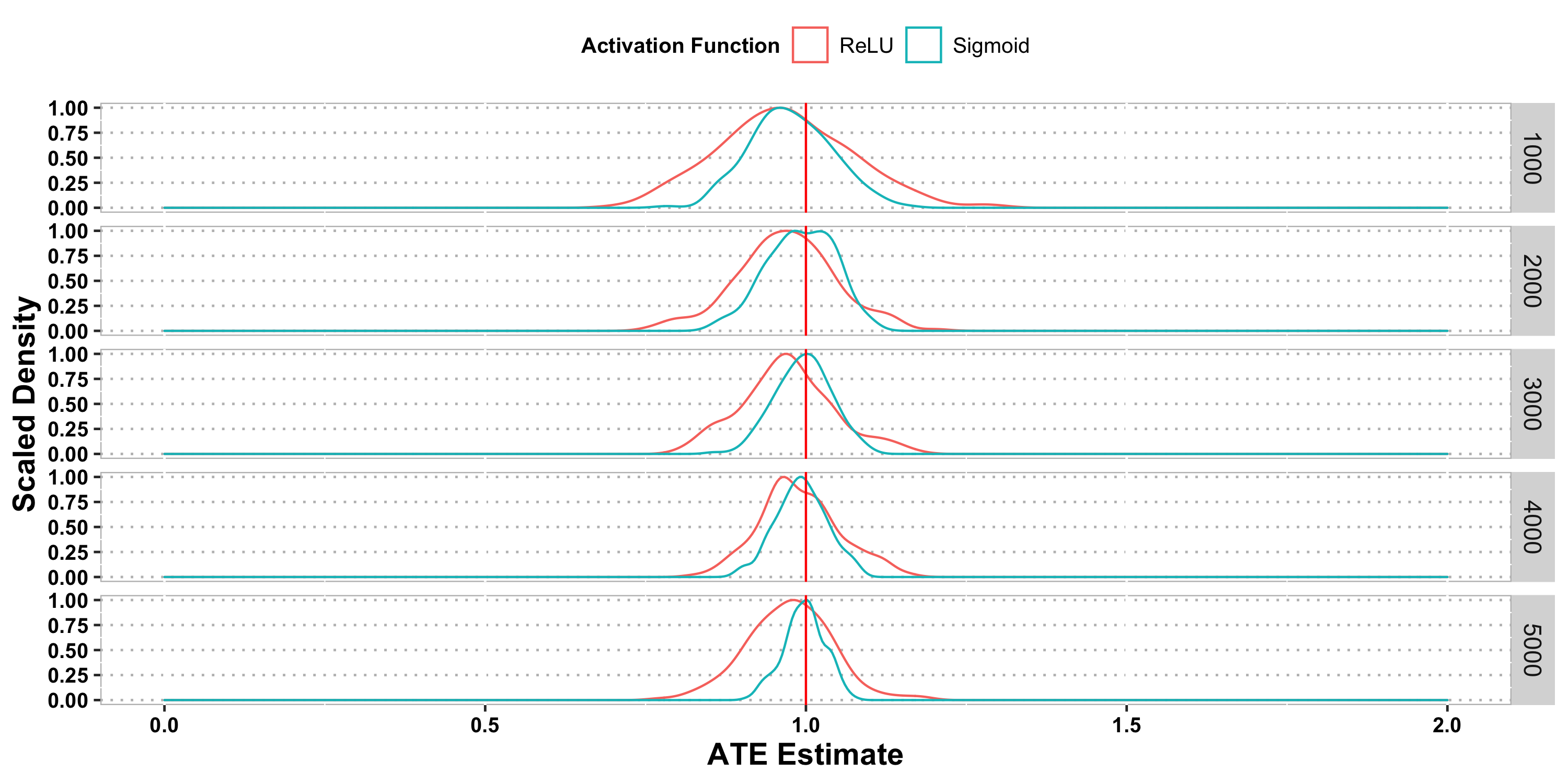}
    \caption{The scaled density of the ATE estimate over 200 replications for different training sample sizes and different activation functions. Here we use a fixed inference sample size of $n = 1000$ and train each network for 400 epochs. The true treatment effect of $\tau = 1$ is shown as a red vertical line.}
    \label{fig:sim_1_overlap_400_epochs}
\end{figure}

\begin{figure}[ht]
    \centering
    \includegraphics[width=\linewidth]{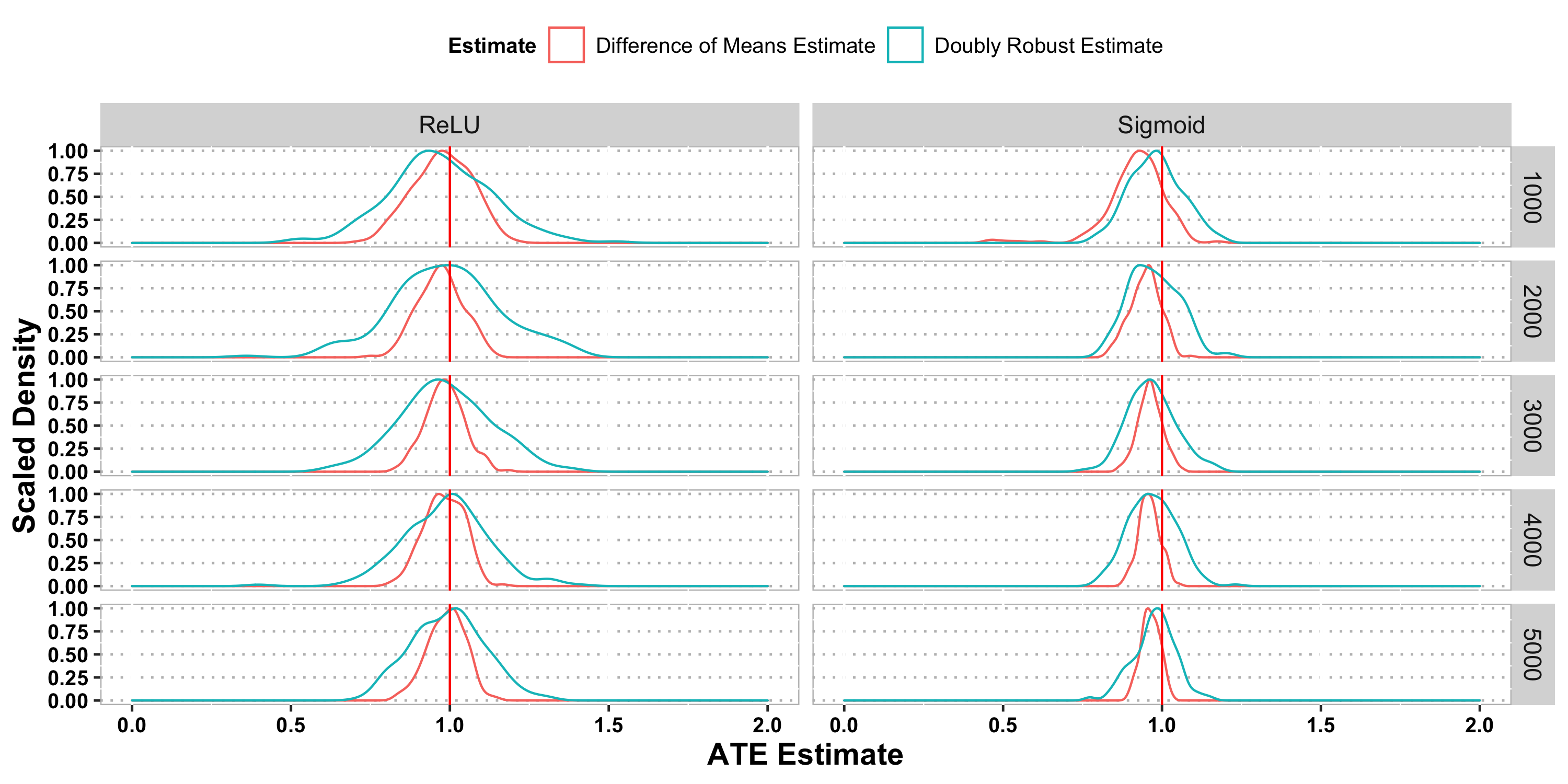}
    \caption{The scaled density of the ATE estimate over 200 replications for different training sample sizes and different activation functions. The red curves correspond to the DNN estimate defined in \eqref{deftauhat} and the blue curves correspond to the doubly robust estimate defined in \eqref{eq: double-robust-split}.   Here we use a fixed inference sample size of $n = 1000$ and train each network for 100 epochs. From top to bottom, the training sample size $n_1$ increases from 1000 to 5000.}
    \label{fig:sim_2_overlap_100_epochs}
\end{figure}

\begin{figure}[ht]
    \centering
    \includegraphics[width=\linewidth]{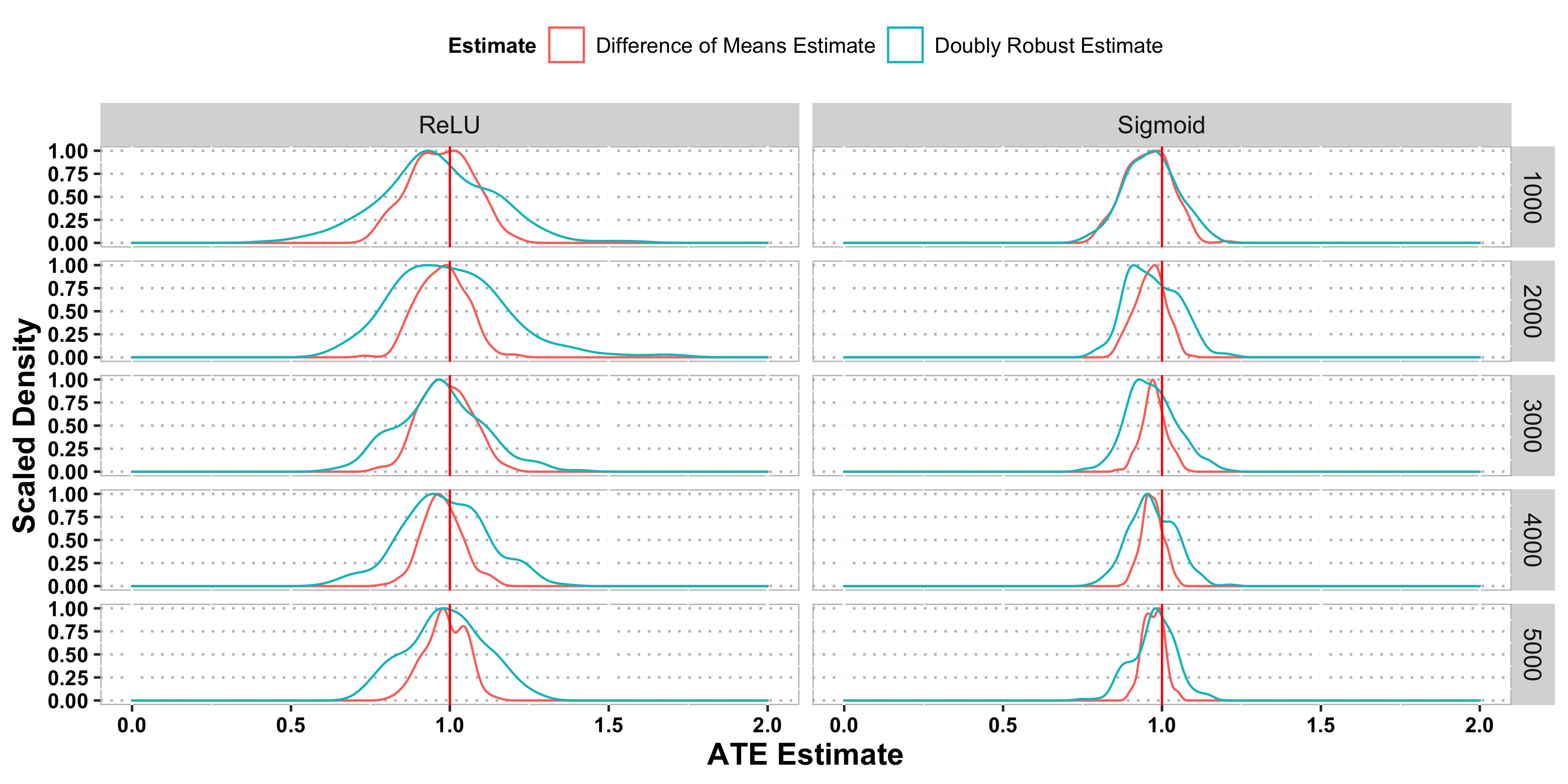}
    \caption{The scaled density of the ATE estimate over 200 replications for different training sample sizes and different activation functions. The red curves correspond to the DNN estimate defined in \eqref{deftauhat} and the blue curves correspond to the doubly robust estimate defined in \eqref{eq: double-robust-split}.   Here we use a fixed inference sample size of $n = 1000$ and train each network for 200 epochs. From top to bottom, the training sample size $n_1$ increases from 1000 to 5000.}
    \label{fig:sim_2_overlap_200_epochs}
\end{figure}

\begin{figure}[ht]
    \centering
    \includegraphics[width=\linewidth]{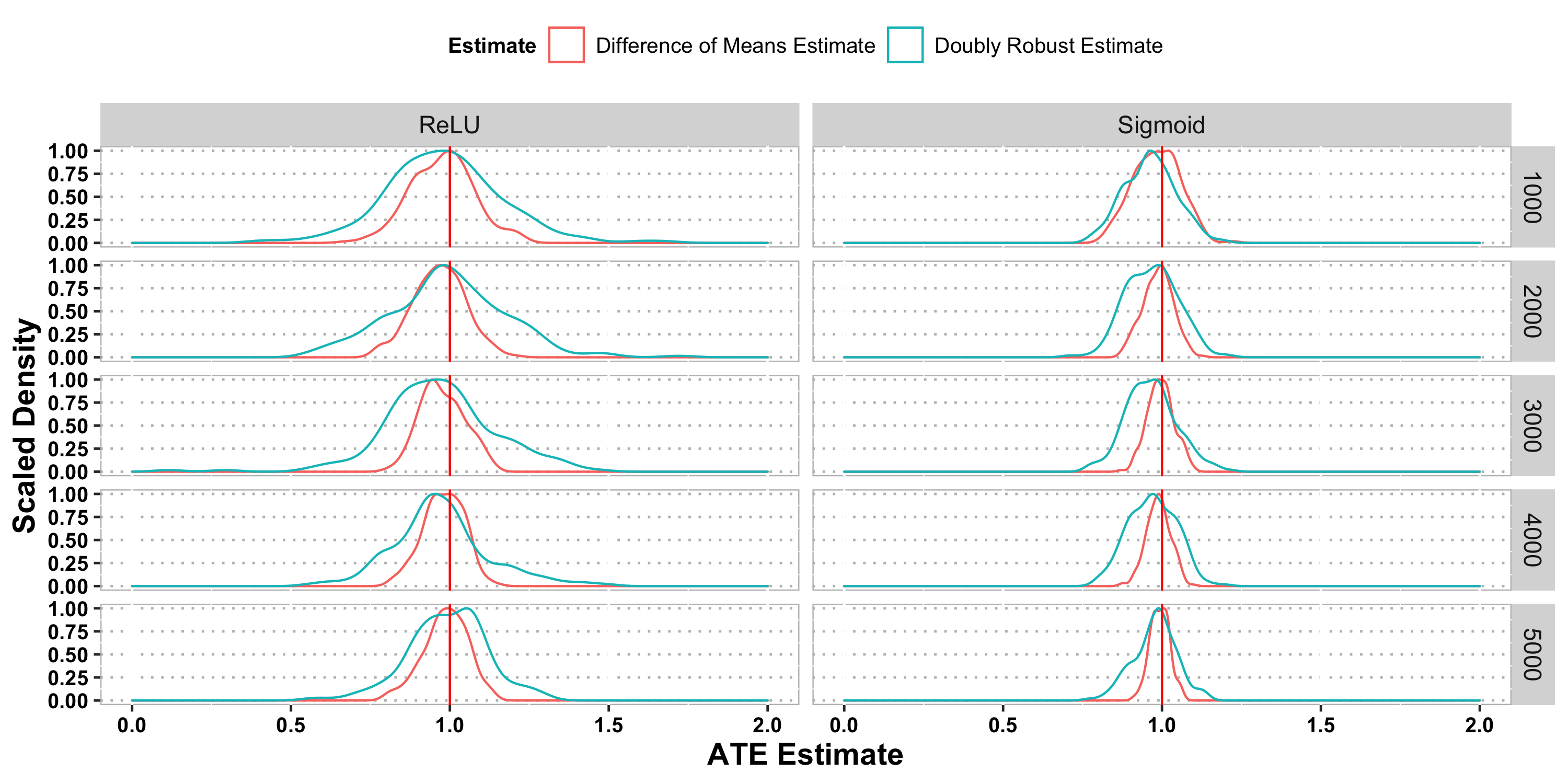}
    \caption{The scaled density of the ATE estimate over 200 replications for different training sample sizes and different activation functions. The red curves correspond to the DNN estimate defined in \eqref{deftauhat} and the blue curves correspond to the doubly robust estimate defined in \eqref{eq: double-robust-split}.   Here we use a fixed inference sample size of $n = 1000$ and train each network for 400 epochs. From top to bottom, the training sample size $n_1$ increases from 1000 to 5000.}
    \label{fig:sim_2_overlap_400_epochs}
\end{figure}

\begin{figure}[ht]
    \centering
    \includegraphics[width=\linewidth]{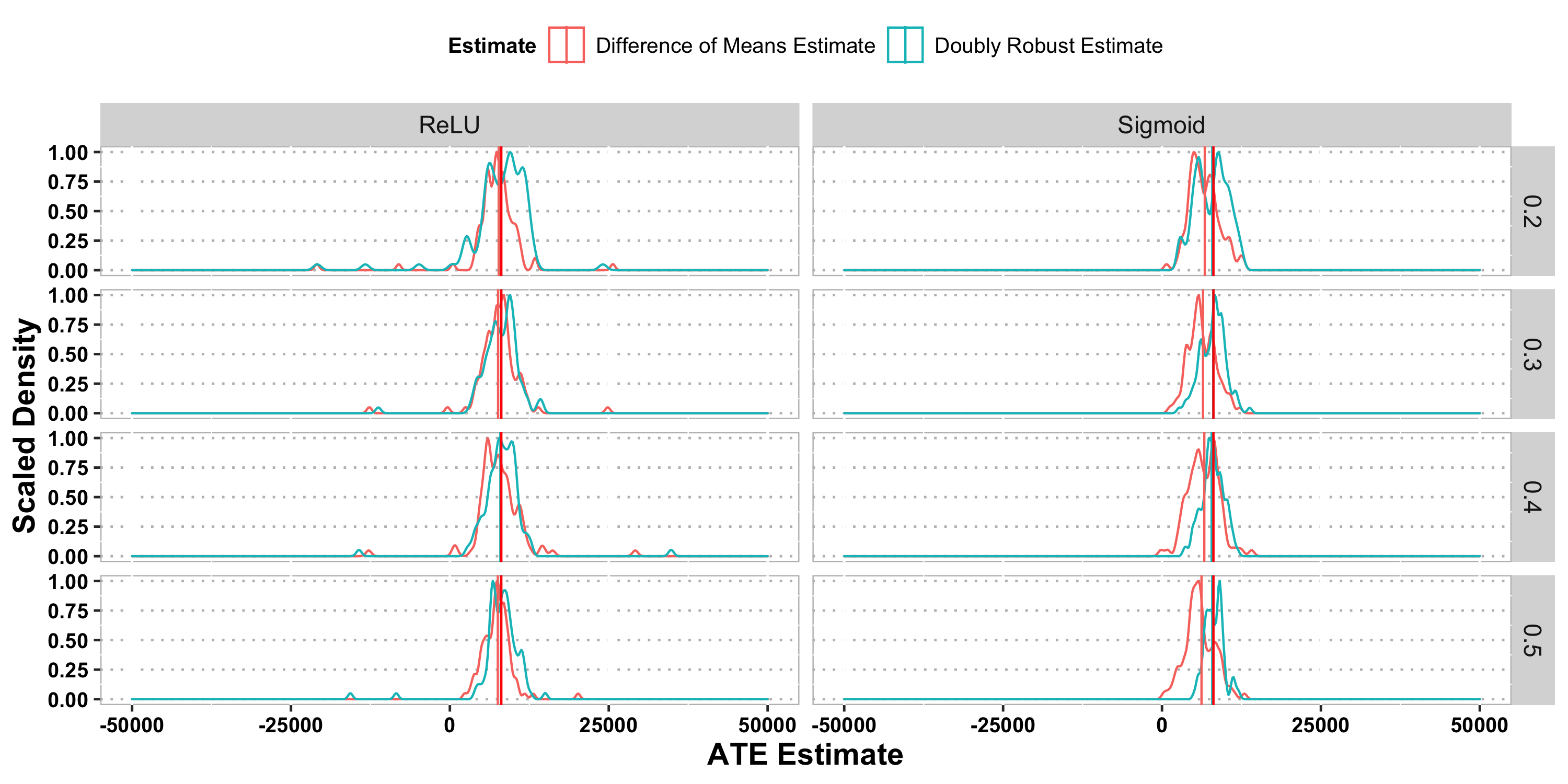}
    \caption{The scaled density of the ATE estimate over 100 replications for different training sample size proportions and different activation functions.  The red curves correspond to the DNN estimate defined in \eqref{deftauhat} and the blue curves correspond to the doubly robust estimate defined in \eqref{eq: double-robust-split}. The red vertical line is the ATE estimate reported in \cite{belloni2017program} from the quadratic spline specification without variable selection of 8093. The rows in the figure correspond to different sizes of the inference set varying from 20\% to 50\% of the data. In this figure, both estimates come from networks trained for 200 epochs.   
    }
    \label{fig:real_data_200_epochs}
\end{figure}

\begin{figure}[ht]
    \centering
    \includegraphics[width=\linewidth]{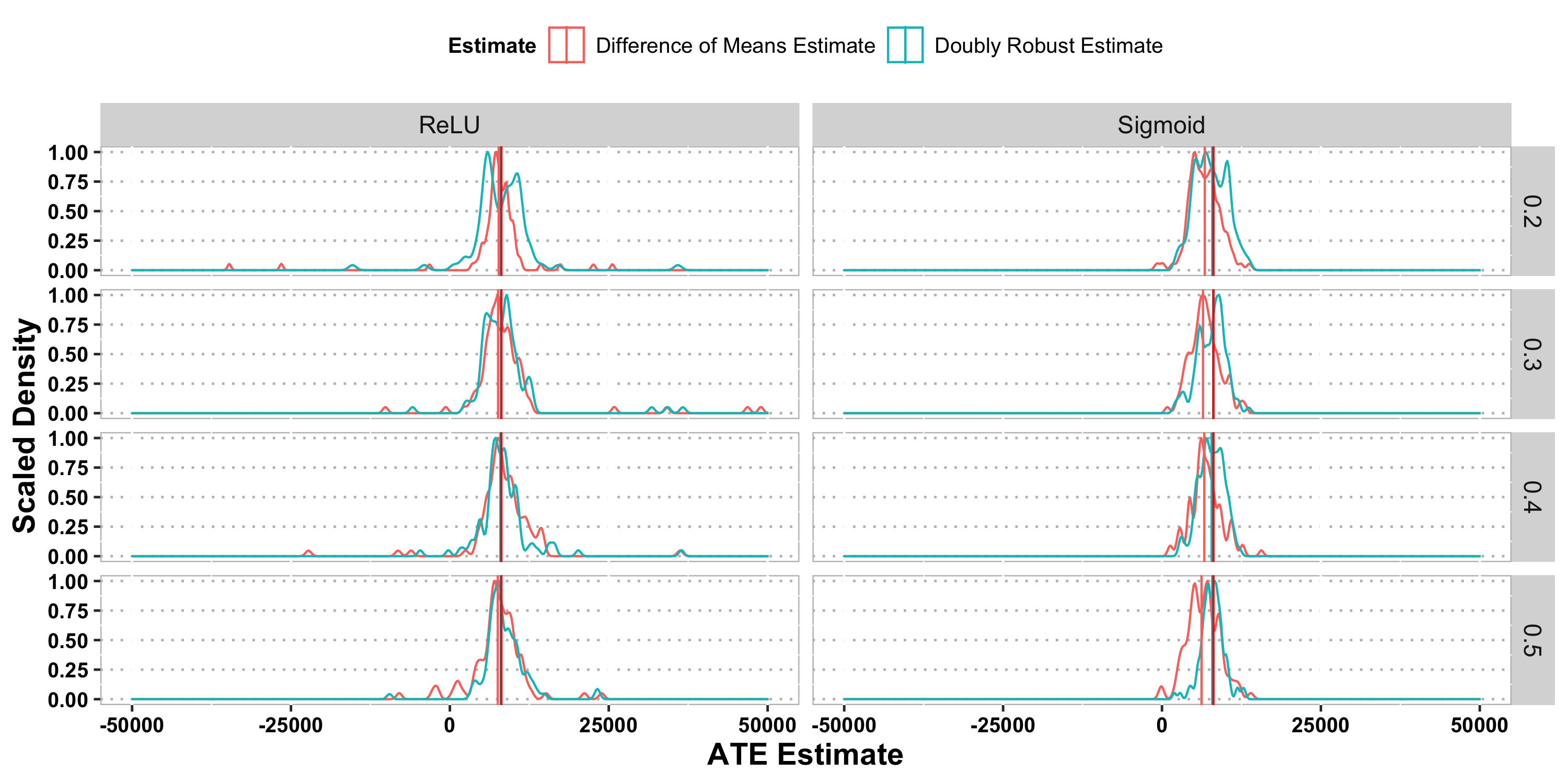}
    \caption{The scaled density of the ATE estimate over 100 replications for different training sample size proportions and different activation functions.  The red curves correspond to the DNN estimate defined in \eqref{deftauhat} and the blue curves correspond to the doubly robust estimate defined in \eqref{eq: double-robust-split}. The red vertical line is the ATE estimate reported in \cite{belloni2017program} from the quadratic spline specification without variable selection of 8093. The rows in the figure correspond to different sizes of the inference set varying from 20\% to 50\% of the data. In this figure, both estimates come from networks trained for 400 epochs.   
    }
    \label{fig:real_data_400_epochs}
\end{figure}

In this section, we present additional simulation and real data results corresponding to different numbers of training epochs. In particular, Figures \ref{fig:sim_1_overlap_100_epochs}--\ref{fig:sim_1_overlap_400_epochs} and Tables \ref{table:sim_1_100_epochs}--\ref{table:sim_1_400_epochs} summarize simulation results parallel to those in Section \ref{Sec.simu1} with the number of epochs ranging from 100 to 400,  Figures \ref{fig:sim_2_overlap_100_epochs}--\ref{fig:sim_2_overlap_400_epochs} and Tables \ref{table:sim_2_100_epochs}--\ref{table:sim_2_400_epochs} summarize simulation results parallel to those in Section \ref{Sec.simu2} with the number of epochs ranging from 100 to 400, and Figures \ref{fig:real_data_200_epochs}--\ref{fig:real_data_400_epochs} and Tables \ref{table:real_data_200_epochs}--\ref{table:real_data_400_epochs} summarize real data results parallel to those in Section \ref{Sec.reda} with the number of epochs ranging from 200 to 400.



\begin{table}[htp]
    \centering
    
\begin{tabular}[t]{cccccc}
\toprule
$n_1$ & Activation & Mean & Median & SD & MSE\\
\midrule
 & ReLU & 0.9808 & 0.9791 & 0.09563 & 0.00947\\

\multirow{-2}{*}{\centering\arraybackslash 1000} & Sigmoid & 0.9239 & 0.9236 & 0.09009 & 0.01386\\
\cmidrule{1-6}
 & ReLU & 0.9807 & 0.9815 & 0.07616 & 0.00614\\

\multirow{-2}{*}{\centering\arraybackslash 2000} & Sigmoid & 0.9479 & 0.9503 & 0.04731 & 0.00494\\
\cmidrule{1-6}
 & ReLU & 0.9867 & 0.9864 & 0.06919 & 0.00494\\

\multirow{-2}{*}{\centering\arraybackslash 3000} & Sigmoid & 0.9612 & 0.9597 & 0.04027 & 0.00312\\
\cmidrule{1-6}
 & ReLU & 0.9754 & 0.9756 & 0.05937 & 0.00411\\

\multirow{-2}{*}{\centering\arraybackslash 4000} & Sigmoid & 0.9627 & 0.9601 & 0.03353 & 0.00251\\
\cmidrule{1-6}
 & ReLU & 0.9883 & 0.9911 & 0.06214 & 0.00398\\

\multirow{-2}{*}{\centering\arraybackslash 5000} & Sigmoid & 0.9636 & 0.9614 & 0.03735 & 0.00271\\
\bottomrule
\end{tabular}

    \caption{Results of the first simulation setting in Section \ref{Sec.simu1} aggregated over 200 replications. In each replication, the networks are trained for 100 epochs.}
    \label{table:sim_1_100_epochs}
\end{table}

\begin{table}[htp]
    \centering
    
\begin{tabular}[t]{cccccc}
\toprule
$n_1$ & Activation & Mean & Median & SD & MSE\\
\midrule
 & ReLU & 0.9847 & 0.9844 & 0.11223 & 0.01277\\

\multirow{-2}{*}{\centering\arraybackslash 1000} & Sigmoid & 0.9555 & 0.9566 & 0.06773 & 0.00654\\
\cmidrule{1-6}
 & ReLU & 0.9823 & 0.9872 & 0.07756 & 0.00630\\

\multirow{-2}{*}{\centering\arraybackslash 2000} & Sigmoid & 0.9672 & 0.9675 & 0.04974 & 0.00354\\
\cmidrule{1-6}
 & ReLU & 0.9877 & 0.9851 & 0.07042 & 0.00508\\

\multirow{-2}{*}{\centering\arraybackslash 3000} & Sigmoid & 0.9771 & 0.9760 & 0.03771 & 0.00194\\
\cmidrule{1-6}
 & ReLU & 0.9837 & 0.9764 & 0.06929 & 0.00504\\

\multirow{-2}{*}{\centering\arraybackslash 4000} & Sigmoid & 0.9779 & 0.9779 & 0.03706 & 0.00186\\
\cmidrule{1-6}
 & ReLU & 0.9806 & 0.9850 & 0.06339 & 0.00437\\

\multirow{-2}{*}{\centering\arraybackslash 5000} & Sigmoid & 0.9732 & 0.9725 & 0.03404 & 0.00187\\
\bottomrule
\end{tabular}

    \caption{Results of the first simulation setting in Section \ref{Sec.simu1} aggregated over 200 replications. In each replication, the networks are trained for 200 epochs.}
    \label{table:sim_1_200_epochs}
\end{table}

\begin{table}[htp]
    \centering
    
\begin{tabular}[t]{cccccc}
\toprule
$n_1$ & Activation & Mean & Median & SD & MSE\\
\midrule
 & ReLU & 0.9670 & 0.9656 & 0.10374 & 0.01180\\

\multirow{-2}{*}{\centering\arraybackslash 1000} & Sigmoid & 0.9764 & 0.9726 & 0.06355 & 0.00457\\
\cmidrule{1-6}
 & ReLU & 0.9692 & 0.9711 & 0.07743 & 0.00692\\

\multirow{-2}{*}{\centering\arraybackslash 2000} & Sigmoid & 0.9917 & 0.9913 & 0.05239 & 0.00280\\
\cmidrule{1-6}
 & ReLU & 0.9711 & 0.9687 & 0.07216 & 0.00602\\

\multirow{-2}{*}{\centering\arraybackslash 3000} & Sigmoid & 0.9958 & 0.9974 & 0.04351 & 0.00190\\
\cmidrule{1-6}
 & ReLU & 0.9903 & 0.9834 & 0.06273 & 0.00401\\

\multirow{-2}{*}{\centering\arraybackslash 4000} & Sigmoid & 0.9930 & 0.9932 & 0.03933 & 0.00159\\
\cmidrule{1-6}
 & ReLU & 0.9741 & 0.9773 & 0.06648 & 0.00507\\

\multirow{-2}{*}{\centering\arraybackslash 5000} & Sigmoid & 0.9971 & 0.9986 & 0.03195 & 0.00102\\
\bottomrule
\end{tabular}

    \caption{Results of the first simulation setting in Section \ref{Sec.simu1} aggregated over 200 replications. In each replication, the networks are trained for 400 epochs.}
    \label{table:sim_1_400_epochs}
\end{table}


\begin{table}[htp]
    \centering
    
\begin{tabular}[t]{ccccccc}
\toprule
$n_1$ & Estimate Type & Activation & Mean & Median & SD & MSE\\
\midrule
 &  & ReLU & 0.9763 & 0.9767 & 0.09150 & 0.00889\\
\cmidrule{3-7}
 & \multirow{-2}{*}{\centering\arraybackslash Difference of Means Estimate} & Sigmoid & 0.9172 & 0.9251 & 0.09622 & 0.01607\\
\cmidrule{2-7}
 &  & ReLU & 0.9703 & 0.9659 & 0.16154 & 0.02684\\
\cmidrule{3-7}
\multirow{-4}{*}{\centering\arraybackslash 1000} & \multirow{-2}{*}{\centering\arraybackslash Doubly Robust Estimate} & Sigmoid & 0.9808 & 0.9788 & 0.08293 & 0.00721\\
\cmidrule{1-7}
 &  & ReLU & 0.9707 & 0.9692 & 0.07016 & 0.00575\\
\cmidrule{3-7}
 & \multirow{-2}{*}{\centering\arraybackslash Difference of Means Estimate} & Sigmoid & 0.9467 & 0.9516 & 0.04716 & 0.00505\\
\cmidrule{2-7}
 &  & ReLU & 0.9891 & 0.9851 & 0.17272 & 0.02980\\
\cmidrule{3-7}
\multirow{-4}{*}{\centering\arraybackslash 2000} & \multirow{-2}{*}{\centering\arraybackslash Doubly Robust Estimate} & Sigmoid & 0.9742 & 0.9669 & 0.07878 & 0.00684\\
\cmidrule{1-7}
 &  & ReLU & 0.9845 & 0.9852 & 0.06145 & 0.00400\\
\cmidrule{3-7}
 & \multirow{-2}{*}{\centering\arraybackslash Difference of Means Estimate} & Sigmoid & 0.9626 & 0.9615 & 0.03774 & 0.00282\\
\cmidrule{2-7}
 &  & ReLU & 0.9880 & 0.9795 & 0.14298 & 0.02049\\
\cmidrule{3-7}
\multirow{-4}{*}{\centering\arraybackslash 3000} & \multirow{-2}{*}{\centering\arraybackslash Doubly Robust Estimate} & Sigmoid & 0.9678 & 0.9657 & 0.07579 & 0.00675\\
\cmidrule{1-7}
 &  & ReLU & 0.9813 & 0.9812 & 0.06223 & 0.00420\\
\cmidrule{3-7}
 & \multirow{-2}{*}{\centering\arraybackslash Difference of Means Estimate} & Sigmoid & 0.9597 & 0.9590 & 0.03335 & 0.00273\\
\cmidrule{2-7}
 &  & ReLU & 0.9884 & 0.9975 & 0.13479 & 0.01821\\
\cmidrule{3-7}
\multirow{-4}{*}{\centering\arraybackslash 4000} & \multirow{-2}{*}{\centering\arraybackslash Doubly Robust Estimate} & Sigmoid & 0.9695 & 0.9662 & 0.07448 & 0.00645\\
\cmidrule{1-7}
 &  & ReLU & 0.9892 & 0.9930 & 0.05888 & 0.00357\\
\cmidrule{3-7}
 & \multirow{-2}{*}{\centering\arraybackslash Difference of Means Estimate} & Sigmoid & 0.9639 & 0.9640 & 0.02923 & 0.00215\\
\cmidrule{2-7}
 &  & ReLU & 0.9941 & 0.9972 & 0.11426 & 0.01302\\
\cmidrule{3-7}
\multirow{-4}{*}{\centering\arraybackslash 5000} & \multirow{-2}{*}{\centering\arraybackslash Doubly Robust Estimate} & Sigmoid & 0.9759 & 0.9810 & 0.06615 & 0.00494\\
\bottomrule
\end{tabular}
    
    \caption{The simulation results corresponding to Figure \ref{fig:sim_2_overlap_100_epochs} for 100 training epochs.}
    \label{table:sim_2_100_epochs}
\end{table}

\begin{table}[htp]
    \centering
    
\begin{tabular}[t]{ccccccc}
\toprule
$n_1$ & Estimate Type & Activation & Mean & Median & SD & MSE\\
\midrule
 &  & ReLU & 0.9735 & 0.9764 & 0.09660 & 0.00999\\
\cmidrule{3-7}
 & \multirow{-2}{*}{\centering\arraybackslash Difference of Means Estimate} & Sigmoid & 0.9553 & 0.9565 & 0.07190 & 0.00714\\
\cmidrule{2-7}
 &  & ReLU & 0.9602 & 0.9495 & 0.17878 & 0.03339\\
\cmidrule{3-7}
\multirow{-4}{*}{\centering\arraybackslash 1000} & \multirow{-2}{*}{\centering\arraybackslash Doubly Robust Estimate} & Sigmoid & 0.9654 & 0.9649 & 0.08281 & 0.00802\\
\cmidrule{1-7}
 &  & ReLU & 0.9767 & 0.9787 & 0.07471 & 0.00610\\
\cmidrule{3-7}
 & \multirow{-2}{*}{\centering\arraybackslash Difference of Means Estimate} & Sigmoid & 0.9626 & 0.9655 & 0.04696 & 0.00359\\
\cmidrule{2-7}
 &  & ReLU & 1.0010 & 0.9830 & 0.17861 & 0.03174\\
\cmidrule{3-7}
\multirow{-4}{*}{\centering\arraybackslash 2000} & \multirow{-2}{*}{\centering\arraybackslash Doubly Robust Estimate} & Sigmoid & 0.9718 & 0.9620 & 0.08075 & 0.00728\\
\cmidrule{1-7}
 &  & ReLU & 0.9862 & 0.9814 & 0.07825 & 0.00628\\
\cmidrule{3-7}
 & \multirow{-2}{*}{\centering\arraybackslash Difference of Means Estimate} & Sigmoid & 0.9749 & 0.9722 & 0.03776 & 0.00205\\
\cmidrule{2-7}
 &  & ReLU & 0.9711 & 0.9655 & 0.13425 & 0.01876\\
\cmidrule{3-7}
\multirow{-4}{*}{\centering\arraybackslash 3000} & \multirow{-2}{*}{\centering\arraybackslash Doubly Robust Estimate} & Sigmoid & 0.9688 & 0.9657 & 0.07743 & 0.00694\\
\cmidrule{1-7}
 &  & ReLU & 0.9766 & 0.9699 & 0.06312 & 0.00451\\
\cmidrule{3-7}
 & \multirow{-2}{*}{\centering\arraybackslash Difference of Means Estimate} & Sigmoid & 0.9695 & 0.9694 & 0.03313 & 0.00202\\
\cmidrule{2-7}
 &  & ReLU & 0.9826 & 0.9743 & 0.13325 & 0.01797\\
\cmidrule{3-7}
\multirow{-4}{*}{\centering\arraybackslash 4000} & \multirow{-2}{*}{\centering\arraybackslash Doubly Robust Estimate} & Sigmoid & 0.9699 & 0.9596 & 0.07169 & 0.00602\\
\cmidrule{1-7}
 &  & ReLU & 0.9854 & 0.9844 & 0.06355 & 0.00423\\
\cmidrule{3-7}
 & \multirow{-2}{*}{\centering\arraybackslash Difference of Means Estimate} & Sigmoid & 0.9729 & 0.9737 & 0.03164 & 0.00173\\
\cmidrule{2-7}
 &  & ReLU & 0.9893 & 0.9891 & 0.12253 & 0.01505\\
\cmidrule{3-7}
\multirow{-4}{*}{\centering\arraybackslash 5000} & \multirow{-2}{*}{\centering\arraybackslash Doubly Robust Estimate} & Sigmoid & 0.9752 & 0.9776 & 0.06742 & 0.00514\\
\bottomrule
\end{tabular}
    
    \caption{The simulation results corresponding to Figure \ref{fig:sim_2_overlap_200_epochs} for 200 training epochs.}
    \label{table:sim_2_200_epochs}
\end{table}

\begin{table}[htp]
    \centering
    
\begin{tabular}[t]{ccccccc}
\toprule
$n_1$ & Estimate Type & Activation & Mean & Median & SD & MSE\\
\midrule
 &  & ReLU & 0.9764 & 0.9798 & 0.09930 & 0.01037\\
\cmidrule{3-7}
 & \multirow{-2}{*}{\centering\arraybackslash Difference of Means Estimate} & Sigmoid & 0.9833 & 0.9834 & 0.07282 & 0.00556\\
\cmidrule{2-7}
 &  & ReLU & 0.9809 & 0.9732 & 0.18509 & 0.03445\\
\cmidrule{3-7}
\multirow{-4}{*}{\centering\arraybackslash 1000} & \multirow{-2}{*}{\centering\arraybackslash Doubly Robust Estimate} & Sigmoid & 0.9626 & 0.9633 & 0.08055 & 0.00785\\
\cmidrule{1-7}
 &  & ReLU & 0.9666 & 0.9646 & 0.08197 & 0.00780\\
\cmidrule{3-7}
 & \multirow{-2}{*}{\centering\arraybackslash Difference of Means Estimate} & Sigmoid & 0.9869 & 0.9904 & 0.04714 & 0.00238\\
\cmidrule{2-7}
 &  & ReLU & 0.9948 & 0.9835 & 0.18331 & 0.03346\\
\cmidrule{3-7}
\multirow{-4}{*}{\centering\arraybackslash 2000} & \multirow{-2}{*}{\centering\arraybackslash Doubly Robust Estimate} & Sigmoid & 0.9678 & 0.9692 & 0.08182 & 0.00770\\
\cmidrule{1-7}
 &  & ReLU & 0.9770 & 0.9660 & 0.06987 & 0.00538\\
\cmidrule{3-7}
 & \multirow{-2}{*}{\centering\arraybackslash Difference of Means Estimate} & Sigmoid & 0.9959 & 0.9971 & 0.04165 & 0.00174\\
\cmidrule{2-7}
 &  & ReLU & 0.9689 & 0.9614 & 0.17931 & 0.03296\\
\cmidrule{3-7}
\multirow{-4}{*}{\centering\arraybackslash 3000} & \multirow{-2}{*}{\centering\arraybackslash Doubly Robust Estimate} & Sigmoid & 0.9660 & 0.9647 & 0.08015 & 0.00755\\
\cmidrule{1-7}
 &  & ReLU & 0.9802 & 0.9814 & 0.06417 & 0.00449\\
\cmidrule{3-7}
 & \multirow{-2}{*}{\centering\arraybackslash Difference of Means Estimate} & Sigmoid & 0.9888 & 0.9895 & 0.03778 & 0.00155\\
\cmidrule{2-7}
 &  & ReLU & 0.9723 & 0.9590 & 0.15317 & 0.02411\\
\cmidrule{3-7}
\multirow{-4}{*}{\centering\arraybackslash 4000} & \multirow{-2}{*}{\centering\arraybackslash Doubly Robust Estimate} & Sigmoid & 0.9696 & 0.9696 & 0.07448 & 0.00645\\
\cmidrule{1-7}
 &  & ReLU & 0.9864 & 0.9919 & 0.06906 & 0.00493\\
\cmidrule{3-7}
 & \multirow{-2}{*}{\centering\arraybackslash Difference of Means Estimate} & Sigmoid & 0.9931 & 0.9944 & 0.03044 & 0.00097\\
\cmidrule{2-7}
 &  & ReLU & 0.9928 & 0.9943 & 0.12544 & 0.01571\\
\cmidrule{3-7}
\multirow{-4}{*}{\centering\arraybackslash 5000} & \multirow{-2}{*}{\centering\arraybackslash Doubly Robust Estimate} & Sigmoid & 0.9780 & 0.9852 & 0.06821 & 0.00511\\
\bottomrule
\end{tabular}
    
    \caption{The simulation results corresponding to Figure \ref{fig:sim_2_overlap_400_epochs} for 400 training epochs.}
    \label{table:sim_2_400_epochs}
\end{table}


\begin{table}[htp]
    \centering
    
\begin{tabular}[t]{ccccc}
\toprule
Inference Proportion & Estimate Type & Activation & Median & Robust SD\\
\midrule
 &  & ReLU & 7328 & 2008\\
\cmidrule{3-5}
 & \multirow{-2}{*}{\centering\arraybackslash Difference of Means Estimate} & Sigmoid & 6300 & 2261\\
\cmidrule{2-5}
 &  & ReLU & 8683 & 3528\\
\cmidrule{3-5}
\multirow{-4}{*}{\centering\arraybackslash 0.2} & \multirow{-2}{*}{\centering\arraybackslash Doubly Robust Estimate} & Sigmoid & 8114 & 3052\\
\cmidrule{1-5}
 &  & ReLU & 7624 & 2154\\
\cmidrule{3-5}
 & \multirow{-2}{*}{\centering\arraybackslash Difference of Means Estimate} & Sigmoid & 5960 & 2152\\
\cmidrule{2-5}
 &  & ReLU & 8159 & 2349\\
\cmidrule{3-5}
\multirow{-4}{*}{\centering\arraybackslash 0.3} & \multirow{-2}{*}{\centering\arraybackslash Doubly Robust Estimate} & Sigmoid & 8281 & 2084\\
\cmidrule{1-5}
 &  & ReLU & 7546 & 2428\\
\cmidrule{3-5}
 & \multirow{-2}{*}{\centering\arraybackslash Difference of Means Estimate} & Sigmoid & 6526 & 2443\\
\cmidrule{2-5}
 &  & ReLU & 8220 & 2301\\
\cmidrule{3-5}
\multirow{-4}{*}{\centering\arraybackslash 0.4} & \multirow{-2}{*}{\centering\arraybackslash Doubly Robust Estimate} & Sigmoid & 8013 & 1689\\
\cmidrule{1-5}
 &  & ReLU & 7472 & 1831\\
\cmidrule{3-5}
 & \multirow{-2}{*}{\centering\arraybackslash Difference of Means Estimate} & Sigmoid & 5819 & 2208\\
\cmidrule{2-5}
 &  & ReLU & 8292 & 1960\\
\cmidrule{3-5}
\multirow{-4}{*}{\centering\arraybackslash 0.5} & \multirow{-2}{*}{\centering\arraybackslash Doubly Robust Estimate} & Sigmoid & 8184 & 1462\\
\bottomrule
\end{tabular}

    \caption{The real data results corresponding to Figure \ref{fig:real_data_200_epochs} for 200 training epochs.}
    \label{table:real_data_200_epochs}
\end{table}

\begin{table}[htp]
    \centering
    
\begin{tabular}[t]{ccccc}
\toprule
Inference Proportion & Estimate Type & Activation & Median & Robust SD\\
\midrule
 &  & ReLU & 7711 & 1664\\
\cmidrule{3-5}
 & \multirow{-2}{*}{\centering\arraybackslash Difference of Means Estimate} & Sigmoid & 6462 & 2212\\
\cmidrule{2-5}
 &  & ReLU & 8200 & 3410\\
\cmidrule{3-5}
\multirow{-4}{*}{\centering\arraybackslash 0.2} & \multirow{-2}{*}{\centering\arraybackslash Doubly Robust Estimate} & Sigmoid & 7495 & 3057\\
\cmidrule{1-5}
 &  & ReLU & 7761 & 2252\\
\cmidrule{3-5}
 & \multirow{-2}{*}{\centering\arraybackslash Difference of Means Estimate} & Sigmoid & 6650 & 2115\\
\cmidrule{2-5}
 &  & ReLU & 7987 & 2547\\
\cmidrule{3-5}
\multirow{-4}{*}{\centering\arraybackslash 0.3} & \multirow{-2}{*}{\centering\arraybackslash Doubly Robust Estimate} & Sigmoid & 8118 & 2252\\
\cmidrule{1-5}
 &  & ReLU & 8064 & 2518\\
\cmidrule{3-5}
 & \multirow{-2}{*}{\centering\arraybackslash Difference of Means Estimate} & Sigmoid & 6722 & 1942\\
\cmidrule{2-5}
 &  & ReLU & 7970 & 2257\\
\cmidrule{3-5}
\multirow{-4}{*}{\centering\arraybackslash 0.4} & \multirow{-2}{*}{\centering\arraybackslash Doubly Robust Estimate} & Sigmoid & 7840 & 1967\\
\cmidrule{1-5}
 &  & ReLU & 7676 & 2400\\
\cmidrule{3-5}
 & \multirow{-2}{*}{\centering\arraybackslash Difference of Means Estimate} & Sigmoid & 6571 & 2456\\
\cmidrule{2-5}
 &  & ReLU & 7935 & 2332\\
\cmidrule{3-5}
\multirow{-4}{*}{\centering\arraybackslash 0.5} & \multirow{-2}{*}{\centering\arraybackslash Doubly Robust Estimate} & Sigmoid & 7959 & 1505\\
\bottomrule
\end{tabular}

    \caption{The real data results corresponding to Figure \ref{fig:real_data_200_epochs} for 400 training epochs.}
    \label{table:real_data_400_epochs}
\end{table}

\end{document}